\documentclass[runningheads]{llncs}

\usepackage{eccv}

\usepackage{eccvabbrv}

\usepackage{graphicx}
\usepackage{float}
\usepackage{booktabs}
\usepackage{subcaption}
\usepackage{amsmath}

\usepackage[dvipsnames]{xcolor}
\usepackage{mathtools}
\usepackage{makecell, tabularx}
\newcolumntype{Y}{>{\centering\arraybackslash}X}
\usepackage{rotating}
\usepackage[export]{adjustbox}
\usepackage{pgfplots}
\pgfplotsset{compat=1.18}
\usepackage{multirow}
\usepackage{pifont}   %
\def\subparagraph{} %
\usepackage{titlesec}

\usepackage[accsupp]{axessibility}  %

\usepackage{hyperref}

\usepackage{orcidlink}

\begin{document}

\title{PointGT: Simultaneous Geometry and Texture Editing for Point-Based Representations} 

\titlerunning{PointGT: Point-Based Geometry and Texture Editing}

\author{Yanshu Zhang\inst{1}\thanks{Corresponding author.}\orcidlink{0009-0002-6046-3966} \and
George Shramko\inst{1}\orcidlink{0009-0001-4664-4368} \and
Pratul P. Srinivasan\inst{4}\orcidlink{0009-0008-8268-3285} \and Ke Li\inst{1,2,3}\orcidlink{0000-0002-3229-271X}}

\authorrunning{Y. Zhang et al.}

\institute{Simon Fraser University, Canada \and Alberta Machine Intelligence Institute (Amii), Canada \and Canadian Institute for Advanced Research (CIFAR), Canada \and Google DeepMind, USA \\ \email{\{yanshu\_zhang, george\_shramko, keli\}@sfu.ca} \\ \email{pratul.srinivasan@gmail.com}}

\maketitle

\begin{abstract}
  We present PointGT, a point-based 3D representation that enables simultaneous editing of object geometry and appearance. Existing reconstruction and view synthesis techniques produce volumetric 3D representations that are high-quality and photorealistic, but are difficult to edit. In particular, recent efforts to enable texture editing for 3D Gaussian Splatting representations are not compatible with geometry edits and deformations. Our method combines a point-based representation that is well-suited for geometry deformations with a learned UV mapping technique that enables high-resolution texture editing. We show that PointGT enables fine-grained editing of both geometry and texture in point-based neural representations with high rendering quality. Project page: \href{https://zvict.github.io/pointgt/}{https://zvict.github.io/pointgt/}.
\end{abstract}

\begin{figure*}
    \centering
    \captionsetup{type=figure}
    \captionsetup[subfloat]{labelformat=empty}
    \centering
    \footnotesize
    \includegraphics[width=1\hsize,valign=m]{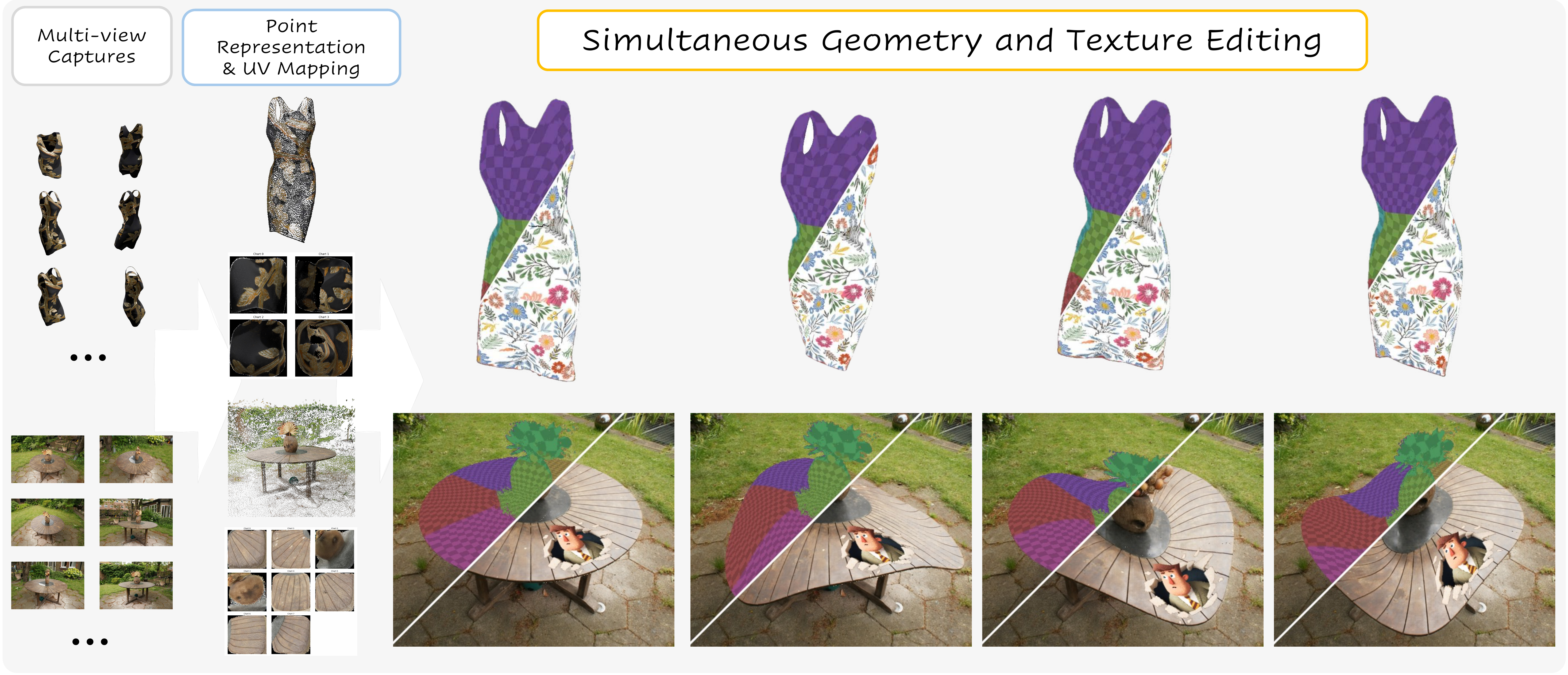}
    \captionof{figure}{\textbf{Overview} Given multi-view captures (left), PointGT reconstructs a point-based 3D representation and learns a global UV mapping onto 2D charts (middle left). By maintaining a deformation-aware correspondence between canonical and deformed space, PointGT ensures persistent texture edits under non-rigid geometry changes. Users can simultaneously deform and retexture a reconstructed dress (top) or a real-world table (bottom), with geometry and appearance fully decoupled and editable.}
    \label{fig:pipeline}
\end{figure*}

\section{Introduction}
\label{sec:intro}

Point-based radiance field representations, most notably 3D Gaussian Splatting (3DGS)~\cite{Kerbl20233DGS}, have emerged as a powerful approach to novel view synthesis and 3D reconstruction. These methods represent scenes as point clouds, where each point is associated with a volumetric primitive (\eg a 3D Gaussian distribution) that has an opacity and a single view-dependent color. Rasterizing these volumetric primitives is efficient, so techniques such as 3DGS can quickly optimize large collections of primitives to represent detailed scenes. 

However, since each Gaussian is assigned a single (view-dependent) color, 3DGS can only represent detailed appearance and textures by using many (often millions) of Gaussian primitives. This coupling of geometry and appearance resolution prevents scenes reconstructed by 3DGS from being easily edited. Editing appearance is difficult because the textures of reconstructed scenes cannot be modified at resolutions finer than the sizes of the reconstructed Gaussians, and editing geometry is difficult because it is not clear how to non-rigidly transform Gaussian primitives based on user manipulations.

Recent efforts to decouple appearance and geometry in Gaussian splatting have extended 3DGS to include either global~\cite{Xu2024TextureGSDT} or per-primitive~\cite{Rong2024GStexPT, Song2024HDGST2, Chao2024TexturedGF, Zhang2025NeuralST, Svitov2024BillBoardS, Tang2025TexGSVolVisES} UV maps. However, these techniques are still unable to enable both geometry deformations and appearance editing. Learned global UV maps from 3D surface points to 2D UV coordinates are no longer valid after geometry deformations, and per-primitive UV maps have difficulties handling deformations that are finer than the size of the Gaussian primitives.

We focus on another point-based representation that uses attention to render a point cloud~\cite{zhang2023papr,chang2023pointersect,ost2022nplf}, which effectively avoids those limitations by modeling scenes as a collection of infinitesimal points with learned features. These methods render rays with a cross-attention mechanism between the ray and the $K$ points closest to the ray. This defines the aggregated feature of a ray as the interpolation of the features from those $K$ nearest points, which enables editing scene geometry since these interpolation weights are still valid after deformations of the scene points. While attention-based point representation is better suited for geometry editing than 3DGS~\cite{zhang2023papr}, appearance editing is still challenging as it is not possible to locally edit textures by modifying the optimized point features. 

In this paper, we propose PointGT, a representation that extends the attention-based point renderer PAPR~\cite{zhang2023papr} to enable simultaneous editing of geometry and appearance. Our key contributions are:
\begin{itemize}
 \item We extend PAPR to include a learned UV mapping from 3D points on objects to a learned 2D texture map. (Sec.~\ref{sec:learning-uv-mapping})
 \item We introduce two novel geometry regularizers that significantly improve the fidelity of PAPR's geometry representation and demonstrate that these are crucial for effective UV mapping. (Sec.~\ref{sec:pre-training-papr})
 \item We propose a deformation-aware correspondence mechanism that explicitly maps deformed-space ray intersections back to canonical space via displacement fusion, preserving texture attachment under non-rigid edits. (Sec.~\ref{sec:geometry-appearance-editing})
\end{itemize}

\section{Related Work}
\label{sec:related_work}

Editing the geometry and appearance of 3D assets is a core part of the computer graphics workflow for digital content creation. While there are many established techniques and tools for editing widely-used mesh-based representations, editing the 3D representations produced by recent view synthesis and 3D reconstruction methods is much more challenging. 

The geometry of neural scene representations such as Neural Radiance Fields (NeRFs)~\cite{mildenhall2021nerf} cannot be directly edited, so many works~\cite{yuan2022nerfedit,xu2022cagedeforming,Peng2022CageNeRFCN,Jambon2023NeRFshop,Wang2023MeshGuidedNI,Zhou2023SERFFI,Yang2022NeuMeshLD,Zheng2022EditableNeRFET} have focused on associating NeRFs with proxy geometry which can be edited, and then converting this edited proxy geometry back into the NeRF representation. However, converting NeRFs to proxy geometry and back can introduce approximation errors, and the choice of proxy geometry can limit the types of edits that are possible.

On the other hand, the appearance of reconstructed NeRFs can be edited by optimizing UV mapping functions that map from 3D coordinates on the object to 2D texture coordinates~\cite{Xiang2021NeuTexNT,Srinivasan2023NuvoNU}. In this work, we use ideas from Nuvo~\cite{Srinivasan2023NuvoNU} to develop a UV mapping technique for point-based reconstruction techniques.

\subsection{3D Shape Deformation and Correspondence}
\label{sec:related-work-deformation}

Classical geometry processing has a rich history of editing 3D shapes, leveraging methods like As-Rigid-As-Possible (ARAP) deformation~\cite{sorkine2007rigid}, Moving Least Squares (MLS), and embedded deformation graphs~\cite{Sumner2007EmbeddedDF} to manipulate surface meshes or cages flexibly. While these approaches effectively manipulate geometry and preserve local structural rigidity, applying appearance edits to the deformed surfaces is relatively straightforward for meshes because texture coordinates (UVs) are explicitly attached to vertices and interpolate naturally across triangles.

However, projecting and interpolating high-resolution 2D textures onto unstructured, dynamically-weighted point clouds is fundamentally difficult. As point-based neural representations (like PAPR or 3DGS) lack explicit connectivity, the set of neighbors and attention weights used to define a surface point can shift non-rigidly under deformation. PointGT directly targets this novel problem of establishing canonical correspondence within dynamic attention fields to enable robust texture persistence.

More recent view synthesis techniques are based on 3D Gaussian Splatting (3DGS), which is a point-based representation that optimizes parameters of a 3D Gaussian along with a view-dependent color (parameterized using spherical harmonics) for each point. 3DGS is more suitable for direct geometry editing, since it is straightforward to move the position, orientation, and scale of any particle in the representation. 

While Gaussian splatting has proved to be a powerful representation for numerous tasks, this paradigm has several limitations: (1) it often requires millions of Gaussian primitives to enable high rendering quality and accurate scene reconstructions, (2) it's challenging to make appearance edits that are finer than the size of an individual Gaussian, and (3) free-form geometry deformation remains inconvenient without introducing geometry proxies. 

Recent works tackle the first two issues by decoupling the appearance and geometry of Gaussian splatting. Texture-GS~\cite{Xu2024TextureGSDT} learns a global UV mapping for all the ray-Gaussian intersections and stores features of these intersections on a 2D texture map. Instead of learning a global UV mapping with surface parameterization constraints (e.g., simple topologies), another line of works~\cite{Rong2024GStexPT, Song2024HDGST2, Chao2024TexturedGF, Zhang2025NeuralST, Svitov2024BillBoardS, Tang2025TexGSVolVisES} learn a per-primitive texture map for each Gaussian, where the Gaussians are often restricted to be 2D surfels~\cite{Huang20242DGS} so texture for each point on the surfel can easily be queried from the texture map with their 2D UV coordinates. Since the frequency of surface textures is often much higher than that of geometry, decoupling the resolutions of geometry and texture representations allows these methods to render scenes photo-realistically with only 5K points~\cite{Zhang2025NeuralST, Xu2024TextureGSDT, Rong2024GStexPT, Chao2024TexturedGF, Svitov2024BillBoardS} and enables manipulation of the textures with fine-grained details~\cite{Rong2024GStexPT, Xu2024TextureGSDT, Tang2025TexGSVolVisES}. 

However, none of these methods can perform simultaneous geometry deformation and texture editing. This limitation is rooted in the fundamental representation of Gaussian splatting, where each point is represented as a volume of space. In the case where per-primitive texture maps are learned, preserving textures after deformation requires not only the correct transformation (rotation, scaling) of each Gaussian but also accurate splitting or cloning of primitives to represent deformations that are finer than a Gaussian (as Gaussians are rigid and cannot be bent). Such a process is often intractable in practice. In the case where a global UV mapping is learned, in addition to the difficulties described above, the UV mapping must be correctly found for the new intersection points after deformation. The combination of these factors makes the simultaneous editing of texture and geometry for Gaussian splatting extremely difficult.

\section{Preliminaries}
\label{sec:preliminaries}

We briefly introduce the relevant background on attention-based point representations and neural UV mappings, and establish the notation used throughout.

\subsection{Attention-Based Point Renderers}
\label{sec:prelim-papr}

Attention-based point renderers (like PAPR~\cite{zhang2023papr} and Pointersect~\cite{chang2023pointersect}) represent scenes as unstructured point clouds coupled with an attention interpolation mechanism. For instance, PAPR represents a scene as $N$ learnable points $\mathcal{P} = \{(\mathbf{p}_j,\, \mathbf{f}_j)\}_{j=1}^{N}$, where $\mathbf{p}_j \in \mathbb{R}^3$ is the spatial position and $\mathbf{f}_j \in \mathbb{R}^h$ is a feature vector. To render ray $\mathbf{r}_i$, PAPR selects the $K$ points nearest to it by perpendicular distance as the local neighborhood. A learned cross-attention mechanism then assigns a softmax weight $a_{ij}$ to each neighbor point $j$, and the ray's feature $\mathbf{f}^{\text{ray}}_i$ is computed as the weighted sum of their features:
\begin{align}
    \mathbf{f}^{\text{ray}}_i = \sum_{j=1}^K a_{ij}\, \mathbf{f}_{ij}, \label{eq:aggregate-feature}
\end{align}
where $\mathbf{f}_{ij}$ denotes the feature associated with the $j$-th selected neighbor point $\mathbf{p}_{ij}$ for ray $i$. The aggregated feature map is then decoded by a UNet~\cite{Ronneberger2015UNetCN} to predict the RGB colors. Given the same attention weights, these methods typically predict the intersection point $\mathbf{x}_i \in \mathbb{R}^3$ where the ray hits the surface by interpolating the point positions:
\begin{align}
    \mathbf{x}_i = \sum_{j=1}^K a_{ij}\, \mathbf{p}_{ij}. \label{eq:aggregate-surface-point}
\end{align}
Note that $\mathbf{x}_i$ is an \emph{attention-weighted average} of point positions, not a true geometric intersection: its quality depends on how concentrated the weights are and how well-placed the supporting points are. This sensitivity motivates the geometry regularizers introduced in Sec.~\ref{sec:pre-training-papr}.

\paragraph{Canonical and deformed space.}
\label{sec:prelim-canonical}
We call the point positions after training the \textbf{canonical positions} $\mathbf{p}_j^{\mathrm{can}}$. The corresponding \textbf{canonical intersection}
\begin{align}
    \mathbf{x}_i^{\mathrm{can}} = \sum_{j=1}^K a_{ij}^{\mathrm{can}}\, \mathbf{p}_{ij}^{\mathrm{can}}
\end{align}
is what the UV mapping is trained on. At edit time, a non-rigid deformation displaces each point to $\mathbf{p}_j^{\mathrm{def}}$, producing a \textbf{deformed intersection}
\begin{align}
    \mathbf{x}_i^{\mathrm{def}} = \sum_{j=1}^K a_{ij}^{\mathrm{def}}\, \mathbf{p}_{ij}^{\mathrm{def}}
\end{align}
in deformed space, where direct UV lookup is no longer valid. Bridging this gap is the central problem addressed in Sec.~\ref{sec:geometry-appearance-editing}.

\subsection{Neural UV Mapping}
\label{sec:prelim-nuvo}

Neural UV mapping methods (like Nuvo~\cite{Srinivasan2023NuvoNU} and NeuTex~\cite{Xiang2021NeuTexNT}) parameterize surface textures by learning continuous mappings from 3D space to 2D texture atlases. For example, Nuvo learns a multi-chart UV atlas and texture map from a set of surface points $\mathcal{G}$ (in our setting, $\mathcal{G}=\{\mathbf{x}_i^{\mathrm{can}}\}$). It uses cycle-consistency (bijectivity) losses and additional regularizers to learn the UV mappings; we refer readers to~\cite{Srinivasan2023NuvoNU} for full details.

To enable texture editing and to encourage an even allocation of texture resolution over the scene, Nuvo also penalizes mapping distortion using conformal and stretch regularizers that operate on the differential of each chart's $3$D$\!\to\!2$D UV mapping~\cite{Srinivasan2023NuvoNU}. These regularizers require the surface normal at each surface point to construct tangent-space directions, which are not explicitly available for raw attention-interpolated intersection points $\mathbf{x}_i^{\mathrm{can}}$. We therefore use a normal-free distortion loss in Sec.~\ref{sec:learning-uv-mapping} that removes the need for surface normals.

\begin{figure*}
    \centering
    \captionsetup{type=figure}
    \captionsetup[subfloat]{labelformat=empty}
    \centering
    \footnotesize
    \includegraphics[width=1\hsize,valign=m]{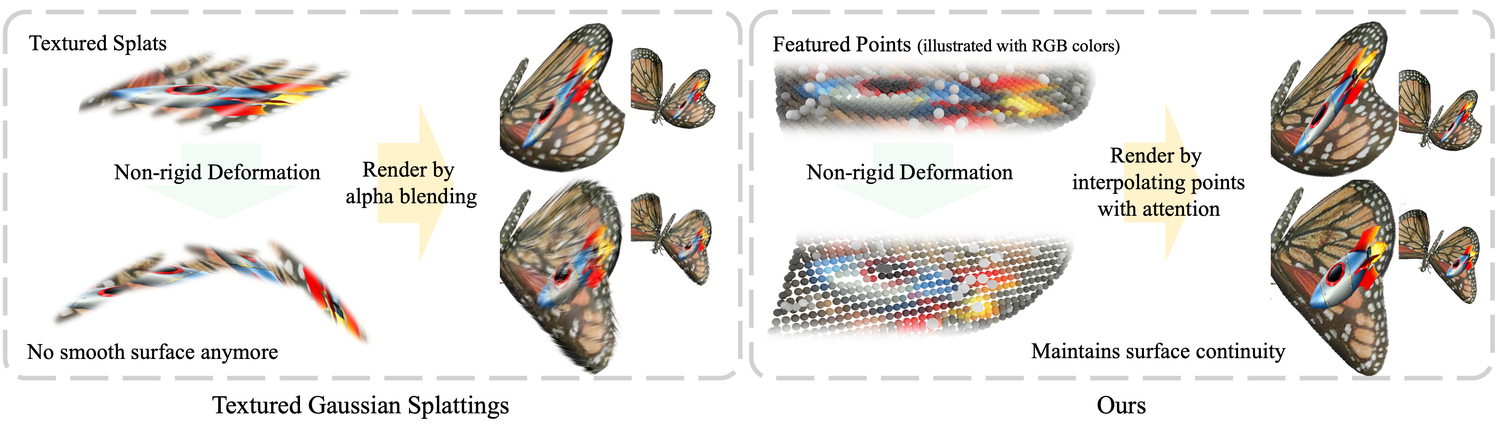}
    \captionof{figure}{Comparison of geometry deformations of textured Gaussian splatting methods and PointGT. \textbf{Left:} Since each Gaussian is a linear primitive, these methods struggle to preserve surface smoothness under large non-rigid deformations. \textbf{Right:} PointGT maintains surface continuity after deformation by predicting ray--surface points via attention-based interpolation on a point cloud.}
    \label{fig:pointgt-vs-gs}
\end{figure*}

\section{Method}
\label{sec:method}

The goal of PointGT is to enable persistent texture edits under non-rigid deformation for point-based neural representations. While applying texture edits to deformed surfaces is relatively straightforward for meshes---where texture coordinates (UVs) are explicitly attached to vertices and interpolate naturally across triangles---point clouds lack this explicit connectivity. Consequently, maintaining a consistent mapping between the deformed geometry and a 2D texture map becomes fundamentally difficult, often causing texture edits to drift or break when the underlying points move non-rigidly.

To tackle this challenge, the choice of the underlying point-based representation is critical. As discussed, Gaussian Splatting-based methods struggle to preserve surface continuity under non-rigid deformations (Fig.~\ref{fig:pointgt-vs-gs}, left), and may break continuous texture mapping. In contrast, attention-based renderers like PAPR~\cite{zhang2023papr} maintain surface continuity via smooth point interpolation (Fig.~\ref{fig:pointgt-vs-gs}, right), providing the stable surface points required for consistent UV mapping.

Building upon this insight, PointGT introduces a framework for simultaneous geometry and texture editing. To ensure the attention-based surface points are suitable for learning a multi-chart UV atlas and texture map~\cite{Srinivasan2023NuvoNU}, we first propose \textbf{geometry regularization} losses that condition the ray--surface intersections to accurately reflect the true geometry (Sec.~\ref{sec:pre-training-papr}). Then, to preserve texture attachment during edits, we introduce a \textbf{deformation-aware canonical correspondence} mechanism (Sec.~\ref{sec:geometry-appearance-editing}). This explicitly maps the deformed-space intersections reliably back to the original canonical space for consistent UV lookup, ensuring that texture edits persist seamlessly regardless of the applied geometric deformation.

\subsection{Learning PAPR with Geometry Regularization}
\label{sec:pre-training-papr}

To enable reliable UV learning and stable editing, we require accurate and well-conditioned ray--surface intersection points. In PAPR, the predicted surface point for ray $i$ is an attention-weighted average of $K$ ray-nearest neighbors (Eq.~\ref{eq:aggregate-surface-point}), which can be sensitive to sparse attention and imperfect point geometry. Without additional constraints, predicted intersection points may \emph{collapse} toward discrete support points and/or be supported by noisy off-surface neighborhoods (Fig.~\ref{fig:abla-sp-close-to-ray} and~\ref{fig:abla-sp-close-to-surface}), which harms the sampled surface point set used to train the UV atlas. We therefore add two geometry regularizers during PAPR pre-training.

\paragraph{On-ray regularizer (close-to-ray).}
As shown in Fig.~\ref{fig:abla-sp-close-to-ray}, sparse attention weights can cause predicted intersection points to cluster around discrete point supports. This degrades the effective geometric resolution of $\mathbf{x}_i$ to the resolution of the point cloud and yields unstable surface samples for UV supervision. Since a valid ray--surface intersection point should lie on its ray by definition, we minimize the distance between each predicted point $\mathbf{x}_i$ and its orthogonal projection $\mathbf{x}'_i$ onto the corresponding ray $\mathbf{r}_i$:
\begin{align}
    \mathcal{L}_{\text{close2ray}} = \frac{1}{M}\sum_{i=1}^M\|\mathbf{x}_i - \mathbf{x}'_i\|_2, \label{eq:on-ray-reg}
\end{align}
where $M$ is the number of rays in a batch.

\paragraph{On-surface neighborhood regularizer (close-to-surface).}
Even when intersections can be predicted, the underlying point cloud may contain off-surface points. This is problematic when starting from sparse points and densifying: new points may gather around an off-surface location, and a ray may then select a neighborhood whose convex hull does not cover the true surface. To regularize neighborhood geometry, we encourage each selected neighbor point $\mathbf{p}_{ij}$ to move toward the predicted surface point $\mathbf{x}_i$:
\begin{align}
    \mathcal{L}_{\text{close2surface}} = \frac{1}{MK}\sum_{i=1}^M \sum_{j=1}^{K}\|\mathbf{p}_{ij}-\operatorname{sg}(\mathbf{x}_i)\|_2
\end{align}
that encourages all the nearby points to move closer to the predicted intersection point. Note that we stop the gradients backpropagating to $\mathbf{x}_i$ from this regularizer for training stability. To prevent early geometry collapse, we apply $\mathcal{L}_{\text{close2surface}}$ in the middle of training after a coarse geometry is learned.

\paragraph{Objective.}
We combine the regularizers with PAPR's rendering loss for pre-training:
\begin{align}
    \mathcal{L}_{\text{rendering}} + \gamma \ \mathcal{L}_{\text{close2ray}} + \eta \ \mathcal{L}_{\text{close2surface}},
\end{align}
where $\mathcal{L}_{\text{rendering}}$ is PAPR's rendering loss, $\gamma$ and $\eta$ are the coefficients of the regularizers.

\subsection{Learning UV Mapping}
\label{sec:learning-uv-mapping}

Building upon the regularized point cloud geometry, we extract ray--surface intersections to learn a multi-chart UV atlas and a 2D texture map. We sample foreground rays from the training views, compute their canonical-space intersection points $\mathbf{x}^{\mathrm{can}}_i$, and optimize a continuous neural UV parameterization over these surface samples.

\paragraph{Distortion loss without normals (Jacobian regularization).}

To enable consistent texture editing and encourage an even allocation of texture resolution across the scene, it is crucial to minimize the distortion of the learned UV mappings. While existing surface parameterization methods~\cite{Srinivasan2023NuvoNU} typically rely on ground-truth geometry normals to compute conformal and stretch penalties, point-based representations natively lack this explicit geometric connectivity and normal information. To overcome this, we define a distortion loss directly over the Jacobians of the texture coordinate mappings.

Specifically, for each chart $k$ in our multi-chart texture maps, we compute the Jacobian matrix $\mathbf{J}_{ik} \in \mathbb{R}^{2\times3}$ of the UV mapping by taking the gradients of the texture coordinates $\mathbf{u}_{ik} \in \mathbb{R}^2$ w.r.t. the 3D point coordinates $\mathbf{x}_i$. The Jacobian's singular values $\sigma_{ik}^1$ and $\sigma_{ik}^2$ characterize the local scaling behavior of the mapping. To minimize anisotropic scaling distortion, we minimize the squared difference of the singular values across all charts:
\begin{align}
    \mathcal{L}_{\text{scaling}} = \frac{1}{Gn}\sum_{i=1}^G \sum_{k=1}^n \big(\sigma_{ik}^1 - \sigma_{ik}^2\big)^2,
\end{align}
where $G$ is the total number of 3D points. This encourages the texture coordinate network to learn mappings with isotropic scaling (where $\sigma^1 = \sigma^2$). While $\mathcal{L}_{\text{scaling}}$ minimizes anisotropy, it does not explicitly prevent area shrinkage or blow-up. To address this, we also define an area distortion penalty, $\mathcal{L}_{\text{area}} = \frac{1}{Gn}\sum_{i=1}^G \sum_{k=1}^n (\log(\sigma_{ik}^1 \sigma_{ik}^2))^2$, which regularizes the local scale to remain close to $1$ and heavily penalizes near-zero areas, preventing degenerate mappings. We define the distortion loss as the sum of these two terms:
\begin{align}
    \mathcal{L}_{\text{distortion}} = \mathcal{L}_{\text{scaling}} + \mathcal{L}_{\text{area}}.
\end{align}

In practice, we optimize the UV atlas by minimizing a weighted sum of Nuvo~\cite{Srinivasan2023NuvoNU}'s losses (excluding the normal-based distortion terms) and the proposed distortion loss $\mathcal{L}_{\text{distortion}}$. 
We jointly learn a 2D RGB texture map by supervising rays with PAPR-predicted colors using bilinear sampling on the atlas; we use the learned texture map for texture editing.

\subsection{Geometry and Texture Editing}
\label{sec:geometry-appearance-editing}

\paragraph{Texture editing in UV space.}
PointGT supports texture editing by directly modifying the 2D texture atlas. Given an edited texture map $\mathbf{T}'$, we generate a mask map $\mathbf{M} \in [0,1]^{n\times h \times w}$ by comparing the original atlas $\mathbf{T}$ and the edited atlas $\mathbf{T}'$, where the pixel values are marked as $1$ in edited regions and $0$ otherwise. During rendering, for each ray we compute the UV coordinate $\mathbf{u}_i$ of its intersection and replace the sampled color with the edited texture when $m_i = \mathbf{M}(\mathbf{u}_i) > 0.95$ (both using bilinear interpolation on the atlas).

\paragraph{Geometry editing and the correspondence problem.}
Geometry editing applies a non-rigid deformation to the learned point cloud, producing deformed point positions $\mathbf{p}^{\mathrm{def}}_j$ from canonical positions $\mathbf{p}^{\mathrm{can}}_j$ (Sec.~\ref{sec:prelim-canonical}). Rendering after deformation produces a deformed-space ray--surface point by interpolating deformed neighbor positions with deformed-space attention weights:
\begin{align}
    \hat{\mathbf{x}}^{\mathrm{def}}_i = \sum_{j=1}^K a^{\mathrm{def}}_{ij}\, \mathbf{p}^{\mathrm{def}}_{ij},
    \label{eq:aggregate-surface-point-deformed-hat}
\end{align}
where the top-$K$ neighbors $\mathbf{p}^{\mathrm{def}}_{ij}$ are selected by ray proximity in the deformed point cloud, and the weights $a^{\mathrm{def}}_{ij}$ are \emph{re-inferred} in deformed space (i.e., both the neighbor set and weights are recomputed after deformation). However, the UV atlas is trained in canonical space, so we must transfer $\hat{\mathbf{x}}^{\mathrm{def}}_i$ back to a canonical coordinate before UV lookup.

\paragraph{Why a naïve transfer fails.}
A natural approach is to reuse the deformed attention weights to directly interpolate \emph{canonical} neighbor positions:
\begin{align}
    \mathbf{x}^{\mathrm{can}}_{i,\text{naive}} = \sum_{j=1}^K a^{\mathrm{def}}_{ij}\, \mathbf{p}^{\mathrm{can}}_{ij}.
    \label{eq:naive-canonical-sum}
\end{align}
In practice, attention weights in PAPR are often sparse; Eq.~\ref{eq:naive-canonical-sum} therefore collapses many rays' canonical lookup points onto a small subset of support points, causing discontinuous UV jumps across neighboring rays and visible texture popping/drift under deformation. While the on-ray regularizer (Sec.~\ref{sec:pre-training-papr}) improves training stability, it does not guarantee that intersections remain exactly on rays under extreme non-rigid, non-training deformations. 

\begin{figure*}
    \centering
    \includegraphics[width=1\linewidth]{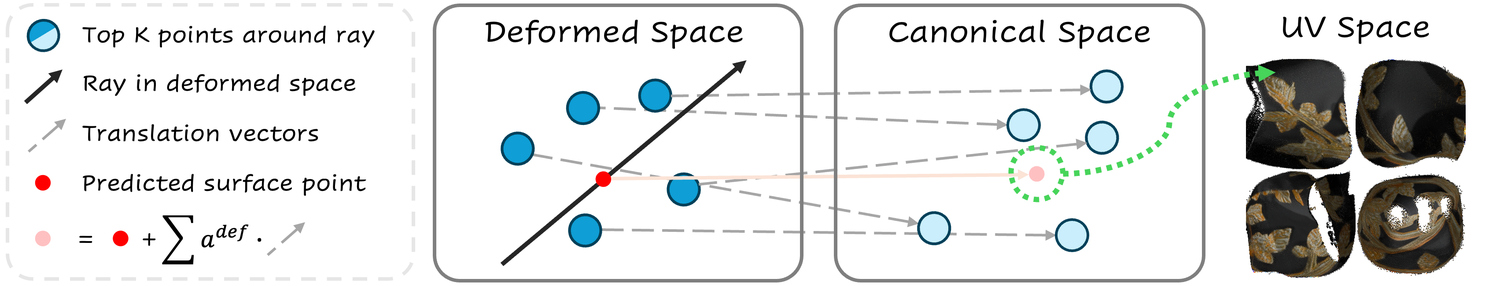}
    \caption{Illustration of the deformation-aware canonical correspondence. In deformed space (left), a ray intersects the deformed point cloud to predict a surface point (red). Each deformed neighbor (dark blue) has a stored canonical counterpart (light blue); the per-neighbor displacements (dashed gray arrow) are fused with the deformed-space attention weights to transfer the intersection back to canonical space (middle), where UV lookup retrieves the texture (right).}
    \label{fig:canonical-correspondence}
\end{figure*}

\paragraph{Deformation-aware canonical correspondence.}
To address this, we propose an edit-time transfer rule that maps each deformed-space ray--surface point back to canonical space while keeping samples well-distributed.
\textbf{Step 1 (point-to-ray projection).} We strictly enforce the on-ray constraint by orthogonally projecting $\hat{\mathbf{x}}^{\mathrm{def}}_i$ onto its ray $\mathbf{r}_i$:
\begin{align}
    \mathbf{x}^{\mathrm{def}}_i = \Pi_{\mathbf{r}_i}\big(\hat{\mathbf{x}}^{\mathrm{def}}_i\big),
    \label{eq:ray-projection}
\end{align}
\textbf{Step 2 (attention-weighted displacement fusion).} Since point identity is preserved under deformation, each deformed neighbor $\mathbf{p}^{\mathrm{def}}_{ij}$ has a corresponding canonical position $\mathbf{p}^{\mathrm{can}}_{ij}$. We compute per-neighbor displacements and fuse them using the \emph{same} deformed-space attention weights, then transfer the ray-projected deformed intersection back to canonical space:
\begin{align}
    \mathbf{t}_i = \sum_{j=1}^K a^{\mathrm{def}}_{ij}\, \big(\mathbf{p}^{\mathrm{can}}_{ij}-\mathbf{p}^{\mathrm{def}}_{ij}\big), \ \ \
    \mathbf{x}^{\mathrm{can}}_i = \mathbf{x}^{\mathrm{def}}_i + \mathbf{t}_i.
    \label{eq:deformation-aware-correspondence}
\end{align}
We then query the UV atlas at $\mathbf{x}^{\mathrm{can}}_i$ and sample the (edited) texture map to render appearance. Intuitively, even if attention weights are sparse, ray projection keeps $\mathbf{x}^{\mathrm{def}}_i$ distributed along rays, and the fused displacement provides a stable canonical transfer for UV lookup.

\paragraph{Edit-order interchangeability.}
PointGT supports flexible and simultaneous geometry and texture editing; the order of applying geometry and texture edits is interchangeable, which is important for applications such as 3D character design where edits are applied iteratively in varying order.

\section{Experiments}
\label{sec:experiments}

We evaluate PointGT along two axes: (i) \textbf{simultaneous geometry and texture editing} under \emph{non-rigid deformation}, where texture edits should persist without drifting (Sec.~\ref{sec:geometry-appearance-editing}), and (ii) \textbf{novel view synthesis} quality compared to recent textured Gaussian splatting baselines. 

\subsection{Implementation Details}

When pre-training PAPR with geometry regularization (Sec.~\ref{sec:pre-training-papr}), we use $\gamma = 0.002$ and $\eta = 0.01$ for the close-to-ray and close-to-surface losses. Since geometry is unreliable early in training, we start applying the regularizers at step $25{,}000$ (i.e., after $1/10$ of total steps). When learning the UV atlas and 2D texture map (Sec.~\ref{sec:learning-uv-mapping}), we follow Nuvo's optimization settings and use weight $0.4$ for the proposed $\mathcal{L}_{\text{distortion}}$ and $0.04$ for $\mathcal{L}_{\text{texture}}$. We use $n=4$ or $8$ charts depending on the complexity of the geometry, and fix memory usage by using a texture resolution of $256\sqrt{2/n} \times 256\sqrt{2/n}$ for each chart. For editing evaluation, we use Objaverse assets with artist-created motion sequences to evaluate persistent editing under realistic deformation scenarios (Sec.~\ref{sec:geometry-appearance-editing}).

\subsection{Simultaneous Geometry and Texture Editing}

PointGT supports texture editing by modifying the 2D texture atlas, and geometry editing by applying non-rigid deformations to the point cloud while preserving canonical UV lookup via our edit-time canonical correspondence (Sec.~\ref{sec:geometry-appearance-editing}). We evaluate this capability under non-rigid deformations and both \textbf{local} edits (e.g., paste an icon into a localized region) and \textbf{global} edits (e.g., recolor or material-style changes).

\begin{figure*}[t]
\centering
\setlength\tabcolsep{1pt}
\renewcommand{\arraystretch}{1.15}
\tiny
\begin{tabular}{l@{\hskip 0.2em}p{2.64cm}p{1.75cm}p{1.75cm}p{2.62cm}p{2.62cm}}
    & \quad\shortstack[c]{Original Scene and\\Edited Texture Map} & \shortstack[c]{Texture Edited \\ (Ours)} & \ \shortstack[c]{UV under \\ Deformation} & \shortstack[c]{Texture + Geometry \\ Edited (Ours)} & \shortstack[c]{Texture + Geometry \\ Edited (GSTex)} \\
\rotatebox[origin=c]{90}{emperor-fish}
    & \includegraphics[height=1.75cm,valign=m]{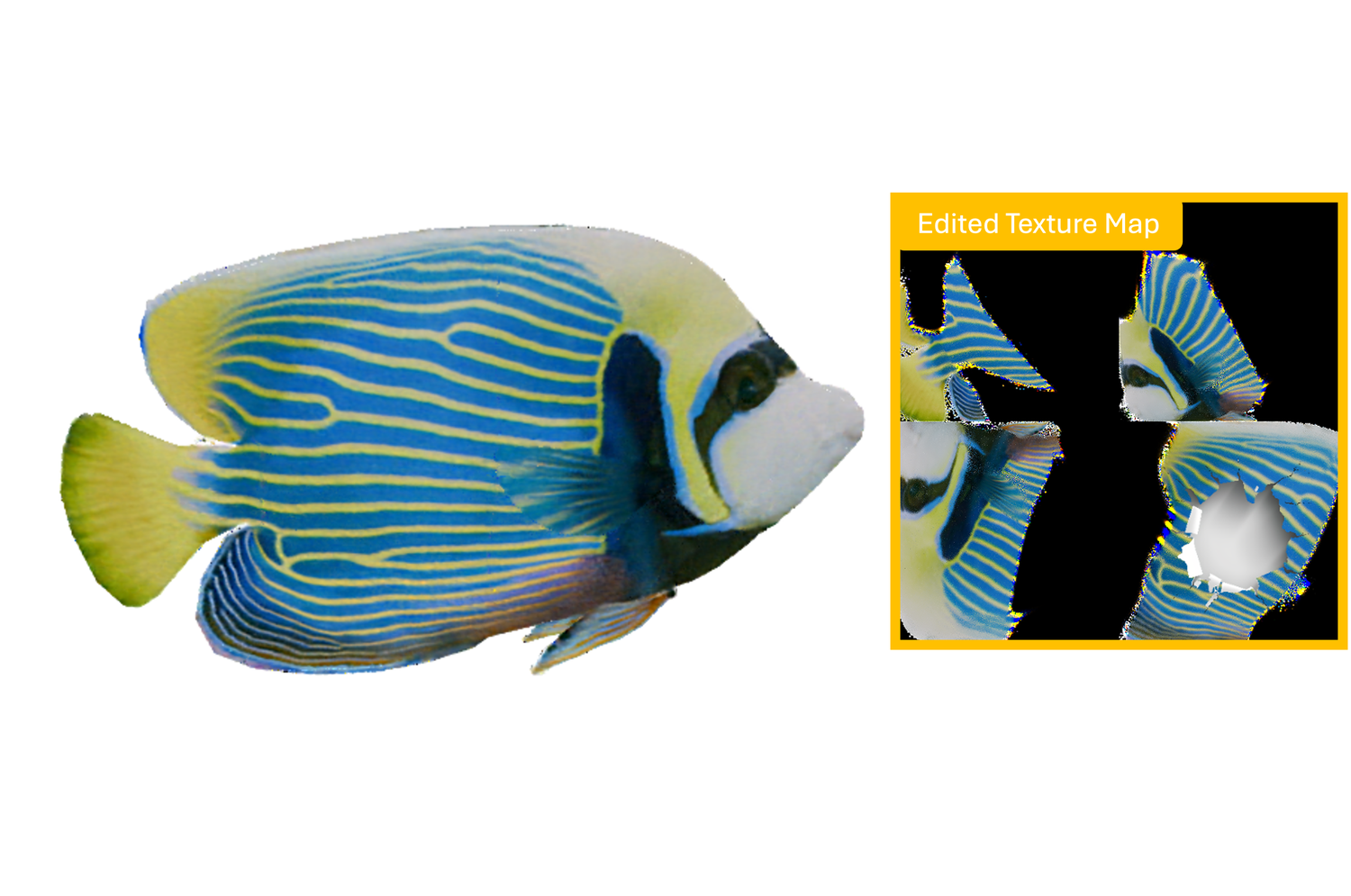}
    & \includegraphics[height=1.75cm,valign=m]{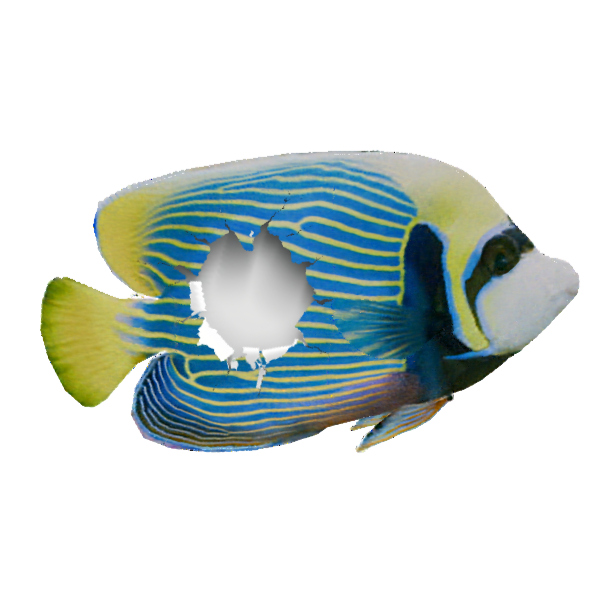}
    & \includegraphics[height=1.75cm,valign=m]{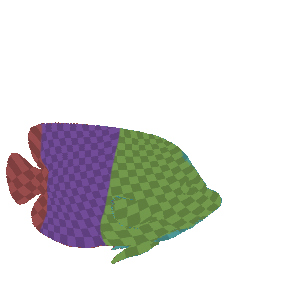}
    & \includegraphics[height=1.75cm,valign=m]{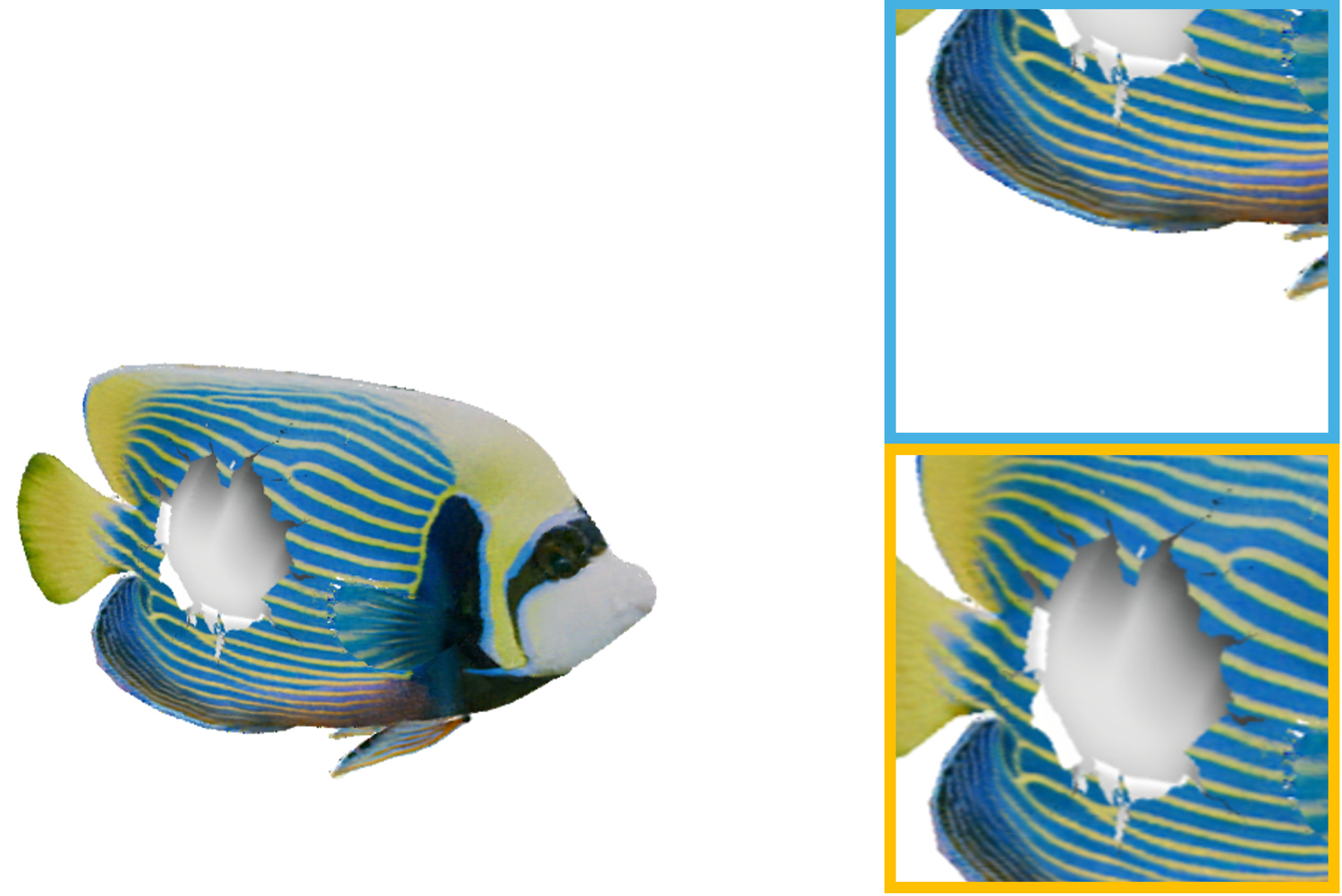}
    & \includegraphics[height=1.75cm,valign=m]{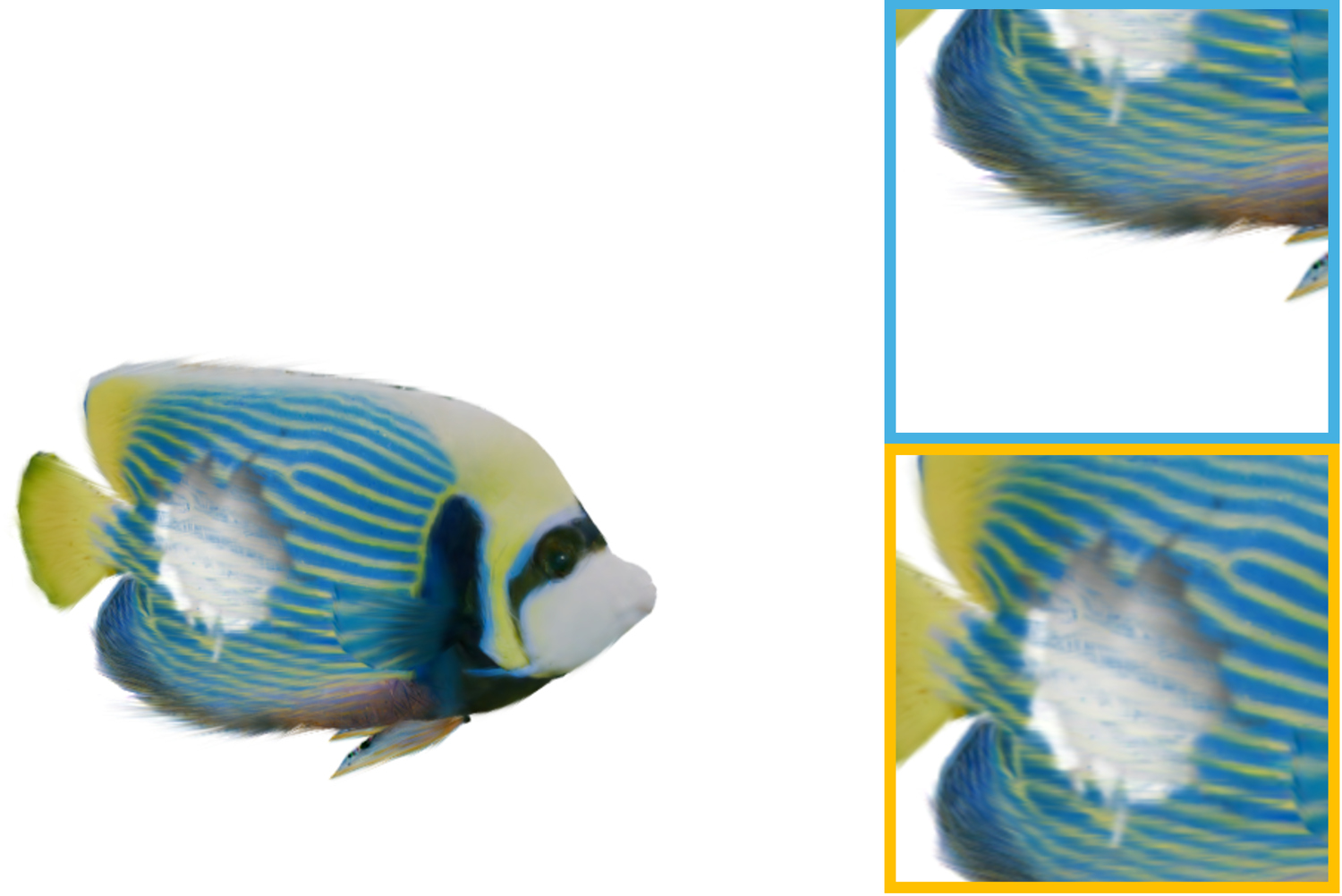} \\
\rotatebox[origin=c]{90}{blossom}
    & \includegraphics[height=1.75cm,valign=m]{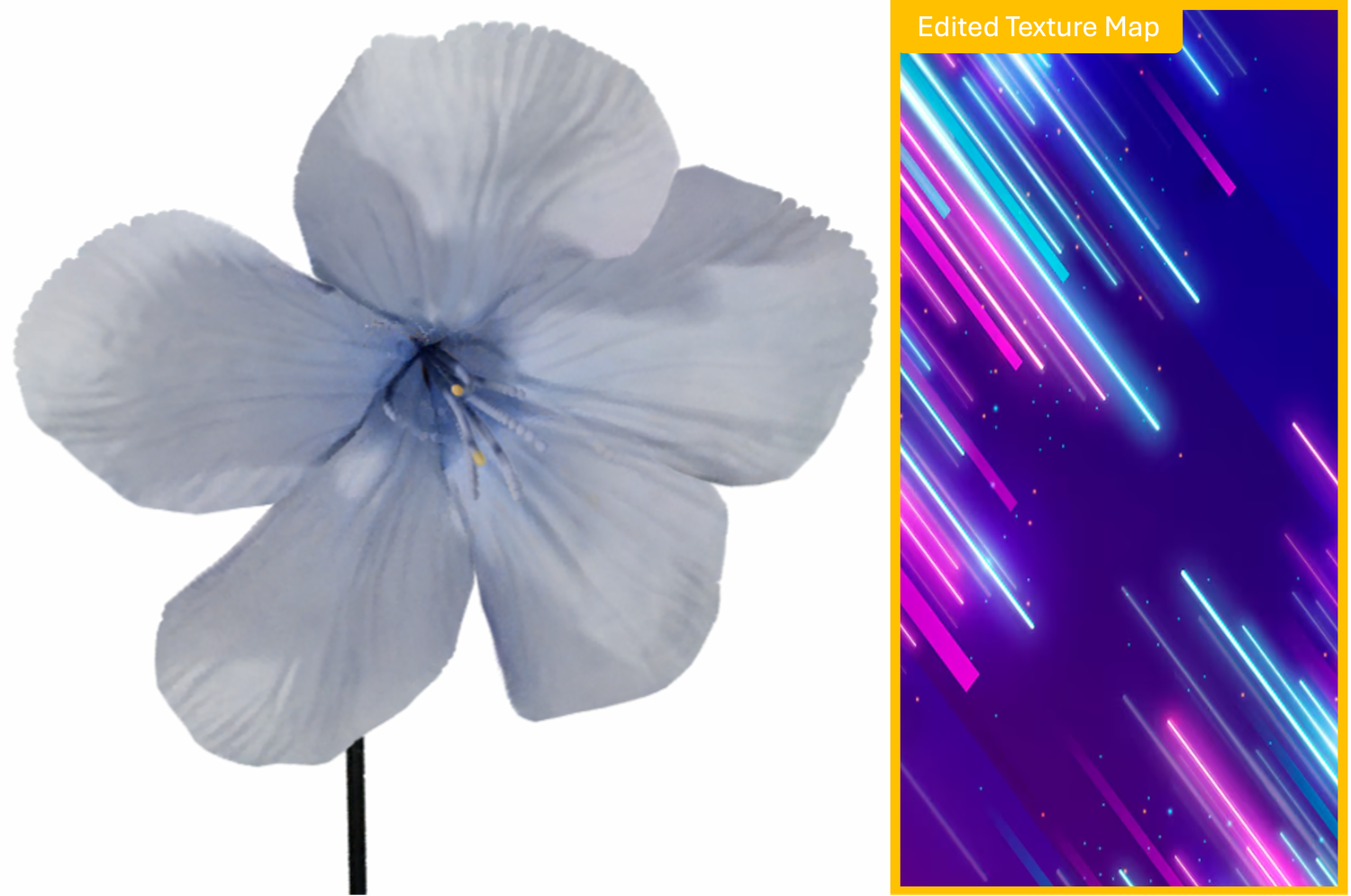}
    & \includegraphics[height=1.75cm,valign=m]{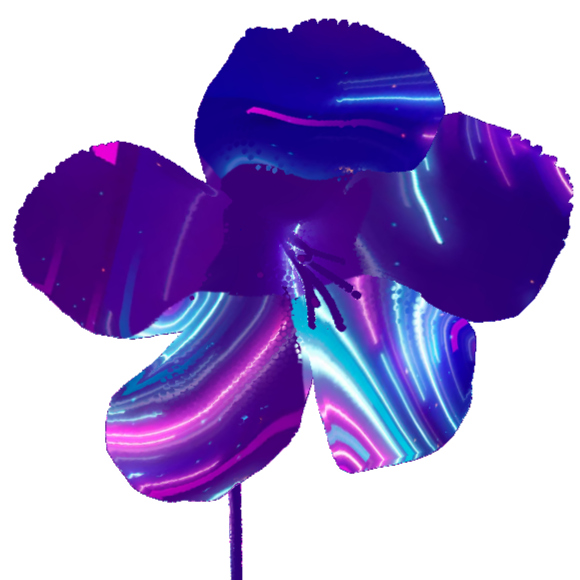}
    & \includegraphics[height=1.75cm,valign=m]{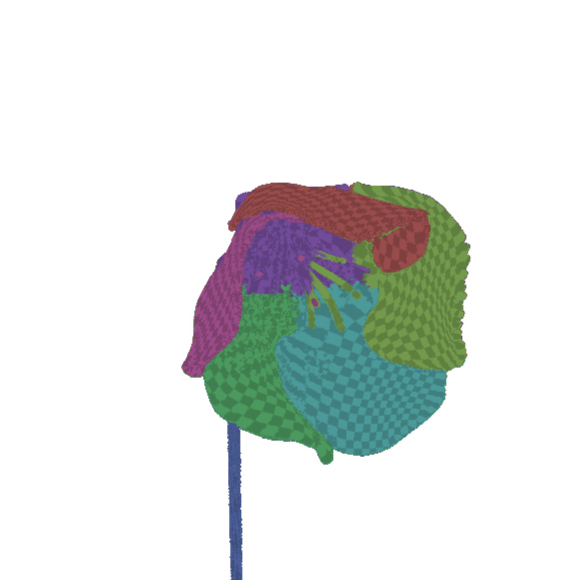}
    & \includegraphics[height=1.75cm,valign=m]{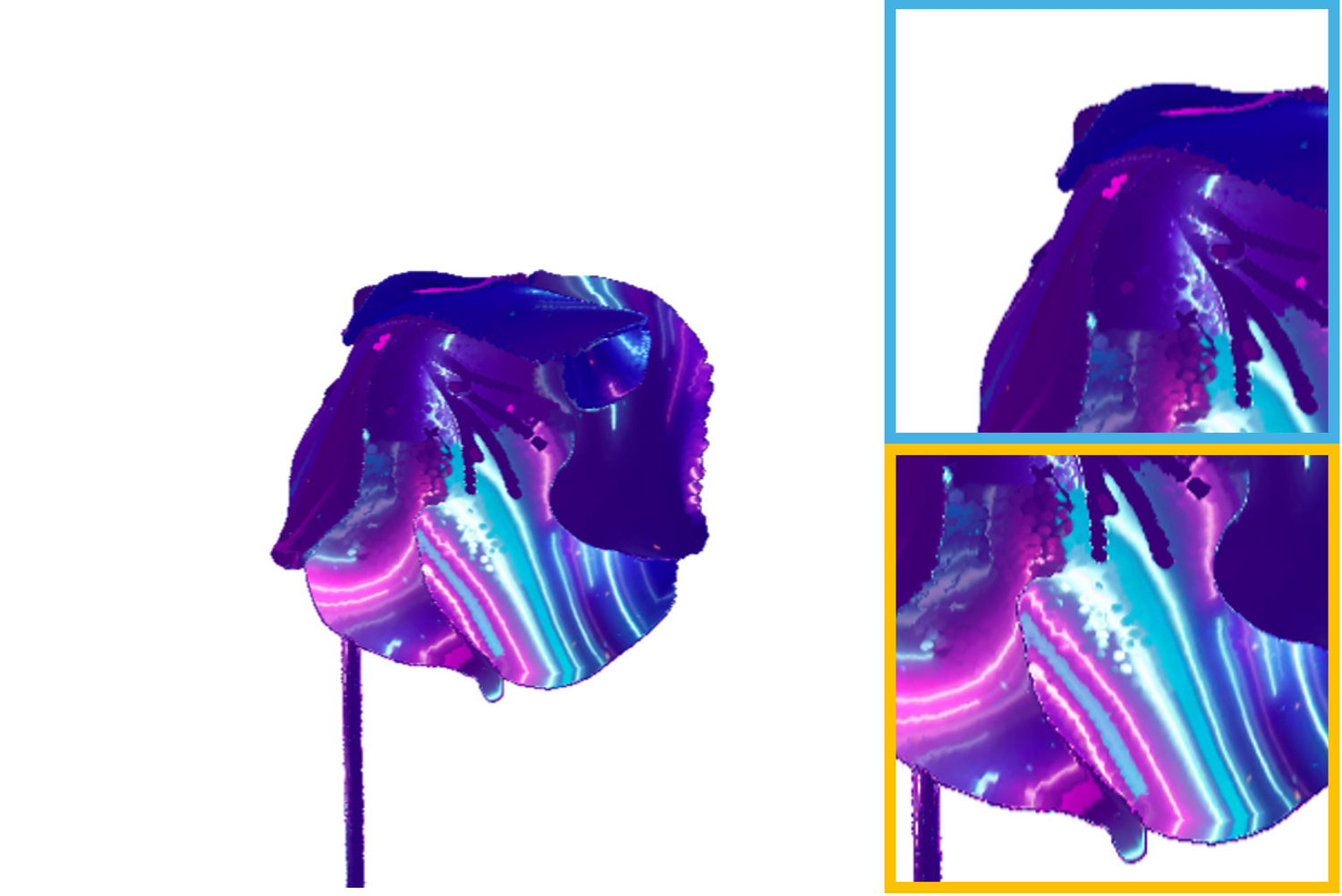}
    & \includegraphics[height=1.75cm,valign=m]{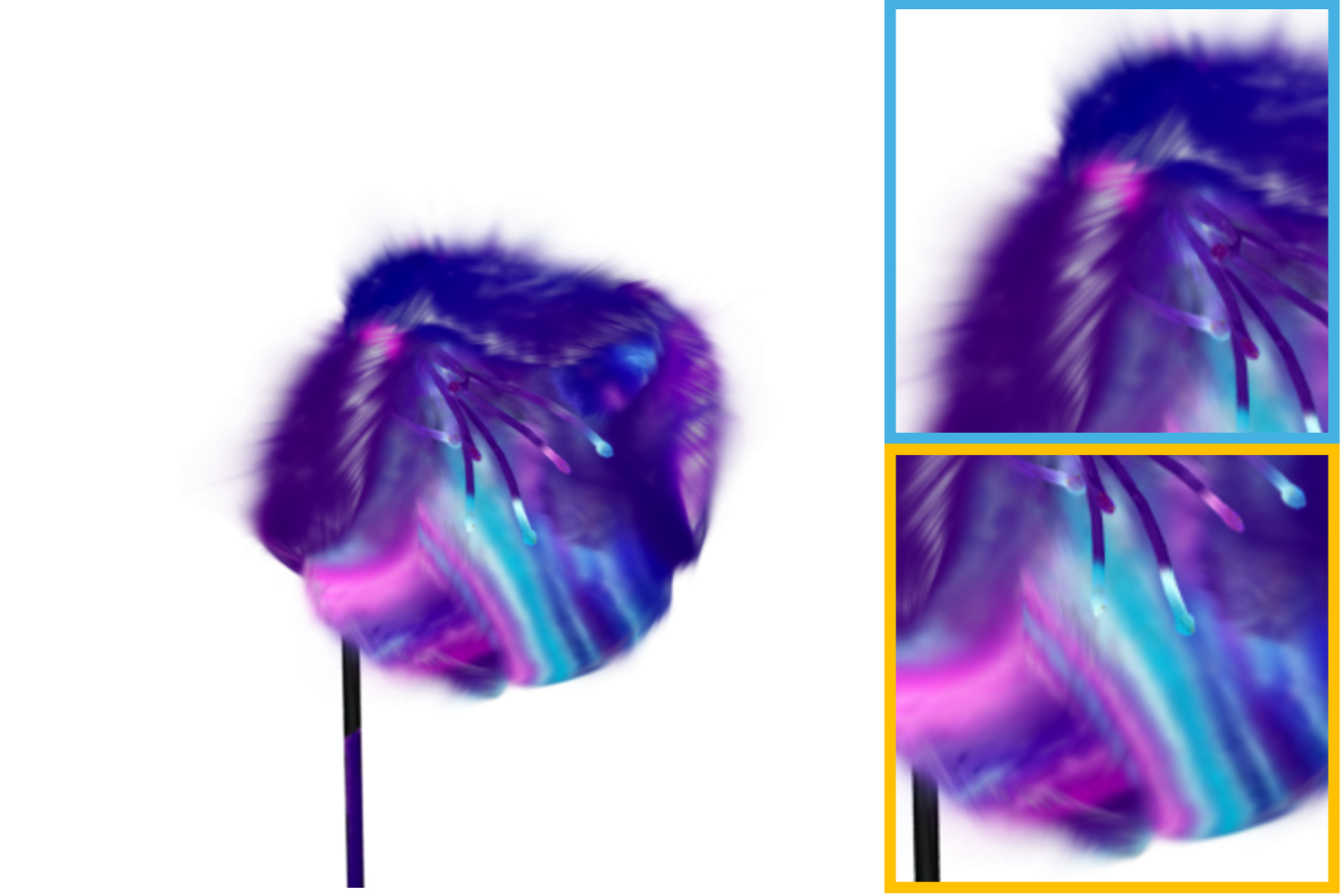} \\
\rotatebox[origin=c]{90}{dress}
    & \includegraphics[height=1.75cm,valign=m]{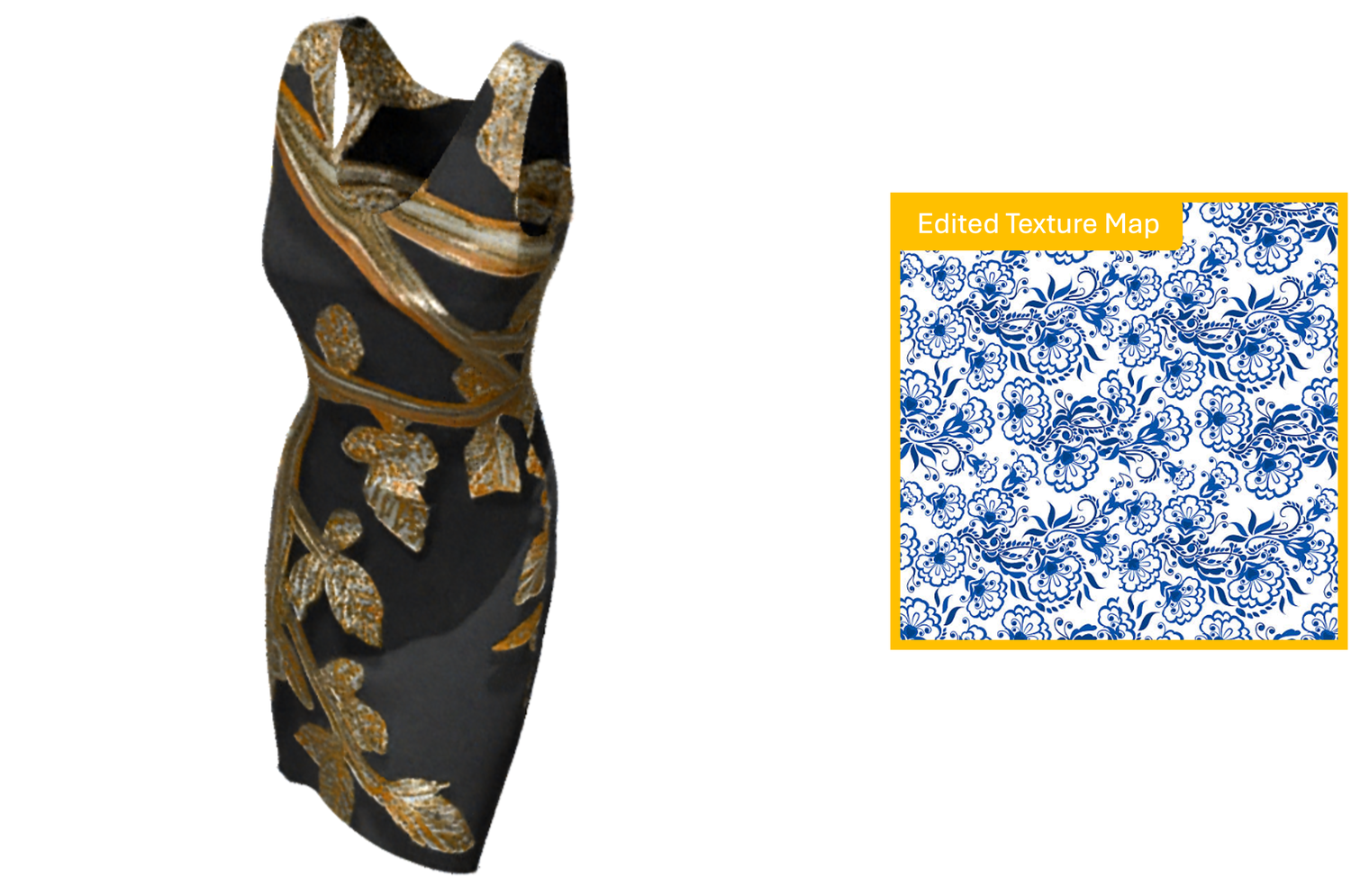}
    & \includegraphics[height=1.75cm,valign=m]{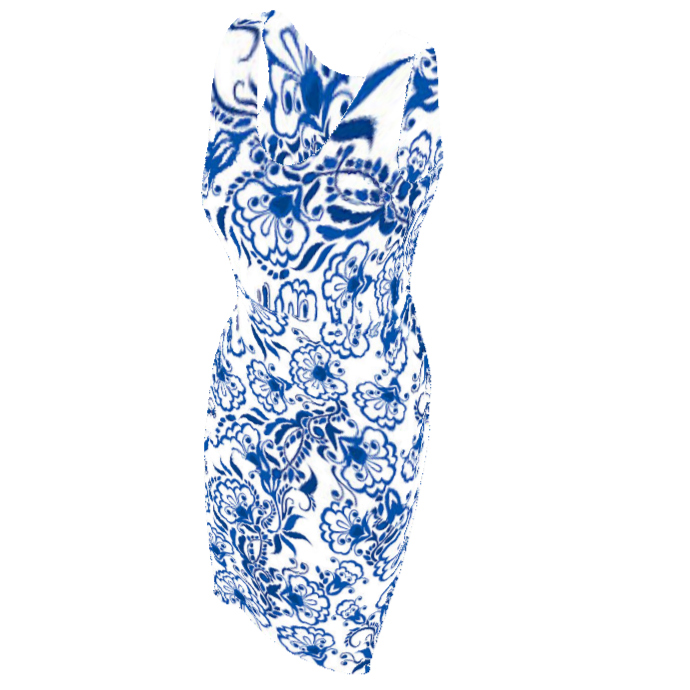}
    & \includegraphics[height=1.75cm,valign=m]{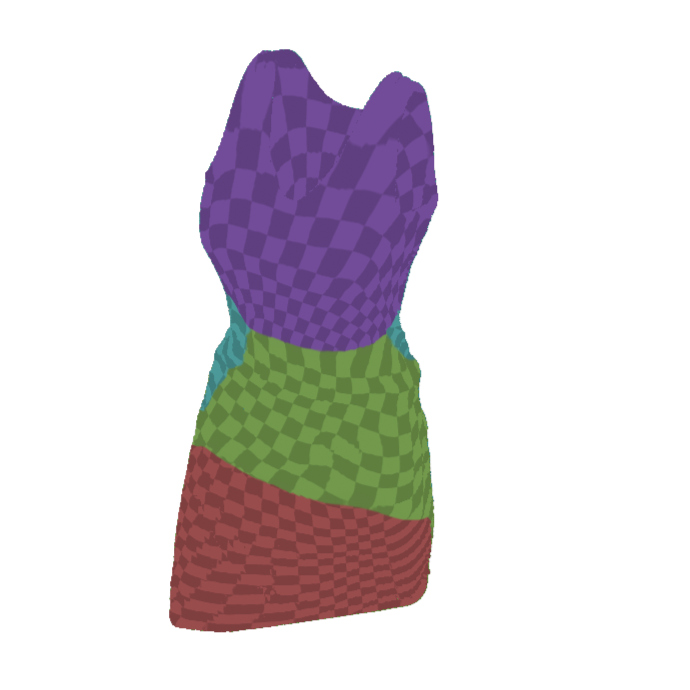}
    & \includegraphics[height=1.75cm,valign=m]{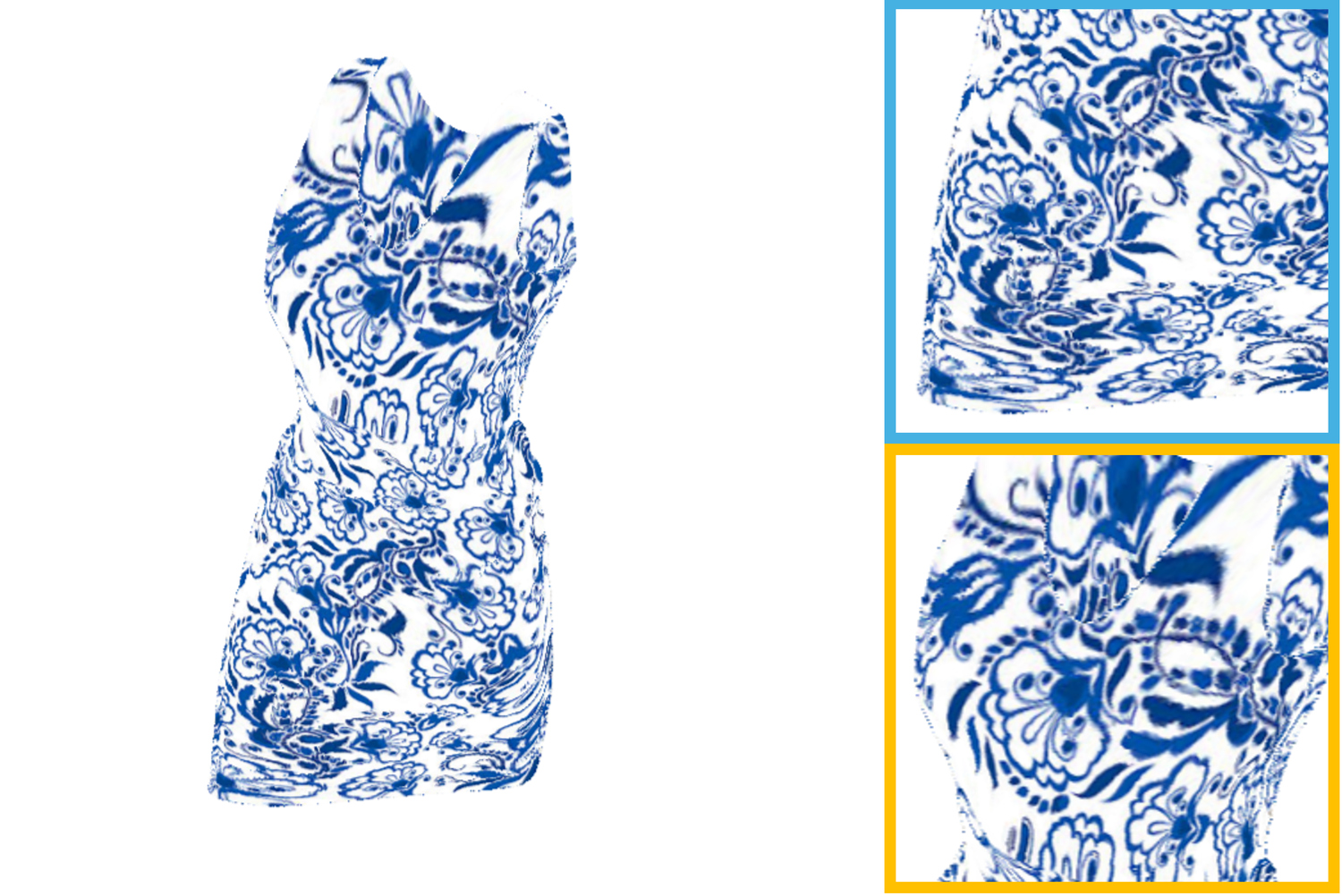}
    & \includegraphics[height=1.75cm,valign=m]{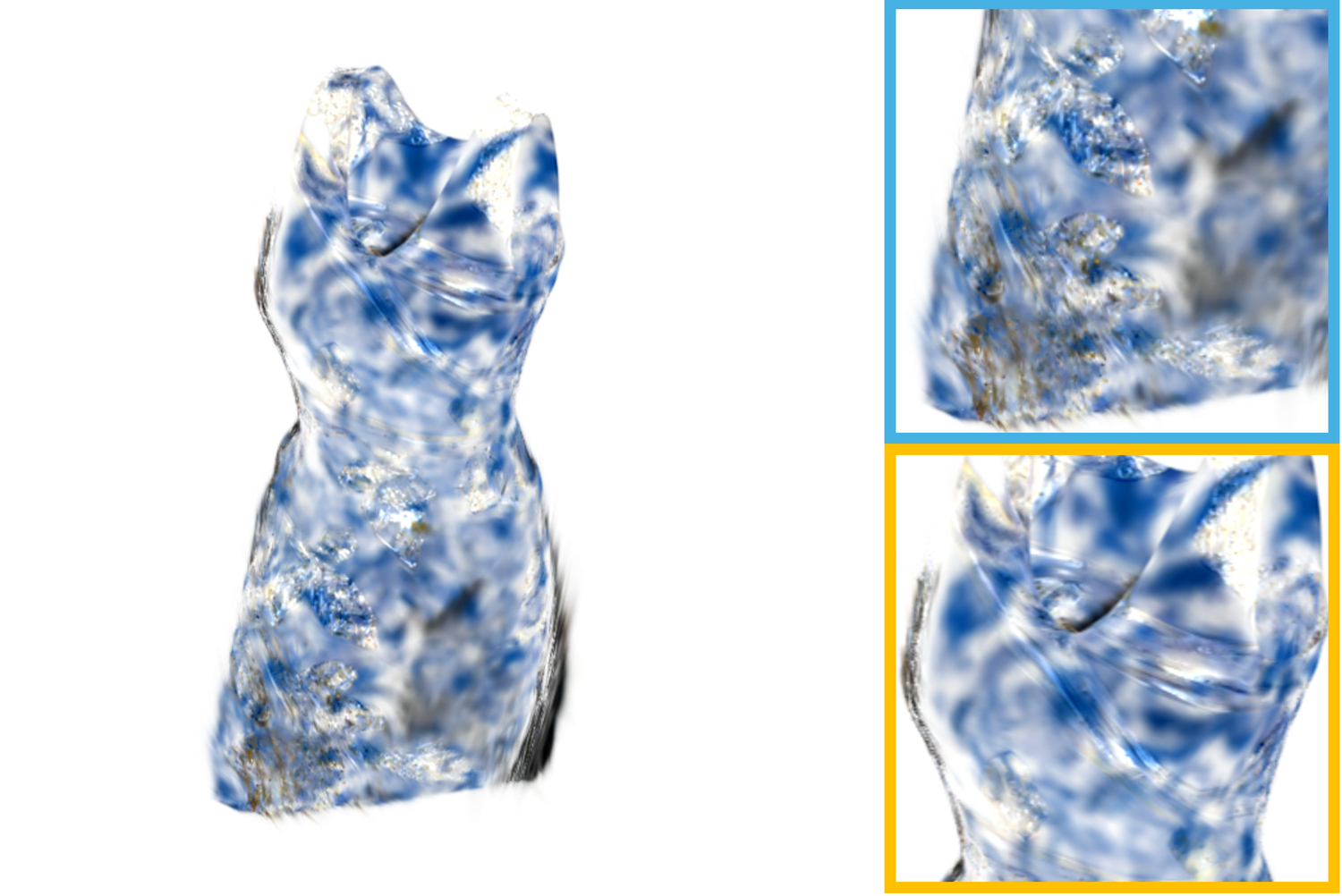} \\
\rotatebox[origin=c]{90}{flag}
    & \includegraphics[height=1.75cm,valign=m]{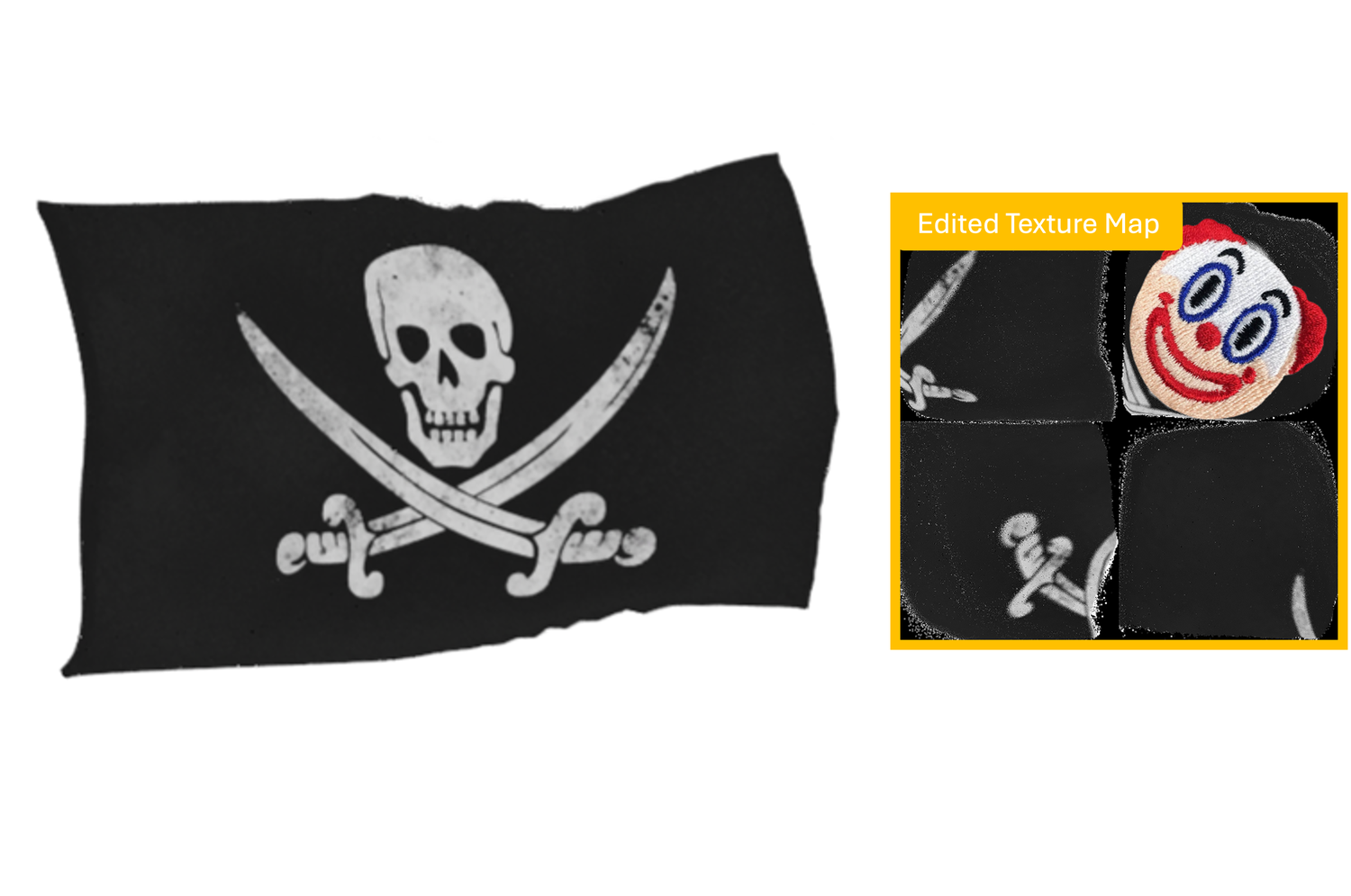}
    & \includegraphics[height=1.75cm,valign=m]{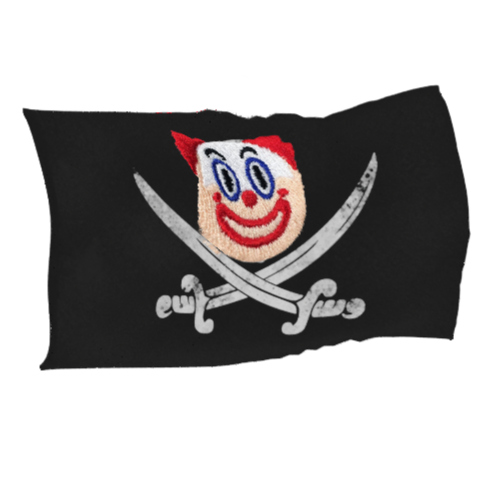}
    & \includegraphics[height=1.75cm,valign=m]{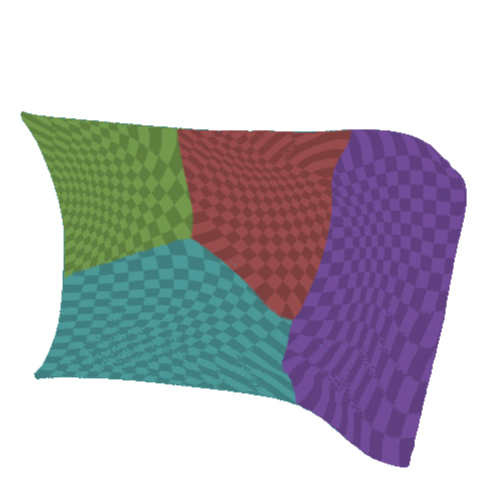}
    & \includegraphics[height=1.75cm,valign=m]{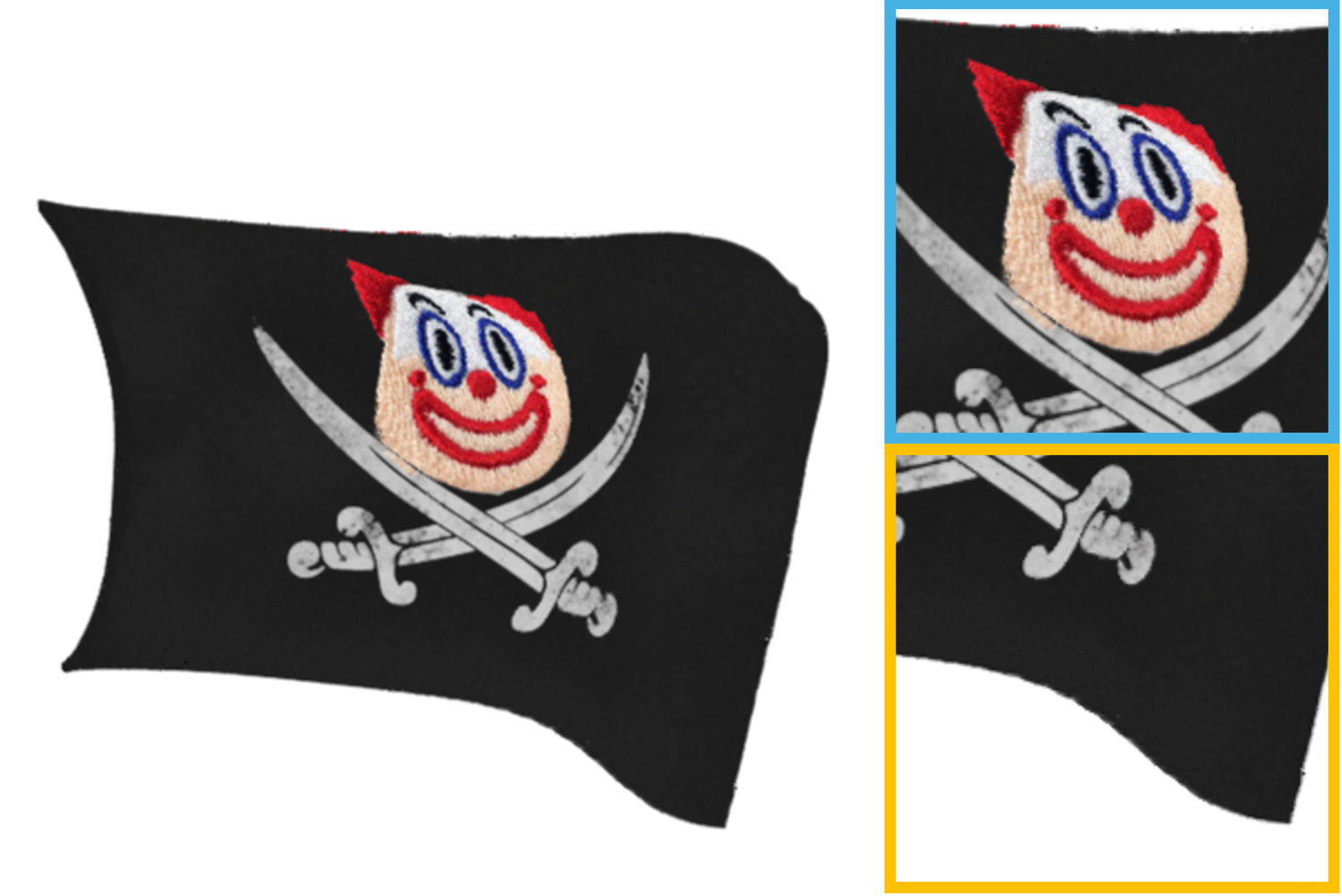}
    & \includegraphics[height=1.75cm,valign=m]{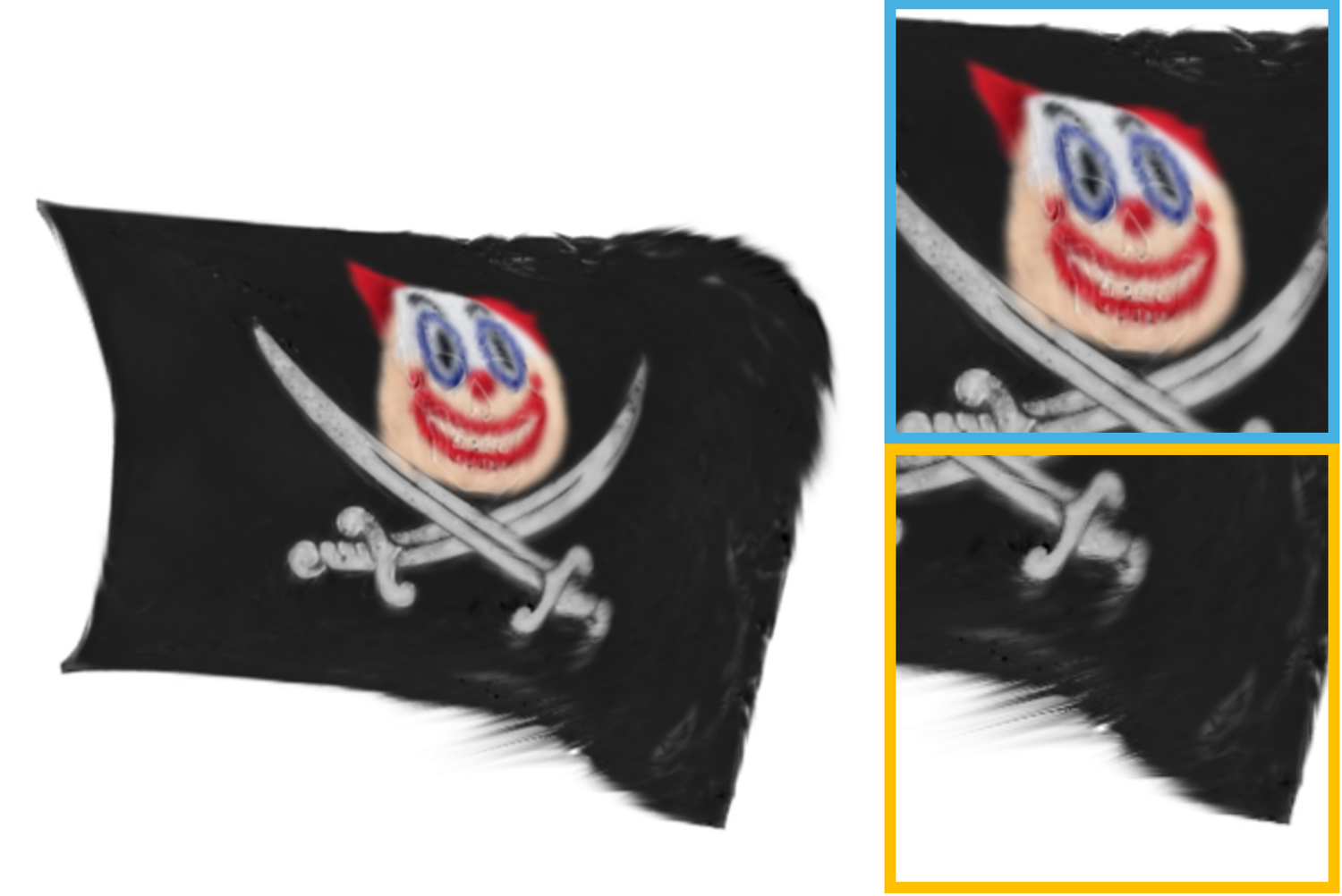} \\
\rotatebox[origin=c]{90}{rexy}
    & \includegraphics[height=1.75cm,valign=m]{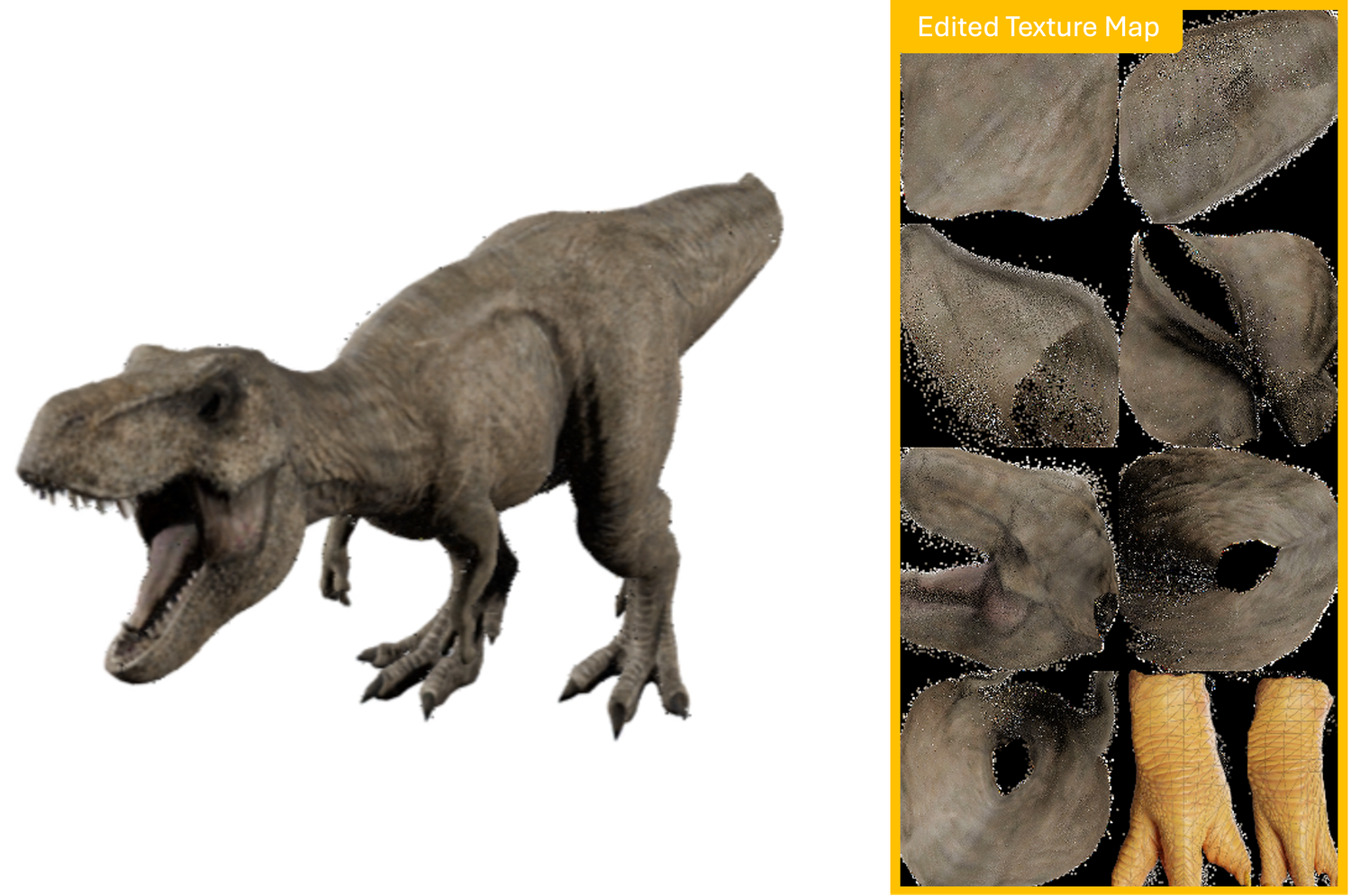}
    & \includegraphics[height=1.75cm,valign=m]{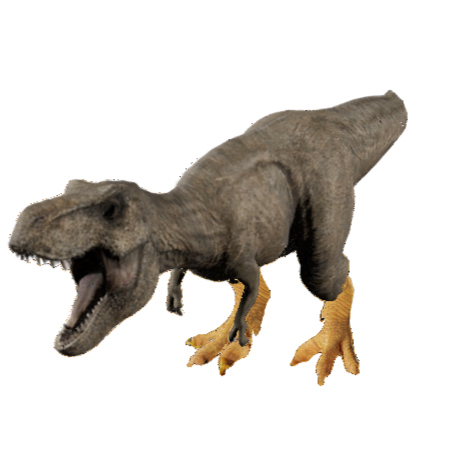}
    & \includegraphics[height=1.75cm,valign=m]{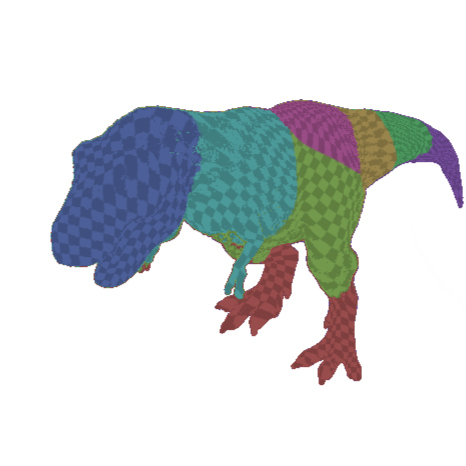}
    & \includegraphics[height=1.75cm,valign=m]{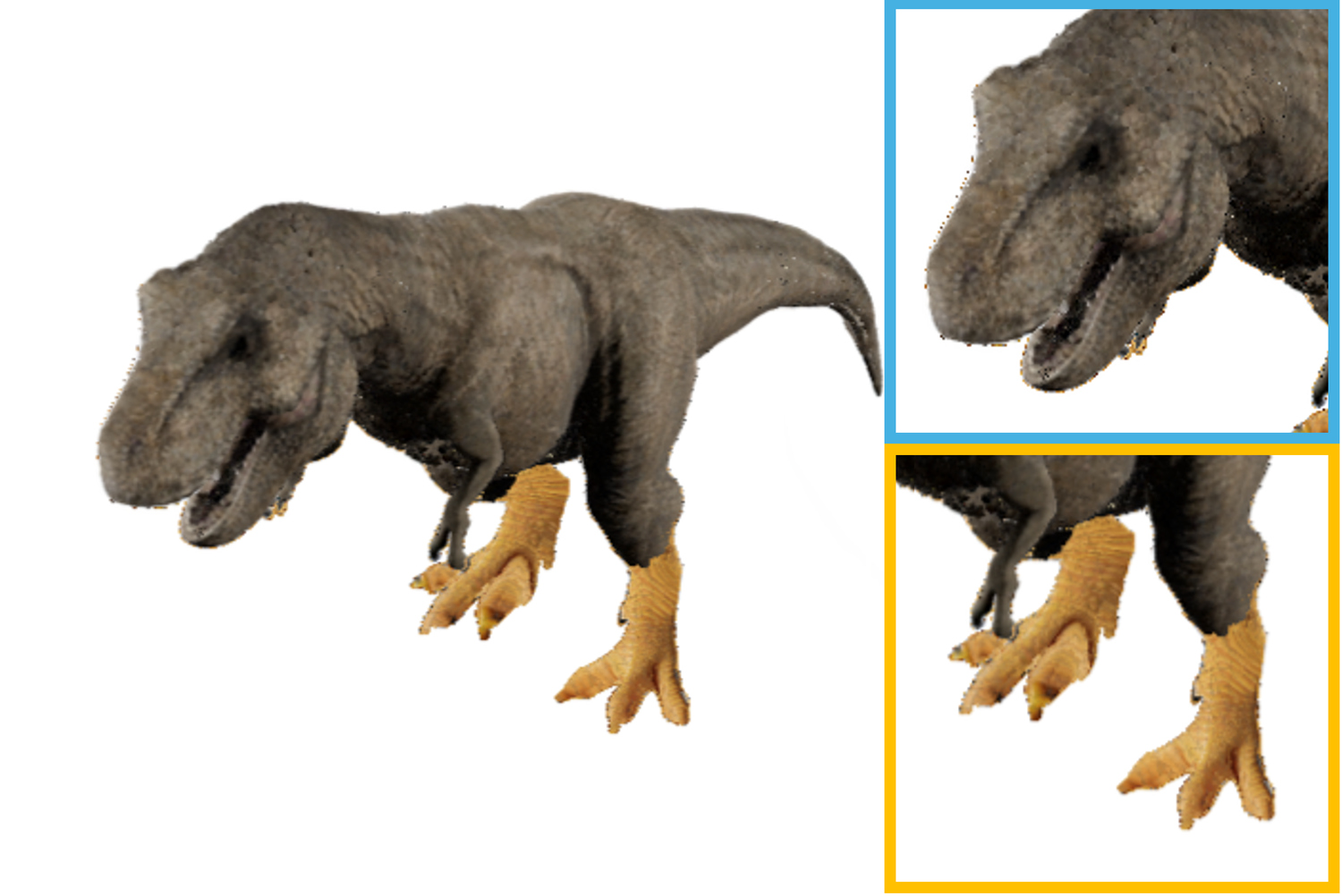}
    & \includegraphics[height=1.75cm,valign=m]{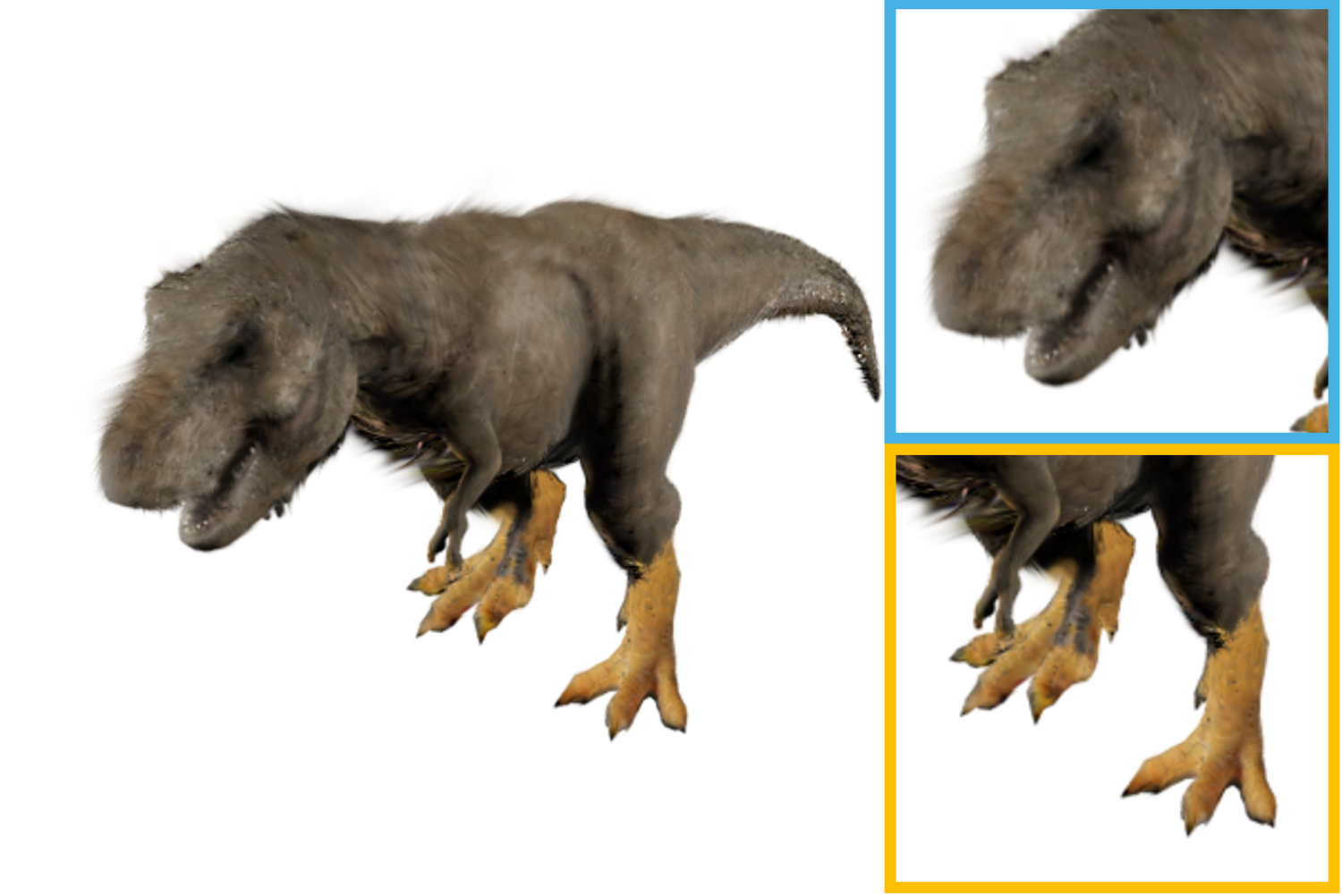} \\
\end{tabular}
\caption{Simultaneous geometry and texture editing under non-rigid deformation with both local and global texture edits.}
\label{fig:simul-edit-large-deform}
\end{figure*}

\paragraph{Non-rigid deformation with local and global edits.}
Figure~\ref{fig:simul-edit-large-deform} shows qualitative results on Objaverse assets with artist-created motion. PointGT preserves surface continuity under deformation and maintains edit attachment: the edited texture remains aligned with the deforming surface without noticeable drifting or popping.

\paragraph{Comparison to GSTex.}
We choose GSTex~\cite{Rong2024GStexPT} as the baseline for simultaneous geometry and texture editing. While Texture-GS~\cite{Xu2024TextureGSDT} learns a texture map, the quality of its learned texture map is insufficient for reliable texture-space editing in practice, with evidence in the supplementary. Textured Gaussians~\cite{Chao2024TexturedGF} is architecturally close to GSTex, they both learn a per-Gaussian texture map which is queried by ray--Gaussian intersection. Since texture editing is not its focus and its released code does not provide an editing pipeline, we treat GSTex as the representative method of this family. NeST-Splatting~\cite{Zhang2025NeuralST} stores texture in a global shell texture anchored in world space rather than to individual Gaussians, so deforming the geometry causes ray--Gaussian intersections to sample the texture at the wrong locations. It thus cannot edit geometry without external proxies such as cages~\cite{xu2022cagedeforming, yuan2022nerfedit}, and it also does not release an editing pipeline. We provide the full analysis of both in the supplementary.

In Figure~\ref{fig:editing-compare} we add a ``rocket'' icon to a butterfly scene from Objaverse~\cite{Deitke2022ObjaverseAU}. We edit PointGT by pasting the icon into the desired region on the 2D texture map. GSTex performs texture editing by editing a rendered view and projecting the edited pixels back to Gaussians. To apply the same local edit to the same region for a fair comparison, we use the edited pixels from our method's rendered view as the editing input for GSTex. As shown, our method maintains sharp edit details across different point budgets (1,000 to 30,000), while GSTex renders blurry textures, especially with 1,000 or 60,000 Gaussians. Moreover, as shown in Figure~\ref{fig:simul-edit-large-deform} our method maintains surface continuity after non-rigid deformation, while GSTex exhibits extruded Gaussians and surface artifacts under deformations.

\begin{figure*}[t]
\centering
\setlength\tabcolsep{2pt}
\renewcommand{\arraystretch}{1.5}
\scriptsize
\begin{tabularx}{1\linewidth}{l@{\hskip 1em}YYY}
\rotatebox[origin=c]{90}{GSTex~\cite{Rong2024GStexPT}}
                        &   \includegraphics[width=\hsize,valign=m]{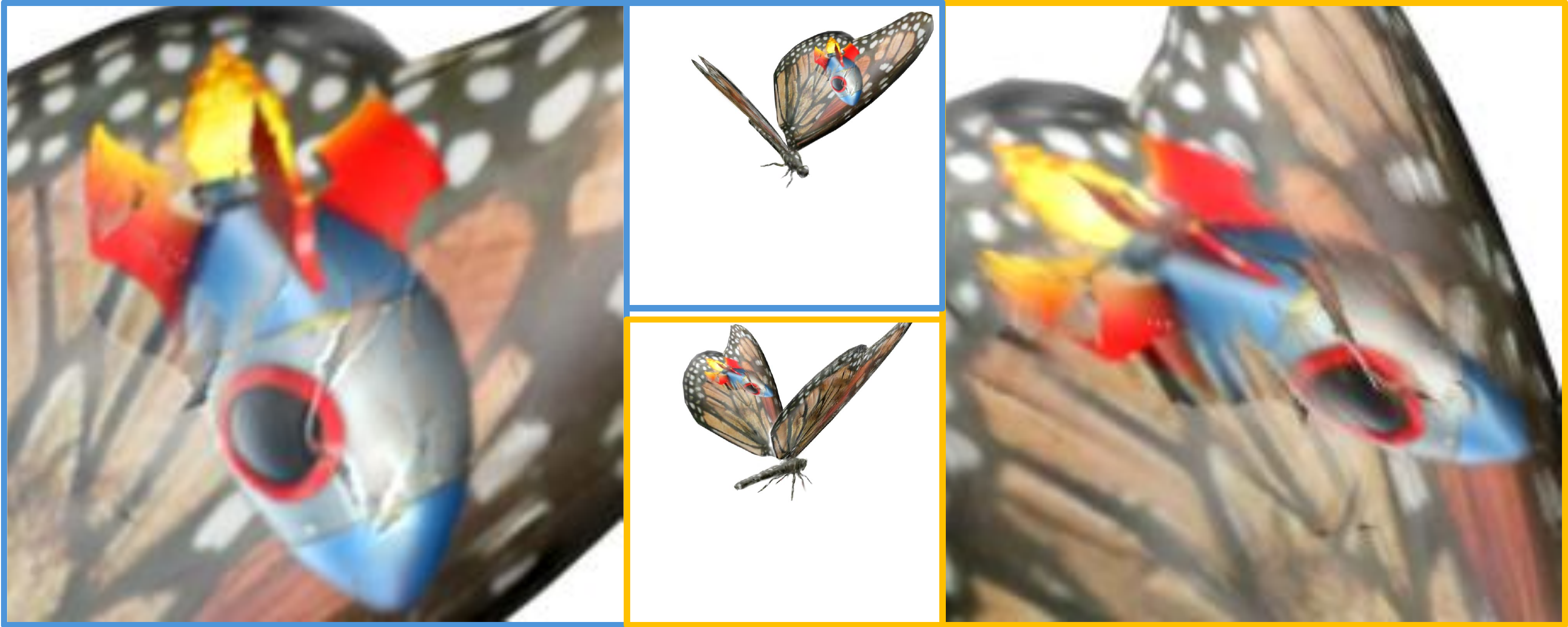}
                        &   \includegraphics[width=\hsize,valign=m]{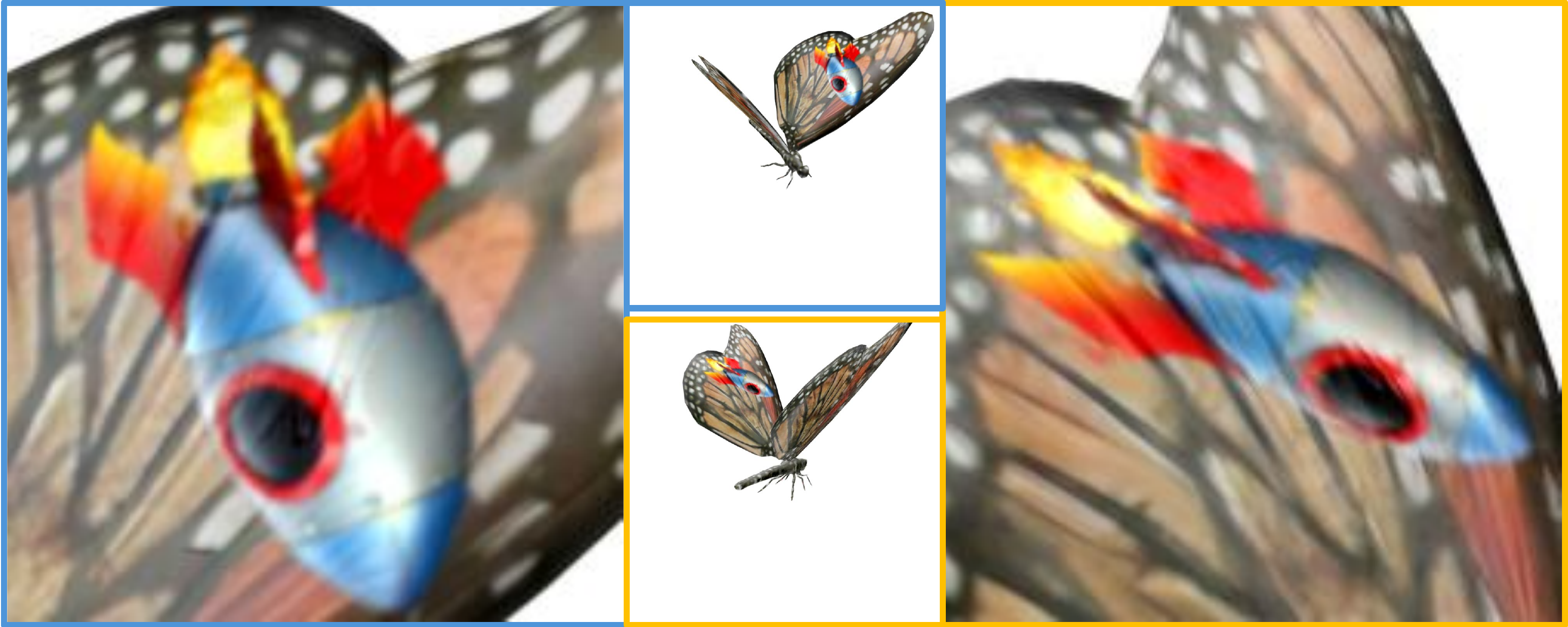}
                        &   \includegraphics[width=\hsize,valign=m]{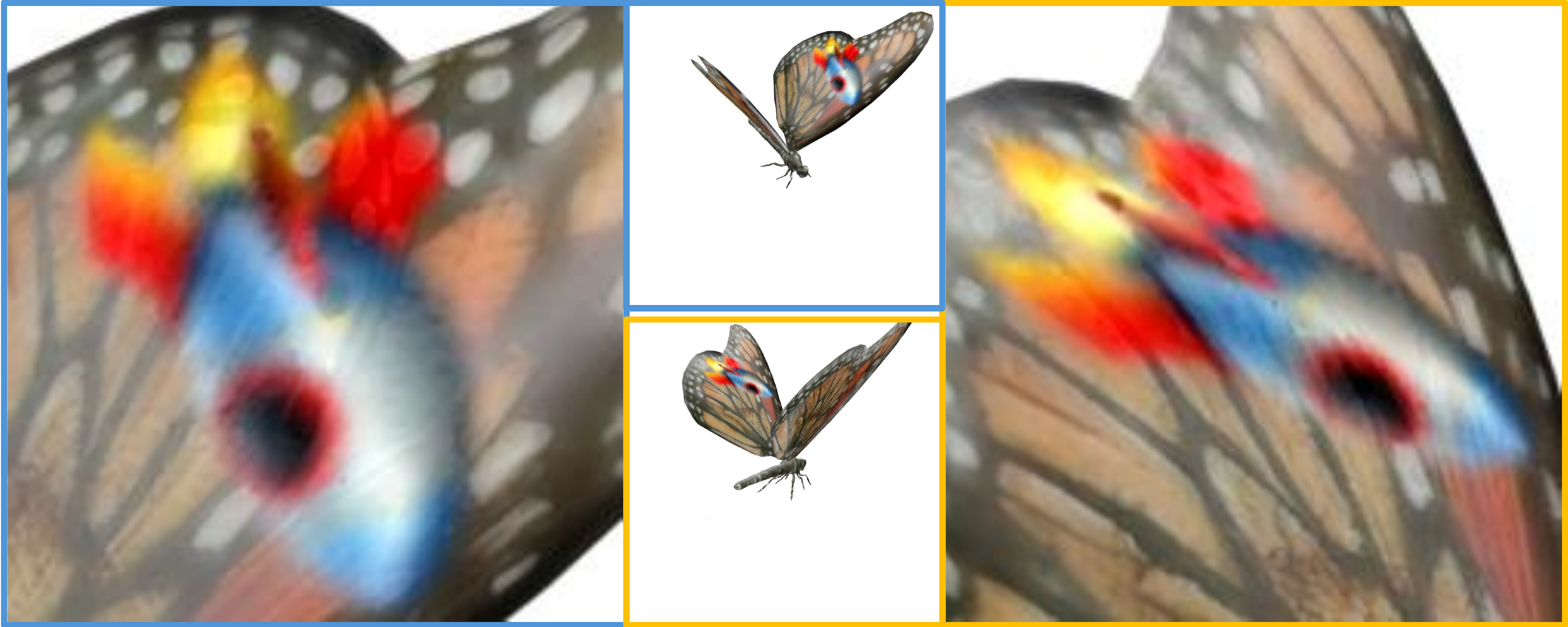}
                        \\[28pt]
\rotatebox[origin=c]{90}{Ours}
                        &   \includegraphics[width=\hsize,valign=m]{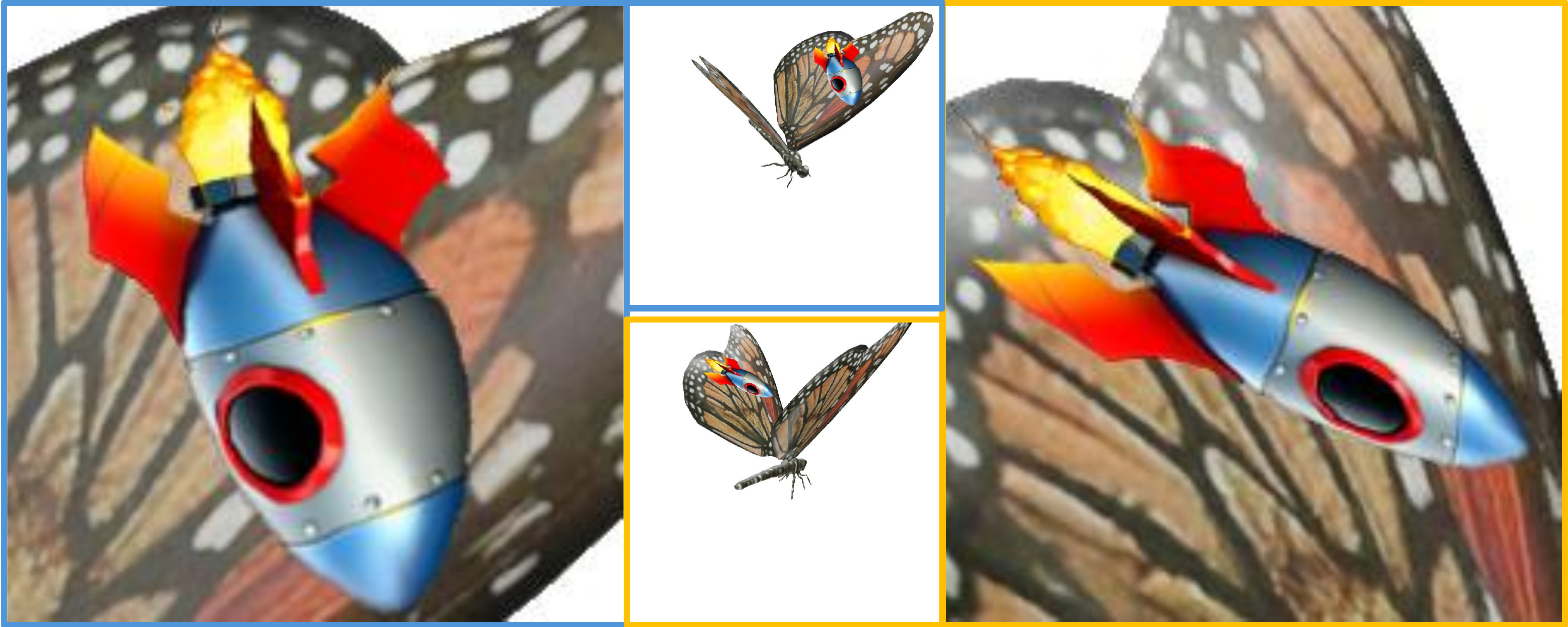}
                        &   \includegraphics[width=\hsize,valign=m]{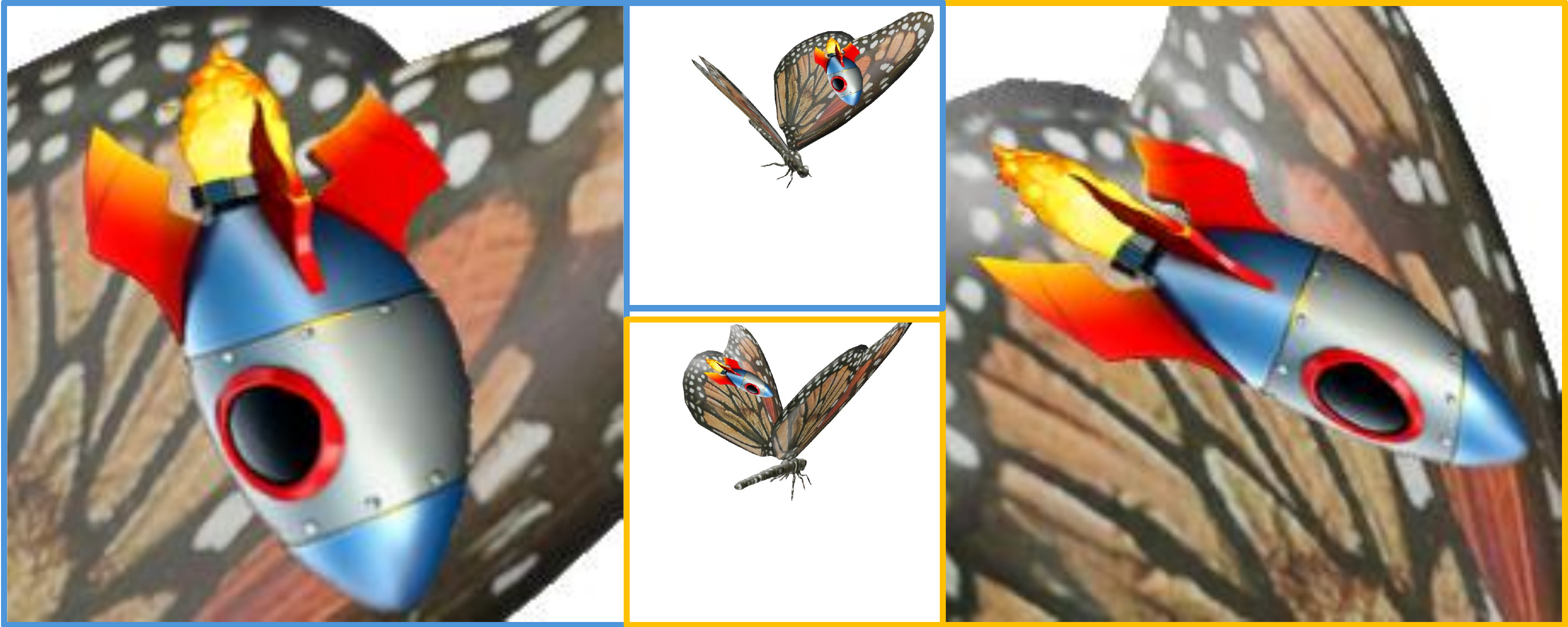}
                        &   \includegraphics[width=\hsize,valign=m]{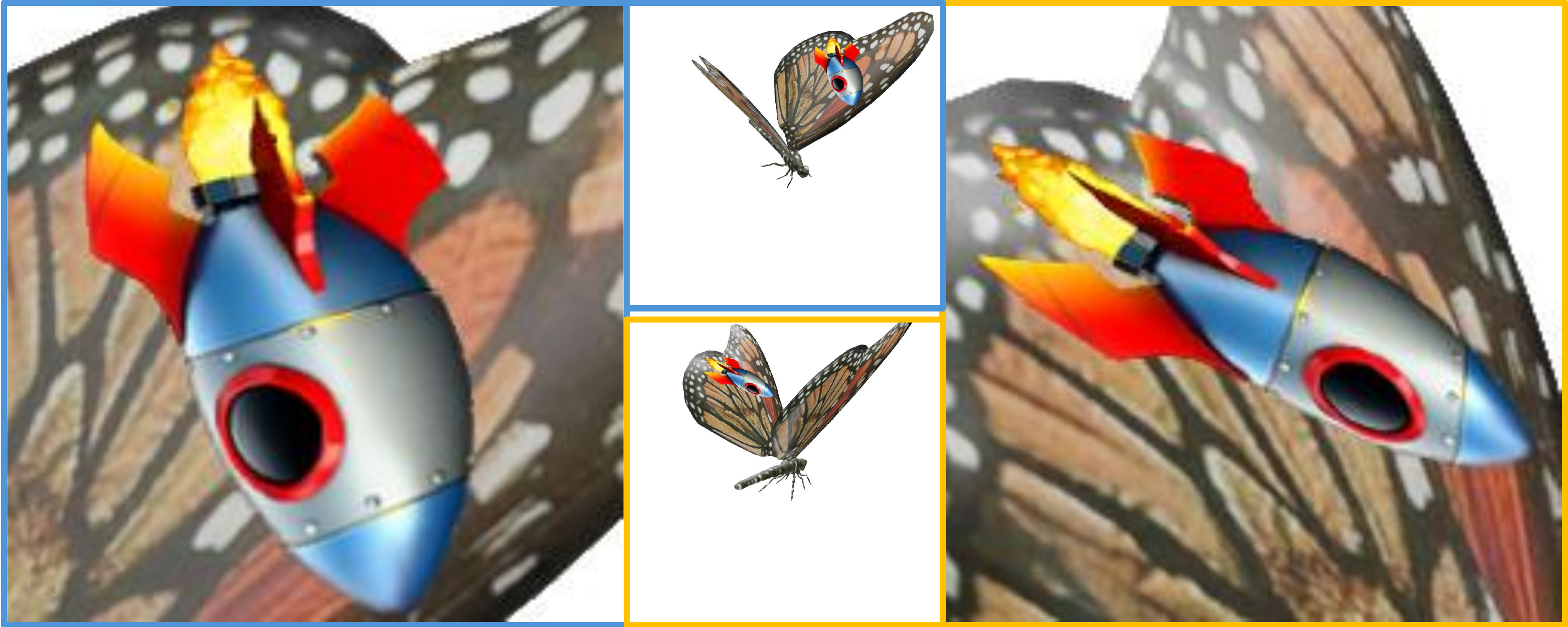}
                        \\
                        &   \# Primitives = 1k
                        &   \# Primitives = 5k
                        &   \# Primitives = 60k
\end{tabularx}
\caption{Texture editing compared with GSTex~\cite{Rong2024GStexPT} with different number of primitives.}
\label{fig:editing-compare}
\end{figure*}

\paragraph{Quantitative Evaluation.}

Due to the lack of ground truth, we use VBench~\cite{huang2023vbench} to quantitatively evaluate the editing quality of our method and GSTex. VBench is a benchmark suite that assesses generated videos along several dimensions, and is now widely used for video quality evaluation in general~\cite{huang2025vbench++,zheng2025vbench2}. Specifically, for each of the five scenes in Fig.~\ref{fig:simul-edit-large-deform}, we render a 360$^\circ$ orbit video in which the edited texture is shown while the geometry is animated over the video's time span using their artist-created motions. As shown in Table~\ref{tab:vbench}, our method either outperforms or matches GSTex across all five dimensions.

\begin{table*}[htb]
\centering
\setlength{\tabcolsep}{3pt}
\caption{VBench evaluation for Figure~\ref{fig:simul-edit-large-deform}. $\Delta\%$ is the relative improvement.}
\label{tab:vbench}
\resizebox{\linewidth}{!}{%
\begin{tabular}{l ccccc}
\toprule
Method & subject consistency & bkg consistency & motion smooth & aesthetic quality & imaging quality \\
\midrule
GSTex      & 0.827          & 0.923          & \textbf{0.985} & 0.434          & 0.520 \\
Ours       & \textbf{0.844} & \textbf{0.925} & 0.982          & \textbf{0.437} & \textbf{0.585} \\
$\Delta\%$ & $+2.16$        & $+0.24$        & $-0.40$        & $+0.55$        & $+12.48$ \\
\bottomrule
\end{tabular}%
}
\end{table*}

\begin{table*}[h]
  \centering
  \caption{Quantitative comparison of Novel-view Synthesis.}
  \label{tab:quantitative_results_alt}
  \scriptsize

  \resizebox{\textwidth}{!}{%
    \begin{tabular}{l cccr cccr}
    \toprule
    & \multicolumn{4}{c}{\textbf{DTU}~\cite{Jensen2014LargeSM}} & \multicolumn{4}{c}{\textbf{NeRF Synthetic}~\cite{mildenhall2021nerf}} \\
    \cmidrule(lr){2-5} \cmidrule(lr){6-9}
    Method & PSNR $\uparrow$ & SSIM $\uparrow$ & LPIPS $\downarrow$ & \# Pts & PSNR $\uparrow$ & SSIM $\uparrow$ & LPIPS $\downarrow$ & \# Pts \\
    \midrule

    \multicolumn{9}{l}{\textit{Authors' recommended primitive counts}} \\
    Texture-GS~\cite{Xu2024TextureGSDT} & 30.53 & 0.920 & 0.083 & 90k & 28.97 & 0.938 & 0.055 & 90k \\
    GSTex~\cite{Rong2024GStexPT} & 32.87 & 0.956 & 0.038 & 186k & 33.25 & 0.969 & 0.024 & 100k \\
    Textured-Gaussian~\cite{Chao2024TexturedGF} & 33.61 & 0.970 & 0.056 & 240k & 33.24 & 0.967 & 0.043 & 190k \\
    NeST Splatting~\cite{Zhang2025NeuralST} & \textbf{33.65} & 0.964 & 0.042 & 80k & 33.37 & 0.967 & 0.032 & 73k \\
    PAPR~\cite{zhang2023papr} & 29.34 & 0.952 & 0.073 & 30k & 32.07 & 0.971 & 0.038 & 30k \\
    Ours & 33.48 & \textbf{0.973} & \textbf{0.023} & 30k & \textbf{33.57} & \textbf{0.982} & \textbf{0.021} & 30k \\
    \midrule

    \multicolumn{9}{l}{\textit{Fixed primitive count (5k)}} \\
    Texture-GS~\cite{Xu2024TextureGSDT} & 26.81 & 0.833 & 0.206 & 5k & 19.29 & 0.795 & 0.213 & 5k \\
    GSTex~\cite{Rong2024GStexPT} & 28.92 & 0.932 & 0.072 & 5k & 30.20 & 0.897 & 0.149 & 5k \\
    Textured-Gaussian~\cite{Chao2024TexturedGF} & 29.58 & 0.912 & 0.080 & 5k & 26.21 & 0.919 & 0.086 & 5k \\
    NeST Splatting~\cite{Zhang2025NeuralST} & 32.68 & 0.966 & 0.056 & 5k & 30.48 & 0.958 & 0.057 & 5k \\
    PAPR~\cite{zhang2023papr} & 24.87 & 0.846 & 0.040 & 5k & 30.38 & 0.963 & 0.048 & 5k \\
    Ours & \textbf{33.25} & \textbf{0.970} & \textbf{0.024} & 5k & \textbf{31.01} & \textbf{0.976} & \textbf{0.039} & 5k \\
    \bottomrule
  \end{tabular}
  } %
  \label{tab:quantitative_results}
\end{table*}

\subsection{Novel View Synthesis}

We compare PointGT for novel view synthesis to PAPR~\cite{zhang2023papr}, Texture-GS~\cite{Xu2024TextureGSDT}, GSTex~\cite{Rong2024GStexPT}, NeST-Splatting~\cite{Zhang2025NeuralST}, and Textured Gaussians~\cite{Chao2024TexturedGF} on the Blender~\cite{mildenhall2021nerf} dataset and the DTU dataset~\cite{Huang20242DGS, Jensen2014LargeSM}. Table~\ref{tab:quantitative_results} shows quantitative results for novel view synthesis on the Blender and DTU datasets. We compare against baselines in two settings: (1) the original settings reported in prior works (no cap on the number of primitives), and (2) capping all methods at 5,000 primitives. As shown in Table~\ref{tab:quantitative_results}, our method outperforms baselines in both settings.

\subsection{Ablation Study}

\subsubsection{Evaluation of the Geometry Regularizers}

We ablate the two regularizers described in Sec.~\ref{sec:pre-training-papr}.

\begin{figure*}[htb]
\centering
\setlength\tabcolsep{2pt}
\renewcommand{\arraystretch}{1.2}
\scriptsize
\begin{tabularx}{1.0\linewidth}{YYYY}
    \includegraphics[width=\hsize,valign=m]{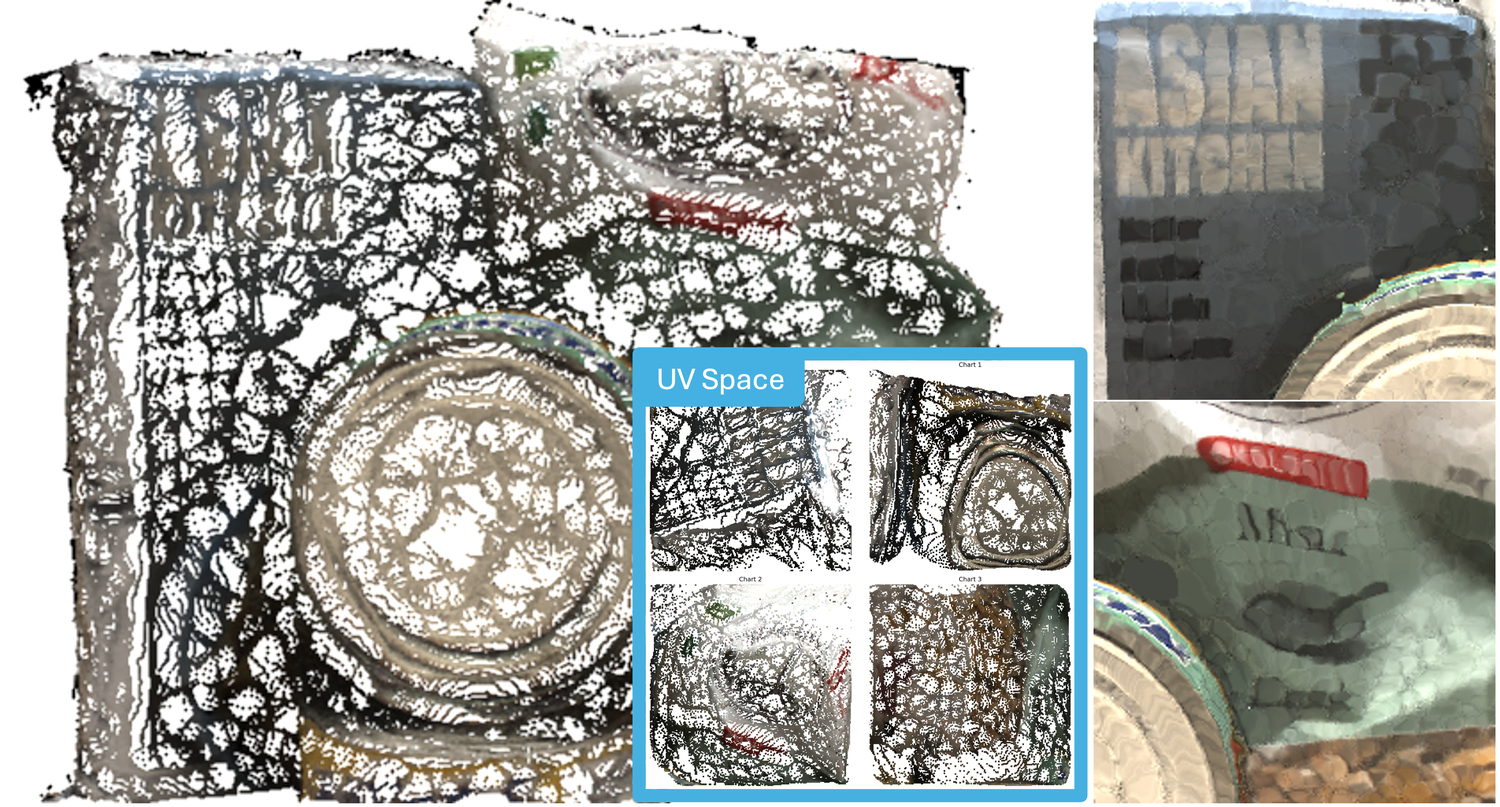}
    & \includegraphics[width=\hsize,valign=m]{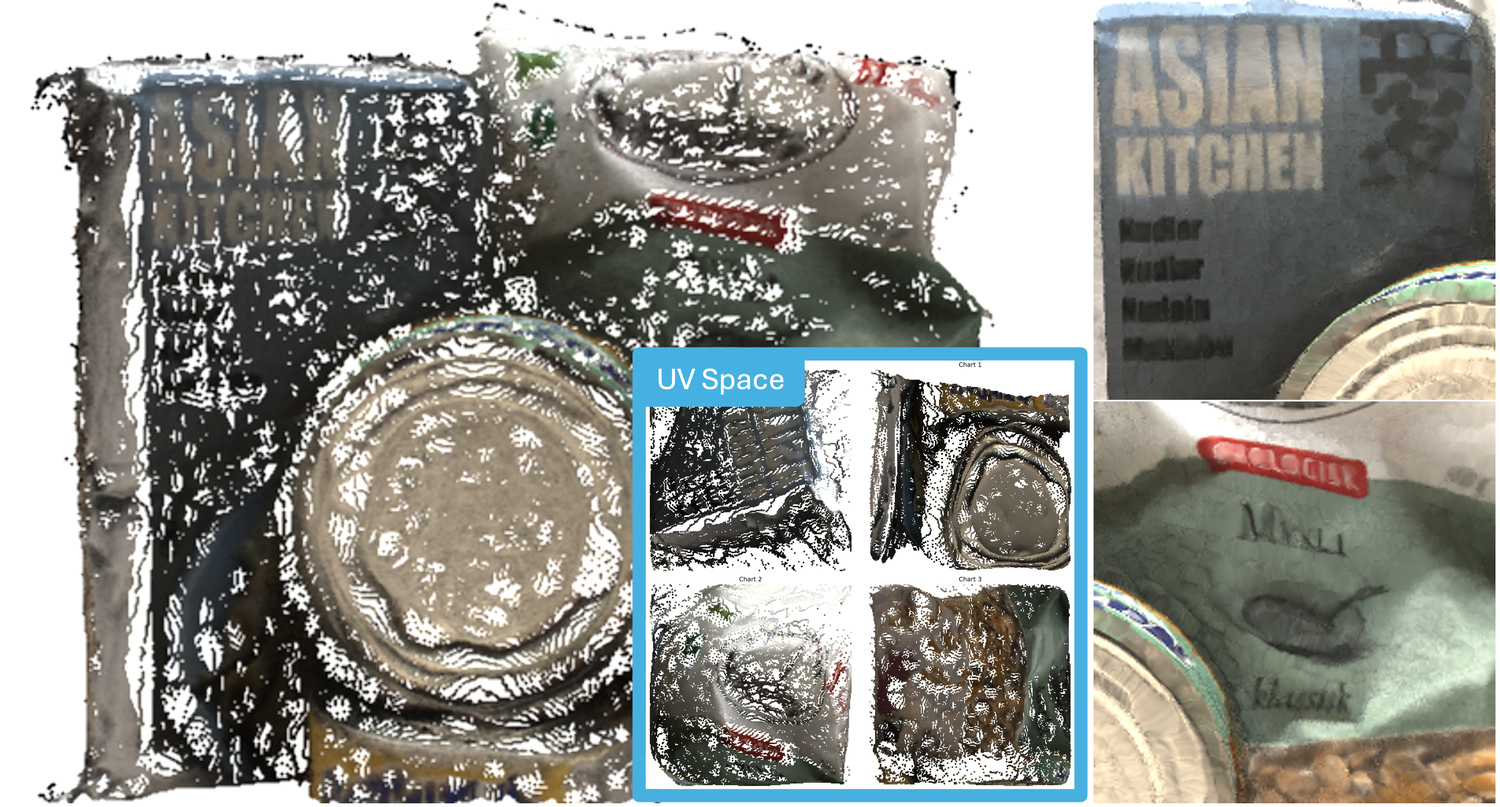}
    & \includegraphics[width=\hsize,valign=m]{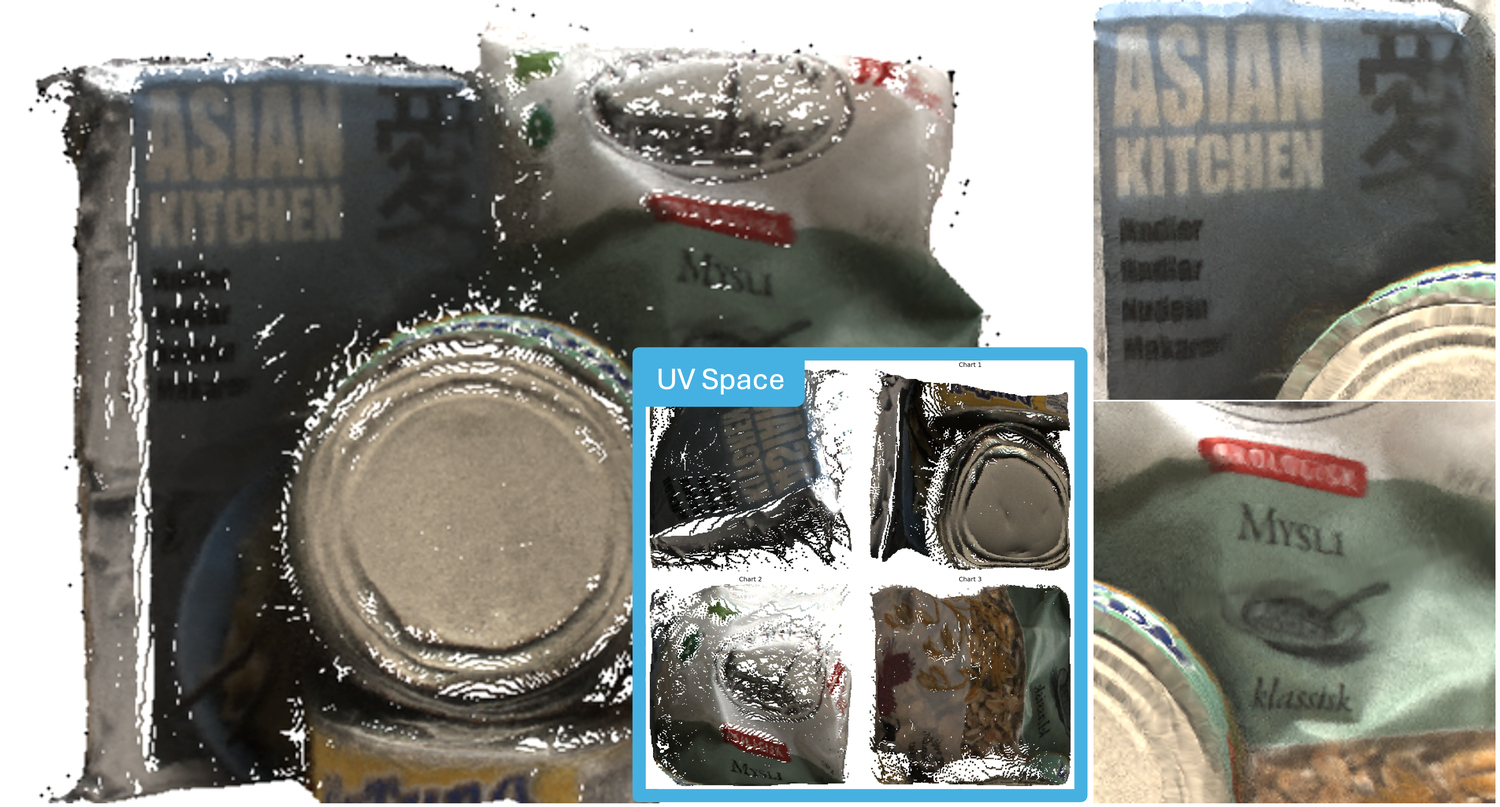}
    & \includegraphics[width=\hsize,valign=m]{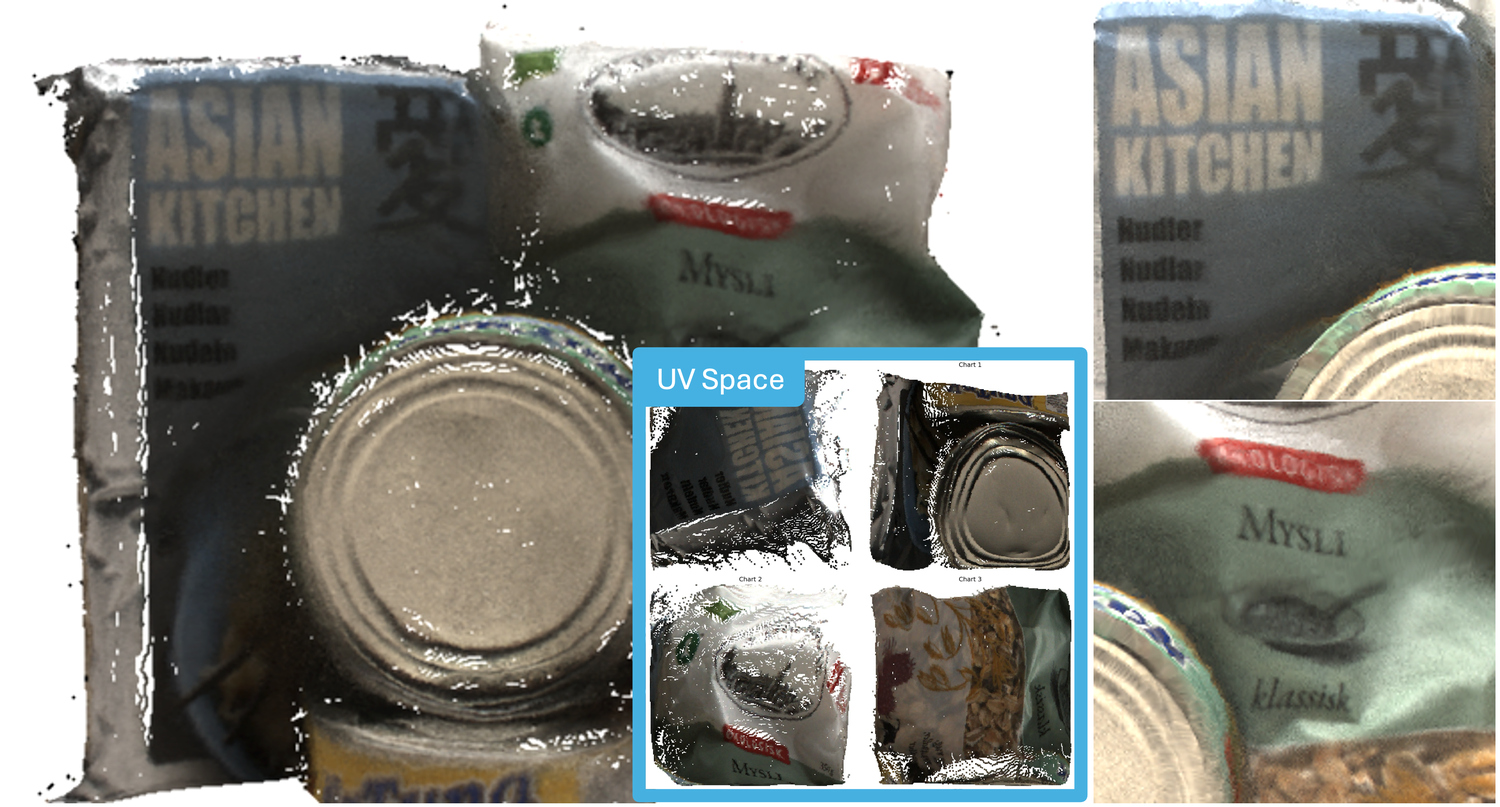}
    \\
    $\lambda_{\text{on-ray}}=0$
    & $\lambda_{\text{on-ray}}=0.001$
    & $\lambda_{\text{on-ray}}=0.005$
    & $\lambda_{\text{on-ray}}=0.01$
\end{tabularx}
\caption{Ablation of the on-ray regularizer with different weights. For each weight, we show the predicted surface points in 3D (left), the corresponding UV coordinates in the 2D atlas (middle), and the final rendered texture from texture maps (right). As shown, the on-ray loss ensures a better coverage of the surface, which leads to more uniform UV coordinates and gives a sharper texture in the renderings.}
\label{fig:abla-sp-close-to-ray}
\end{figure*}

\paragraph{On-ray regularizer.} Figure~\ref{fig:abla-sp-close-to-ray} shows the ablation of the on-ray loss (close-to-ray). We visualize the predicted surface points from a camera view, the same points in UV space (across 4 atlas maps), and crops of the final renderings using varying loss weights. Without the on-ray loss, the predicted surface points collapse and cluster around discrete support points. Consequently, querying the texture map with these points fails to smoothly interpolate the texture, degrading the effective texture resolution. With the on-ray loss, the surface points are continuous and smoothly interpolate the surface, yielding much clearer details in the renderings.

\begin{figure*}[htb]
    \centering
    \begin{subfigure}[t]{0.38\linewidth}
        \centering
        \resizebox{\linewidth}{!}{%
        \setlength\tabcolsep{2pt}
        \begin{tabular}{cc}
            \includegraphics[height=1.9cm,valign=m]{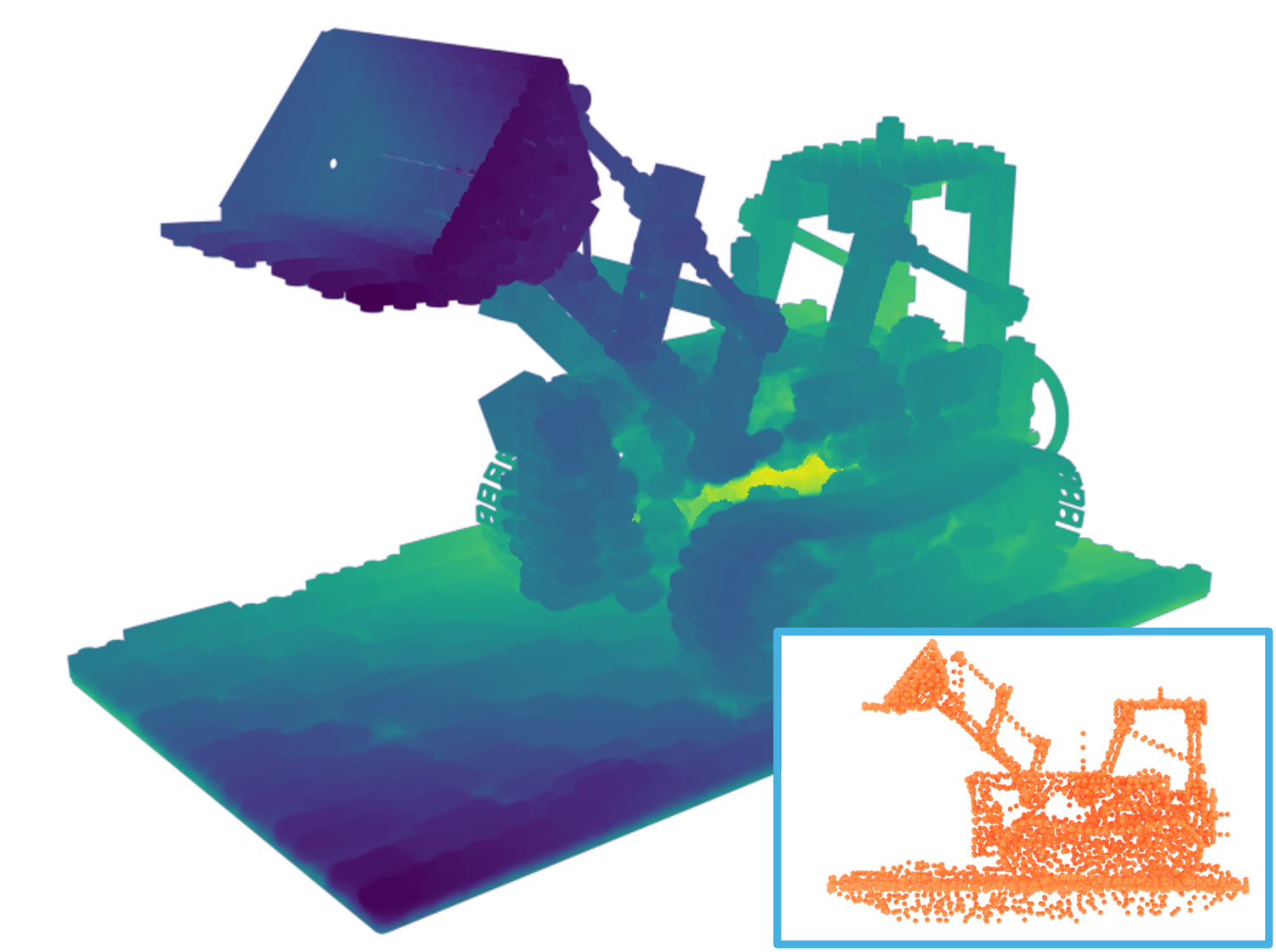} &
            \includegraphics[height=1.9cm,valign=m]{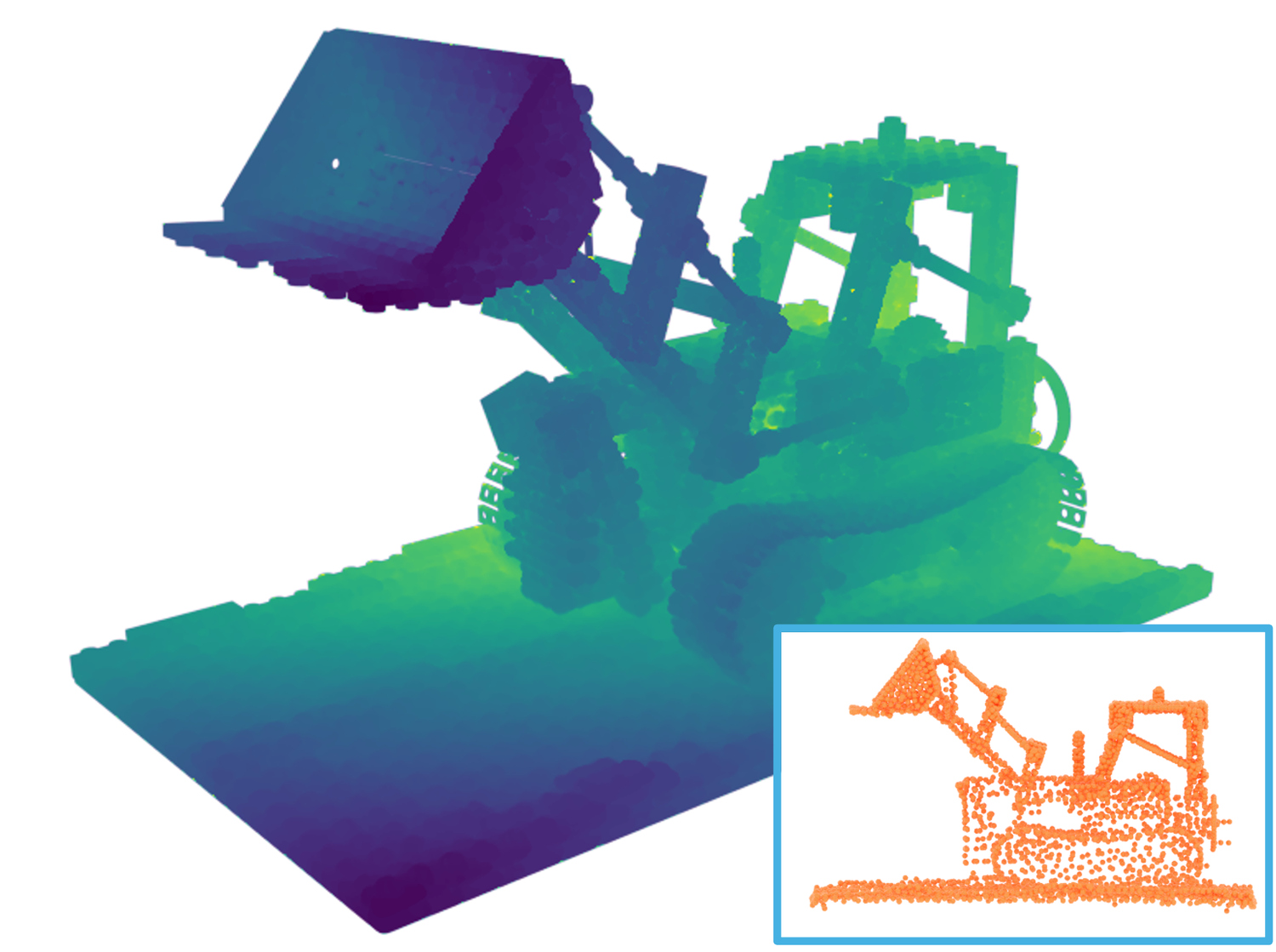} \\
            \scriptsize w/o on-surface loss &
            \scriptsize w/ on-surface loss 
        \end{tabular}
        }
        \caption{The on-surface regularizer reduces noisy points and produces a smoother surface as shown in the depth maps.}
        \label{fig:abla-sp-close-to-surface}
    \end{subfigure}
    \hfill
    \begin{subfigure}[t]{0.59\linewidth}
        \centering
        \resizebox{\linewidth}{!}{%
        \setlength\tabcolsep{1pt}
        \begin{tabular}{cccc}
            \includegraphics[height=1.9cm,valign=m]{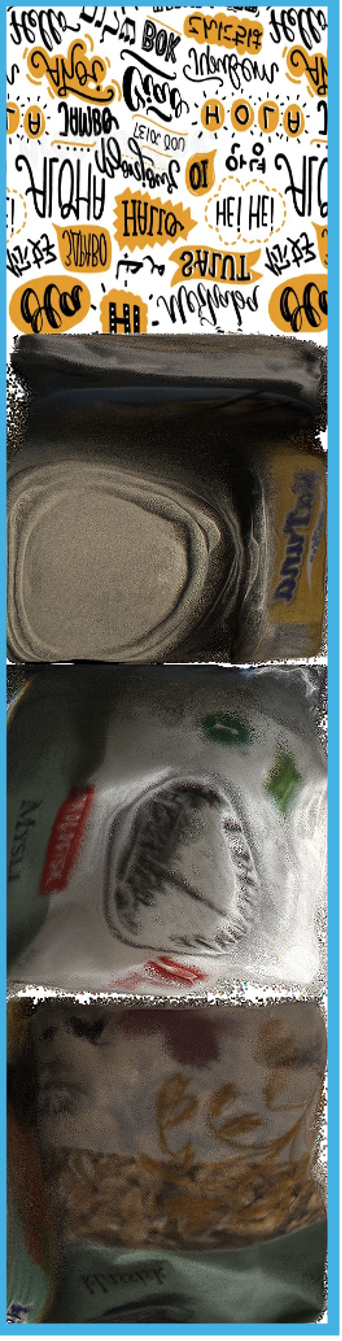} &
            \includegraphics[height=1.9cm,valign=m]{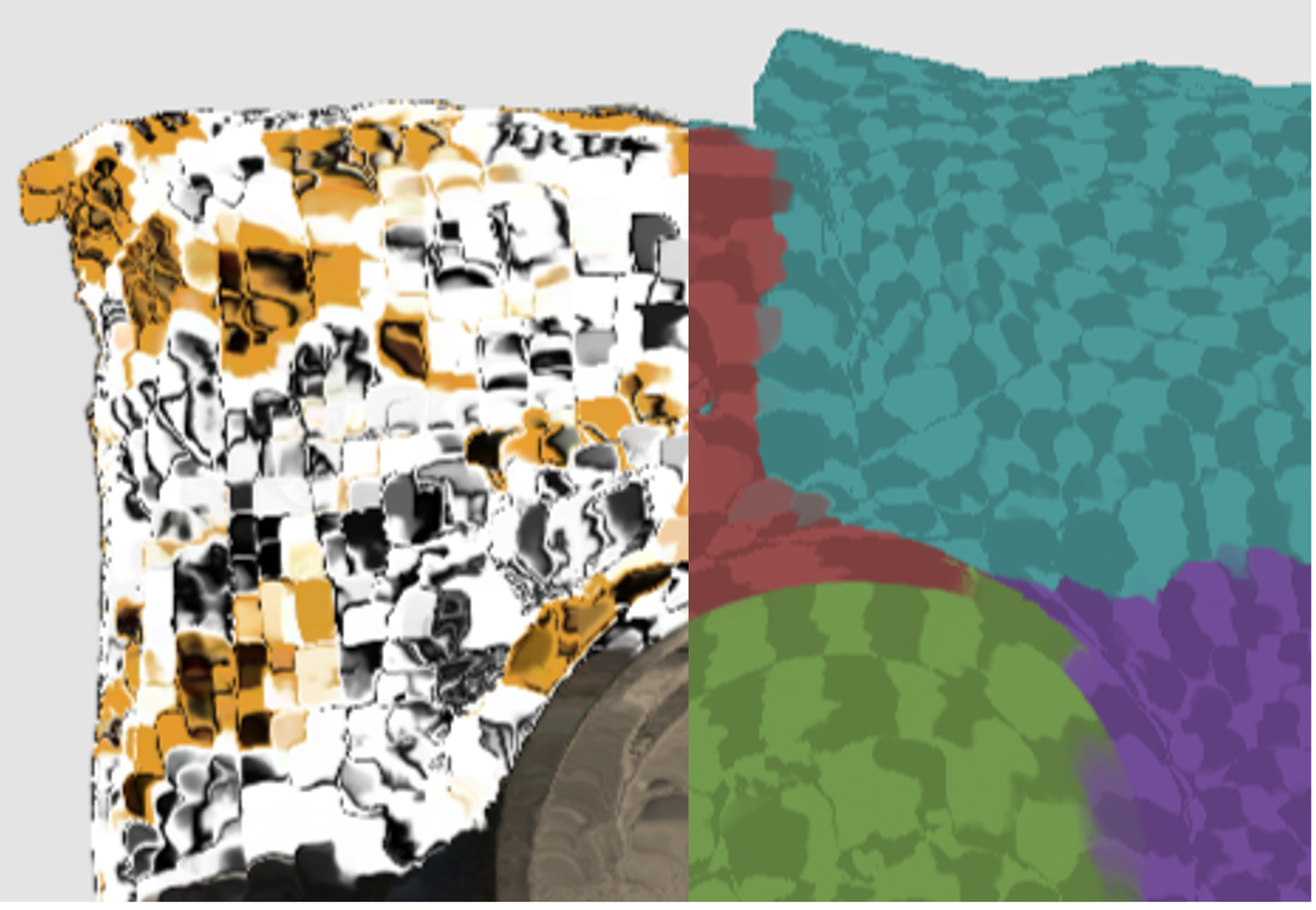} &
            \includegraphics[height=1.9cm,valign=m]{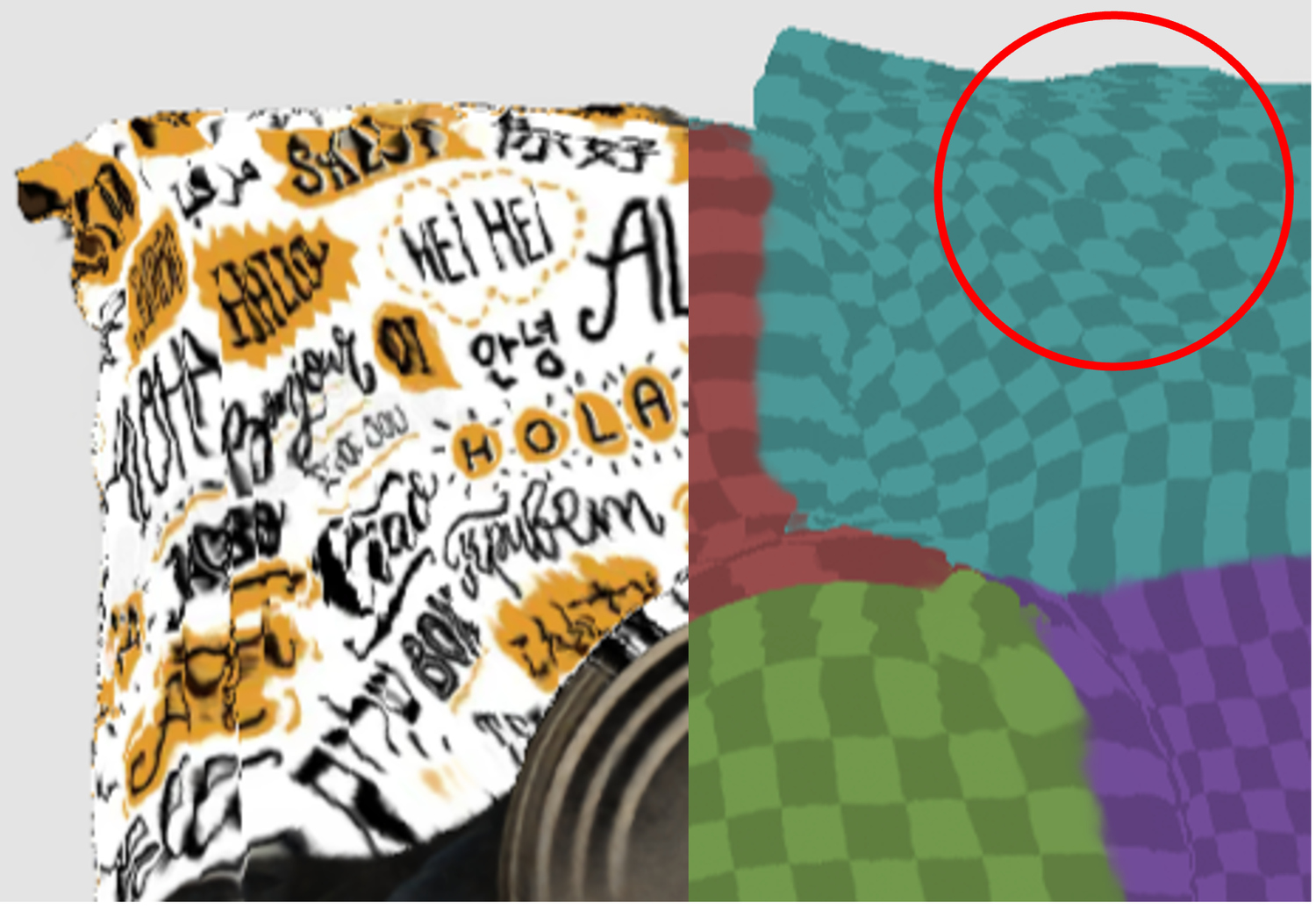} &
            \includegraphics[height=1.9cm,valign=m]{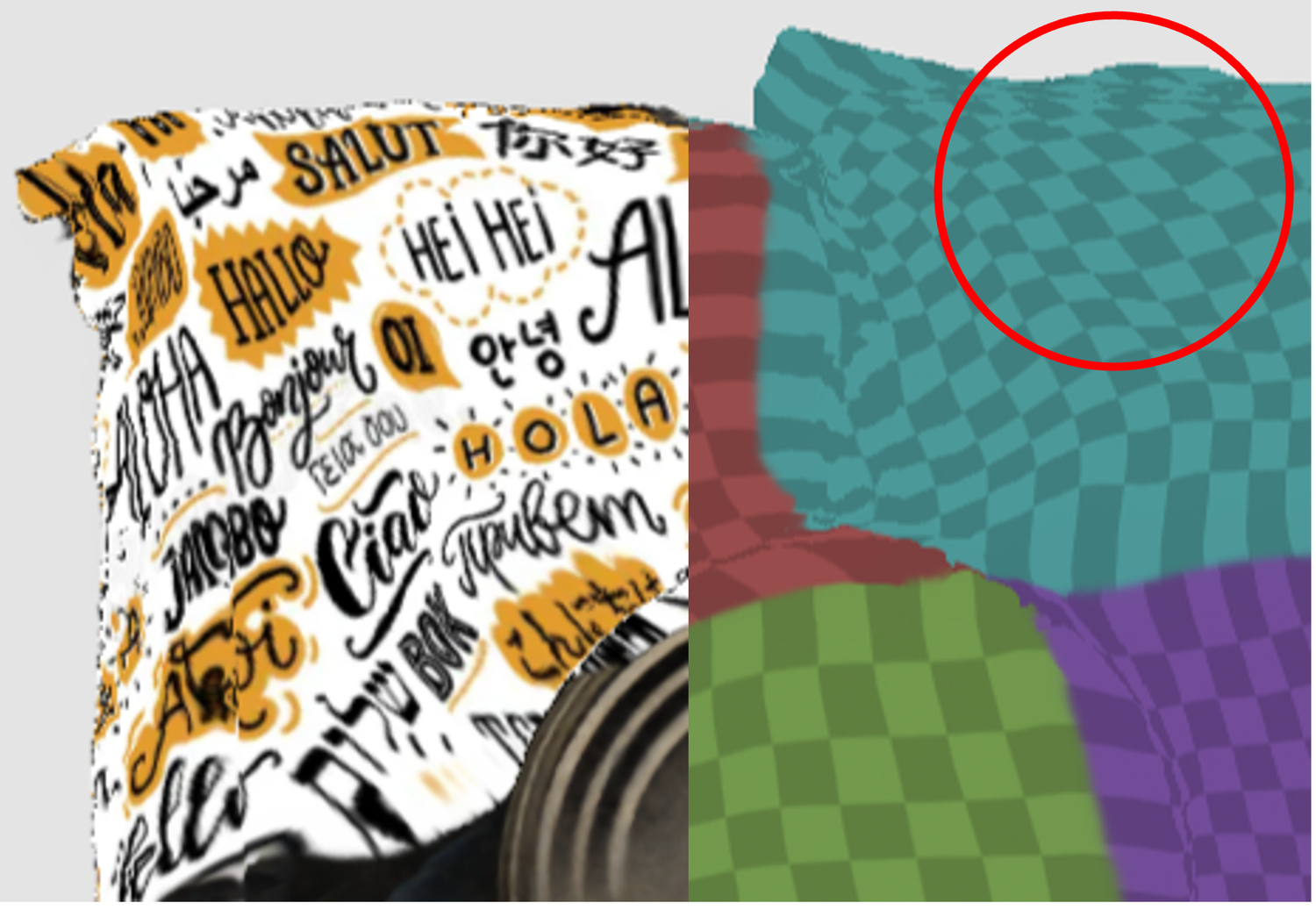} \\
            &
            \scriptsize \shortstack[c]{baseline} &
            \scriptsize \shortstack[c]{+ on-ray regularizer} &
            \scriptsize \shortstack[c]{+ deform-aware corr.}
        \end{tabular}
        }
        \caption{Ablation of canonical correspondence. Naïve correspondence (Eq.~\ref{eq:naive-canonical-sum}) exhibits texture drift while our deformation-aware correspondence (Eq.~\ref{eq:deformation-aware-correspondence}) improves stability.}
        \label{fig:abla-correspondence-stability}
    \end{subfigure}
    \caption{Ablation studies on geometry regularization and canonical correspondence.}
    \label{fig:ablations-combined}
\end{figure*}

\begin{figure}[htb]
\centering
    \setlength\tabcolsep{10pt}
    \renewcommand{\arraystretch}{1.0}
    \resizebox{1\textwidth}{!}{%
    \begin{tabular}{ccc}
        $\lambda_{\text{distortion}}=0$ &
        $\lambda_{\text{distortion}}=0.01$ &
        $\lambda_{\text{distortion}}=0.4$ \\
        \includegraphics[width=0.32\linewidth]{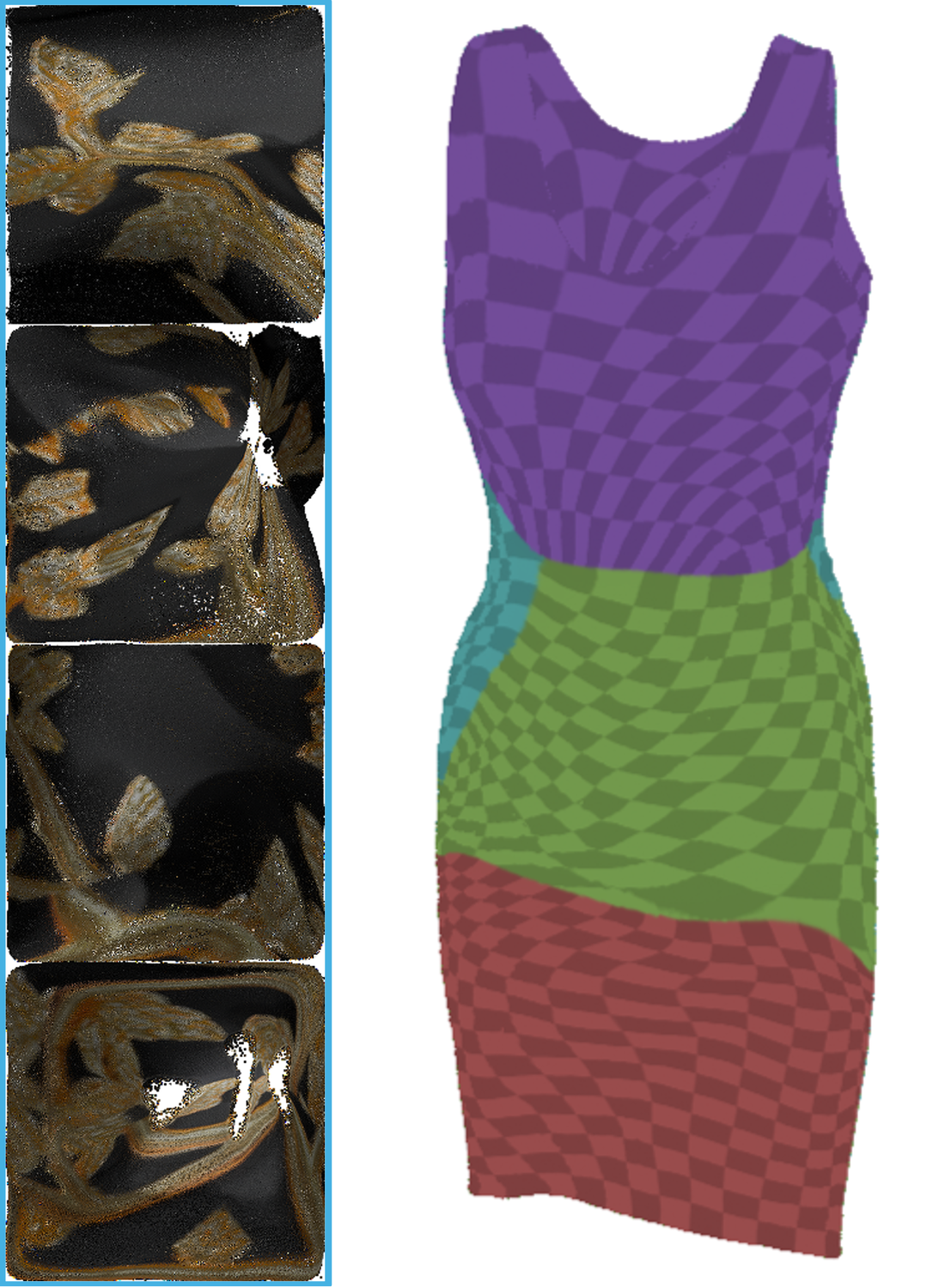} &
        \includegraphics[width=0.32\linewidth]{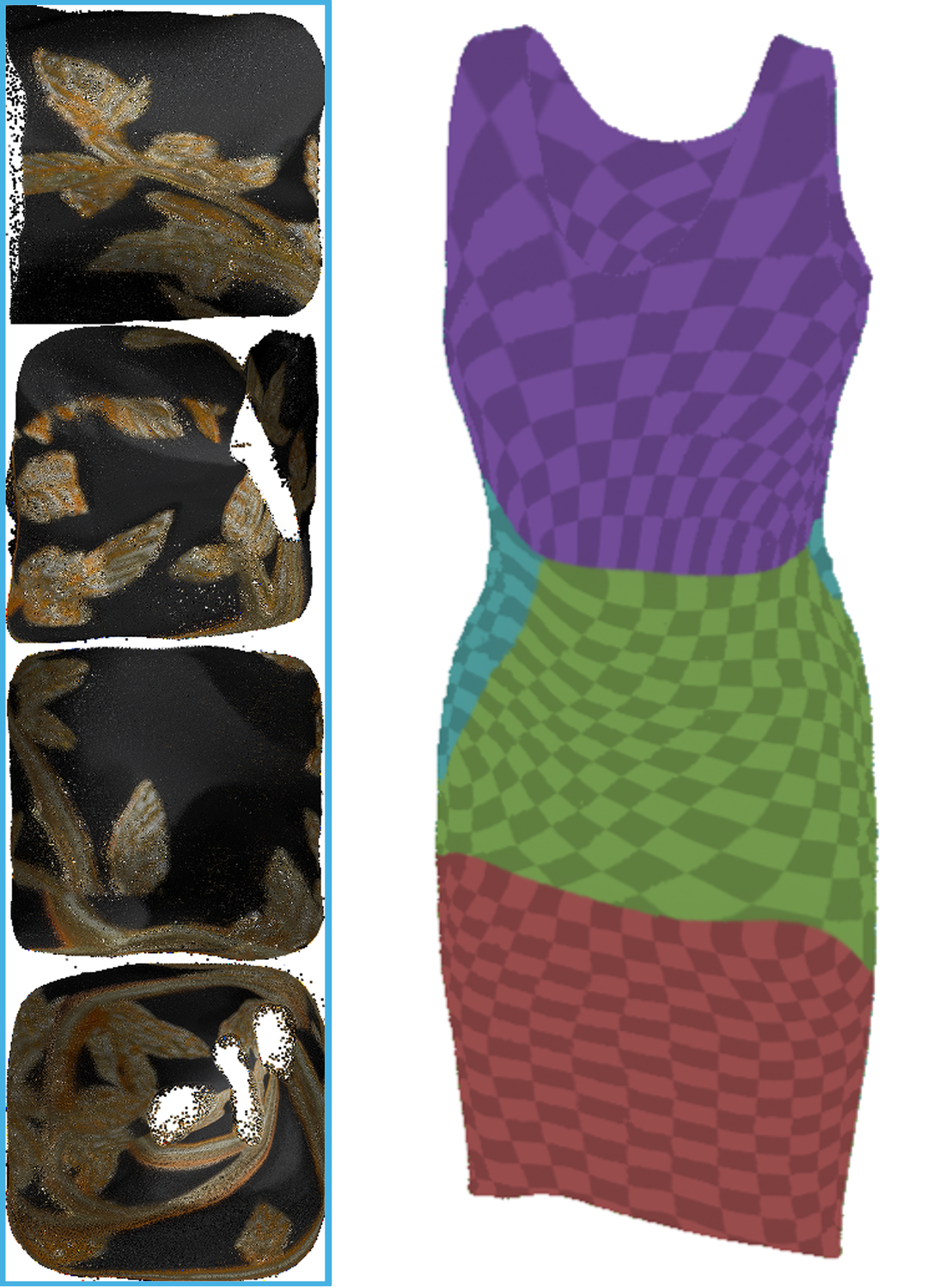} &
        \includegraphics[width=0.32\linewidth]{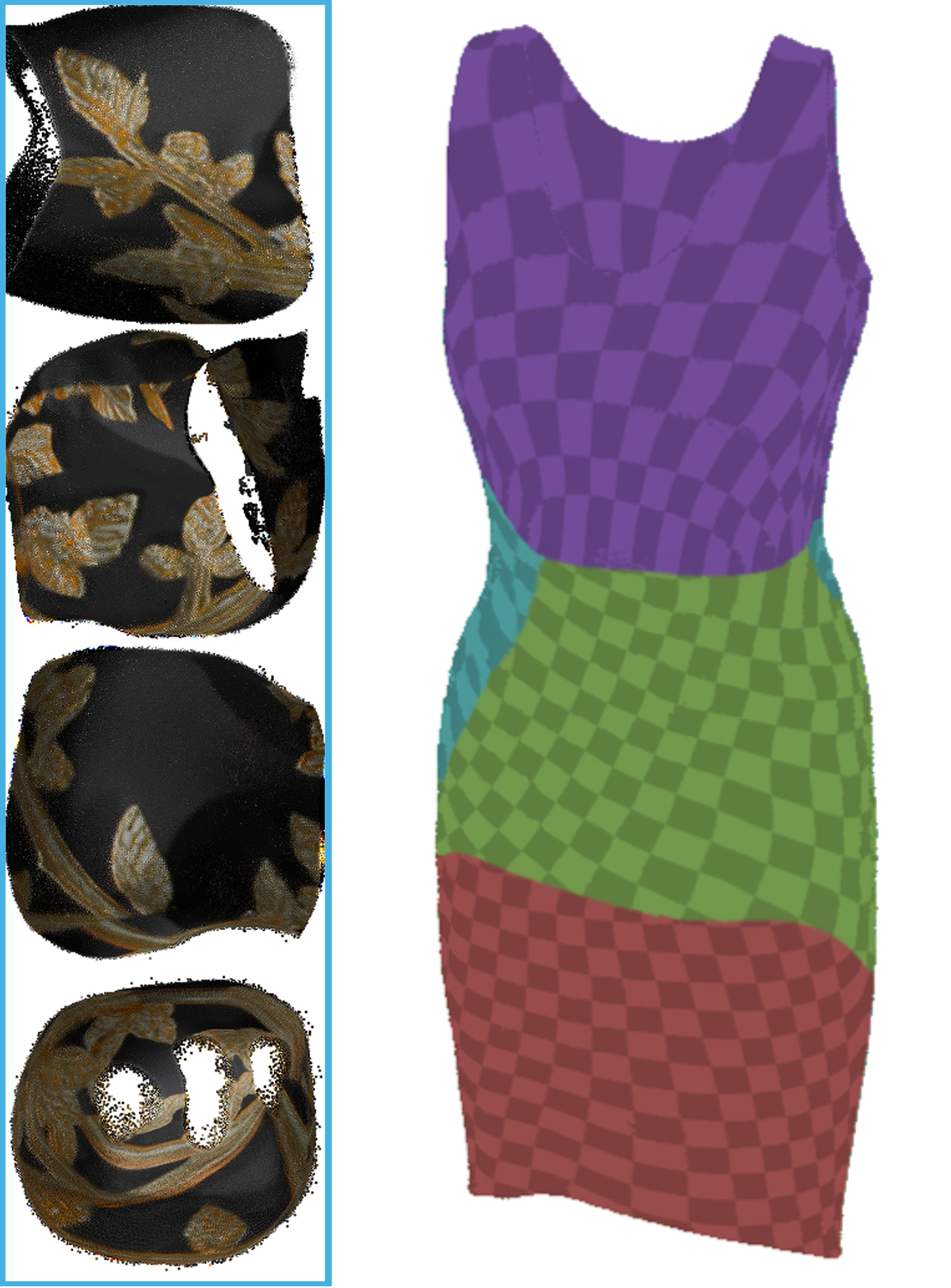}
    \end{tabular}
    }
    \caption{Learned UV mapping under different weights of the proposed distortion loss.}
    \label{fig:abla-texture-map-distortion-loss}
\end{figure}

\paragraph{On-surface regularizer.} Figure~\ref{fig:abla-sp-close-to-surface} shows the ablation of the on-surface loss (close-to-surface). We visualize a depth map from a selected view alongside a side-view of the learned point cloud. Without the on-surface loss, the point cloud contains many noisy, off-surface points. Adding the on-surface loss significantly reduces these noisy points, resulting in a much smoother depth map with fewer holes.

\subsubsection{Evaluation of Distortion Loss}

Figure~\ref{fig:abla-texture-map-distortion-loss} shows how the learned texture map depends on the weight of the proposed distortion loss in Sec.~\ref{sec:learning-uv-mapping}. Increasing the distortion-loss weight reduces UV distortion, which is crucial for preserving the shape of the edited texture.

\subsubsection{Correspondence Stability Under Deformation}
\label{sec:abla-correspondence-stability}
We ablate the edit-time canonical correspondence used for UV lookup after deformation (Sec.~\ref{sec:geometry-appearance-editing}). Specifically, we compare our deformation-aware canonical transfer (Eq.~\ref{eq:deformation-aware-correspondence}) against the naïve alternative that directly interpolates canonical neighbor positions using deformed-space attention weights (Eq.~\ref{eq:naive-canonical-sum}). Figure~\ref{fig:abla-correspondence-stability} shows that the naïve transfer leads to discontinuous UV jumps and visible texture drift/popping under non-rigid deformations, while our transfer better preserves correspondence and keeps texture edits attached.

\section{Conclusion and Limitation}
\label{sec:conclusion}

We presented PointGT, a representation based on PAPR that enables simultaneous editing of geometry and texture for point-based representations. PointGT successfully achieves this by learning a UV mapping which maps 3D points to a learned 2D texture map. To improve geometry accuracy, we introduced two novel geometry regularizers, which have been demonstrated to improve the geometry significantly. We also introduced a novel correspondence mechanism to maintain UV mapping under deformation, which has been demonstrated to be highly effective. Comparison with baselines showed that PointGT preserves texture edits under large deformations while the baseline exhibits artifacts and surface discontinuities. Furthermore, we showed that PointGT outperforms competitive baselines on novel view synthesis with a much smaller number of primitives. 
Our method currently relies on an optimization-based UV parameterization~\cite{Srinivasan2023NuvoNU}, which can struggle to produce clean and low-distortion charts for objects with complex topology, which we believe could be alleviated by learning-based UV-mapping methods~\cite{li2025auto,ma2026meshtailorcuttingseamsgenerative} in the future.

\section{Acknowledgments}

This research was enabled in part by support provided by NSERC, the BC DRI Group, the Digital Research Alliance of Canada and the Canada CIFAR AI Chairs Program.

\bibliographystyle{splncs04}
\bibliography{main}

\begin{thebibliography}{10}
\providecommand{\url}[1]{\texttt{#1}}
\providecommand{\urlprefix}{URL }
\providecommand{\doi}[1]{https://doi.org/#1}

\bibitem{Barron2021MipNeRF3U}
Barron, J.T., Mildenhall, B., Verbin, D., Srinivasan, P.P., Hedman, P.:
  Mip-nerf 360: Unbounded anti-aliased neural radiance fields. 2022 IEEE/CVF
  Conference on Computer Vision and Pattern Recognition (CVPR) pp. 5460--5469
  (2021)

\bibitem{chang2023pointersect}
Chang, J.H.R., Chen, W.Y., Ranjan, A., Yi, K.M., Tuzel, O.: Pointersect: Neural
  rendering with cloud-ray intersection. 2023 IEEE/CVF Conference on Computer
  Vision and Pattern Recognition (CVPR) pp. 8359--8369 (2023)

\bibitem{Chao2024TexturedGF}
Chao, B., Tseng, H.Y., Porzi, L., Gao, C., Li, T., Li, Q., Saraf, A., Huang,
  J.B., Kopf, J., Wetzstein, G., Kim, C.: Textured gaussians for enhanced 3d
  scene appearance modeling. 2025 IEEE/CVF Conference on Computer Vision and
  Pattern Recognition (CVPR) pp. 8964--8974 (2024)

\bibitem{Deitke2022ObjaverseAU}
Deitke, M., Schwenk, D., Salvador, J., Weihs, L., Michel, O., VanderBilt, E.,
  Schmidt, L., Ehsani, K., Kembhavi, A., Farhadi, A.: Objaverse: A universe of
  annotated 3d objects. 2023 IEEE/CVF Conference on Computer Vision and Pattern
  Recognition (CVPR) pp. 13142--13153 (2022)

\bibitem{Huang20242DGS}
Huang, B., Yu, Z., Chen, A., Geiger, A., Gao, S.: 2d gaussian splatting for
  geometrically accurate radiance fields. ACM SIGGRAPH 2024 Conference Papers
  (2024)

\bibitem{huang2023vbench}
Huang, Z., He, Y., Yu, J., Zhang, F., Si, C., Jiang, Y., Zhang, Y., Wu, T.,
  Jin, Q., Chanpaisit, N., Wang, Y., Chen, X., Wang, L., Lin, D., Qiao, Y.,
  Liu, Z.: {VBench}: Comprehensive benchmark suite for video generative models.
  In: CVPR (2024)

\bibitem{huang2025vbench++}
Huang, Z., Zhang, F., Xu, X., He, Y., Yu, J., Dong, Z., Ma, Q., Chanpaisit, N.,
  Si, C., Jiang, Y., Wang, Y., Chen, X., Chen, Y.C., Wang, L., Lin, D., Qiao,
  Y., Liu, Z.: {VBench++}: Comprehensive and versatile benchmark suite for
  video generative models. IEEE Transactions on Pattern Analysis and Machine
  Intelligence  (2025)

\bibitem{Jambon2023NeRFshop}
Jambon, C., Kerbl, B., Kopanas, G., Diolatzis, S., Leimk{\"u}hler, T.,
  Drettakis, G.: Nerfshop. Proceedings of the ACM on Computer Graphics and
  Interactive Techniques  \textbf{6},  1 -- 21 (2023)

\bibitem{Jensen2014LargeSM}
Jensen, R.R., Dahl, A., Vogiatzis, G., Tola, E., Aan{\ae}s, H.: Large scale
  multi-view stereopsis evaluation. 2014 IEEE Conference on Computer Vision and
  Pattern Recognition pp. 406--413 (2014)

\bibitem{Kerbl20233DGS}
Kerbl, B., Kopanas, G., Leimkuehler, T., Drettakis, G.: 3d gaussian splatting
  for real-time radiance field rendering. ACM Transactions on Graphics (TOG)
  \textbf{42},  1 -- 14 (2023)

\bibitem{li2025auto}
Li, Y., Cheung, V., Liu, X., Chen, Y., Luo, Z., Lei, B., Weng, H., Zhao, Z.,
  Huang, J., Chen, Z., et~al.: Auto-regressive surface cutting. arXiv preprint
  arXiv:2506.18017  (2025)

\bibitem{ma2026meshtailorcuttingseamsgenerative}
Ma, X., Yan, X., Zhang, C., Huang, H.: Meshtailor: Cutting seams via generative
  mesh traversal (2026)

\bibitem{mildenhall2021nerf}
Mildenhall, B., Srinivasan, P.P., Tancik, M., Barron, J.T., Ramamoorthi, R.,
  Ng, R.: {NeRF}: Representing scenes as neural radiance fields for view
  synthesis. Commun. ACM  \textbf{65},  99--106 (2020)

\bibitem{ost2022nplf}
Ost, J., Laradji, I., Newell, A., Bahat, Y., Heide, F.: Neural point light
  fields. In: Proceedings of the IEEE/CVF Conference on Computer Vision and
  Pattern Recognition. pp. 18419--18429 (2022)

\bibitem{Peng2022CageNeRFCN}
Peng, Y., Yan, Y., Liu, S., Cheng, Y., Guan, S., Pan, B., Zhai, G., Yang, X.:
  Cagenerf: Cage-based neural radiance field for generalized 3d deformation and
  animation. In: Neural Information Processing Systems (2022)

\bibitem{Rong2024GStexPT}
Rong, V., Chen, J., Bahmani, S., Kutulakos, K.N., Lindell, D.B.: Gstex:
  Per-primitive texturing of 2d gaussian splatting for decoupled appearance and
  geometry modeling. 2025 IEEE/CVF Winter Conference on Applications of
  Computer Vision (WACV) pp. 3508--3518 (2024)

\bibitem{Ronneberger2015UNetCN}
Ronneberger, O., Fischer, P., Brox, T.: U-net: Convolutional networks for
  biomedical image segmentation. ArXiv  \textbf{abs/1505.04597} (2015)

\bibitem{Song2024HDGST2}
Song, Y., Lin, H., Lei, J., Liu, L., Daniilidis, K.: Hdgs: Textured 2d gaussian
  splatting for enhanced scene rendering. ArXiv  \textbf{abs/2412.01823} (2024)

\bibitem{sorkine2007rigid}
Sorkine, O., Alexa, M.: As-rigid-as-possible surface modeling. In: Symposium on
  Geometry processing. vol.~4, pp. 109--116. Citeseer (2007)

\bibitem{Srinivasan2023NuvoNU}
Srinivasan, P.P., Garbin, S.J., Verbin, D., Barron, J.T., Mildenhall, B.: Nuvo:
  Neural uv mapping for unruly 3d representations. ArXiv
  \textbf{abs/2312.05283} (2023)

\bibitem{Sumner2007EmbeddedDF}
Sumner, R.W., Schmid, J., Pauly, M.: Embedded deformation for shape
  manipulation. ACM SIGGRAPH 2007 papers  (2007)

\bibitem{Svitov2024BillBoardS}
Svitov, D., Morerio, P., de~Agapito, L., Bue, A.D.: Billboard splatting
  (bbsplat): Learnable textured primitives for novel view synthesis. ArXiv
  \textbf{abs/2411.08508} (2024)

\bibitem{Tang2025TexGSVolVisES}
Tang, K., Ai, K., Han, J., Wang, C.: Texgs-volvis: Expressive scene editing for
  volume visualization via textured gaussian splatting. ArXiv
  \textbf{abs/2507.13586} (2025)

\bibitem{Wang2023MeshGuidedNI}
Wang, C., He, M., Chai, M., Chen, D., Liao, J.: Mesh-guided neural implicit
  field editing. ArXiv  \textbf{abs/2312.02157} (2023)

\bibitem{Xiang2021NeuTexNT}
Xiang, F., Xu, Z., Havsan, M., Hold-Geoffroy, Y., Sunkavalli, K., Su, H.:
  Neutex: Neural texture mapping for volumetric neural rendering. 2021 IEEE/CVF
  Conference on Computer Vision and Pattern Recognition (CVPR) pp. 7115--7124
  (2021)

\bibitem{xu2022cagedeforming}
Xu, T., Harada, T.: Deforming radiance fields with cages. ArXiv
  \textbf{abs/2207.12298} (2022)

\bibitem{Xu2024TextureGSDT}
Xu, T., Hu, W., Lai, Y.K., Shan, Y., Zhang, S.: Texture-gs: Disentangling the
  geometry and texture for 3d gaussian splatting editing. In: European
  Conference on Computer Vision (2024)

\bibitem{Yang2022NeuMeshLD}
Yang, B., Bao, C., Zeng, J., Bao, H., Zhang, Y., Cui, Z., Zhang, G.: Neumesh:
  Learning disentangled neural mesh-based implicit field for geometry and
  texture editing. ArXiv  \textbf{abs/2207.11911} (2022)

\bibitem{yuan2022nerfedit}
Yuan, Y.J., Sun, Y.T., Lai, Y.K., Ma, Y., Jia, R., Gao, L.: Nerf-editing:
  Geometry editing of neural radiance fields. 2022 IEEE/CVF Conference on
  Computer Vision and Pattern Recognition (CVPR) pp. 18332--18343 (2022)

\bibitem{Zhang2025NeuralST}
Zhang, X., Chen, A., Xiong, J., Dai, P., Shen, Y., Xu, W.: Neural shell texture
  splatting: More details and fewer primitives. ArXiv  \textbf{abs/2507.20200}
  (2025)

\bibitem{zhang2023papr}
Zhang, Y., Peng, S., Moazenipourasil, S.A., Li, K.: {PAPR}: Proximity attention
  point rendering. In: Thirty-seventh Conference on Neural Information
  Processing Systems (2023)

\bibitem{Zheng2022EditableNeRFET}
Zheng, C., chang Lin, W., Xu, F.: Editablenerf: Editing topologically varying
  neural radiance fields by key points. 2023 IEEE/CVF Conference on Computer
  Vision and Pattern Recognition (CVPR) pp. 8317--8327 (2022)

\bibitem{zheng2025vbench2}
Zheng, D., Huang, Z., Liu, H., Zou, K., He, Y., Zhang, F., Zhang, Y., He, J.,
  Zheng, W.S., Qiao, Y., Liu, Z.: {VBench-2.0}: Advancing video generation
  benchmark suite for intrinsic faithfulness. arXiv preprint arXiv:2503.21755
  (2025)

\bibitem{Zhou2023SERFFI}
Zhou, K., Hong, L., Xie, E., Yang, Y., Li, Z., Zhang, W.: Serf: Fine-grained
  interactive 3d segmentation and editing with radiance fields. ArXiv
  \textbf{abs/2312.15856} (2023)

\end{thebibliography}

\newpage
\appendix

\section{Multi-View Editing Results}
\label{supp-sec:multiview-editing}

\begin{figure}[H]
\centering
\setlength\tabcolsep{1.5pt}
\renewcommand{\arraystretch}{1.0}
\scriptsize
\begin{tabularx}{\linewidth}{l@{\hskip 0.3em}YYYYY}
\rotatebox[origin=c]{90}{Emperor Fish}
    & \includegraphics[width=\hsize,valign=m]{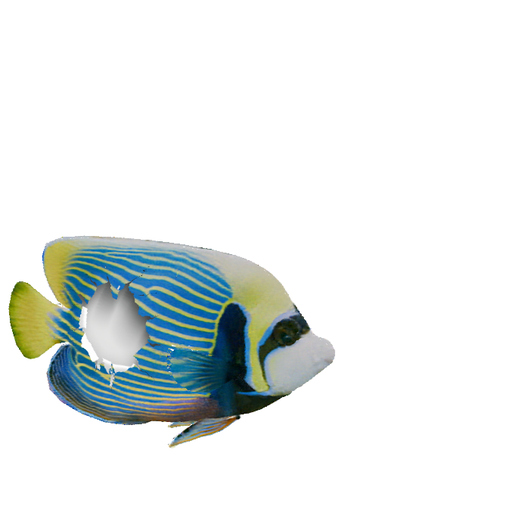}
    & \includegraphics[width=\hsize,valign=m]{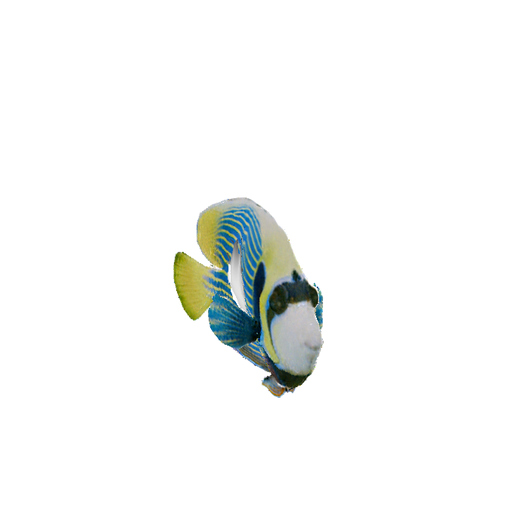}
    & \includegraphics[width=\hsize,valign=m]{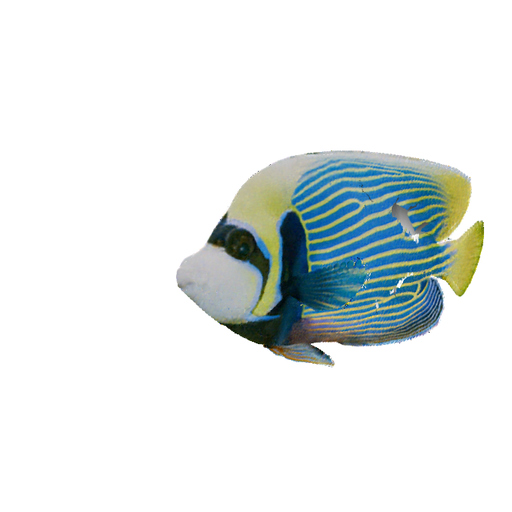}
    & \includegraphics[width=\hsize,valign=m]{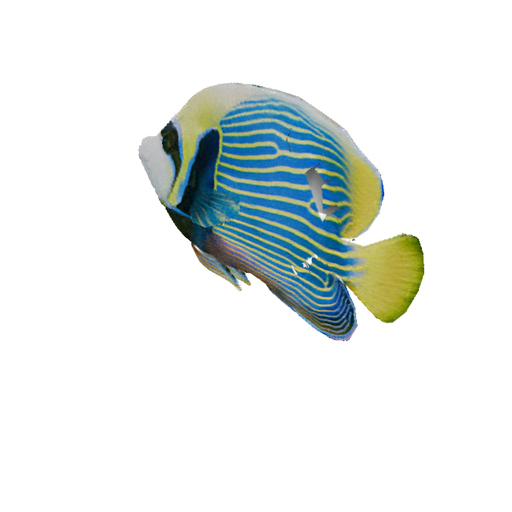}
    & \includegraphics[width=\hsize,valign=m]{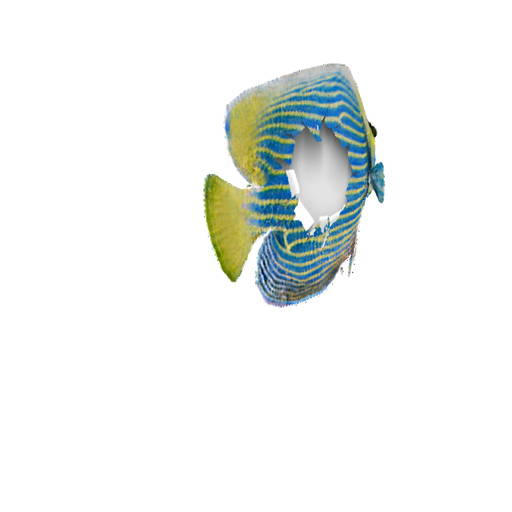} \\
\rotatebox[origin=c]{90}{Blossom}
    & \includegraphics[width=\hsize,valign=m]{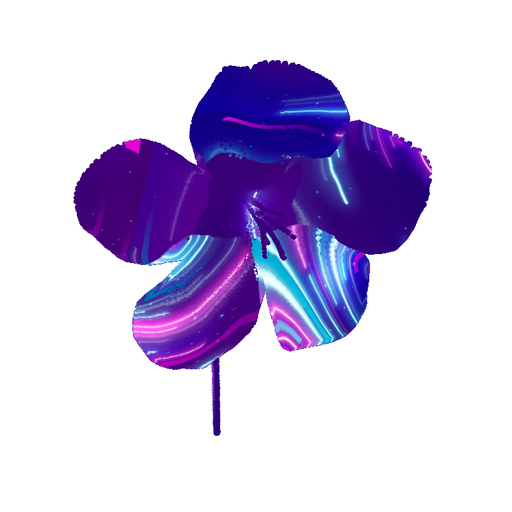}
    & \includegraphics[width=\hsize,valign=m]{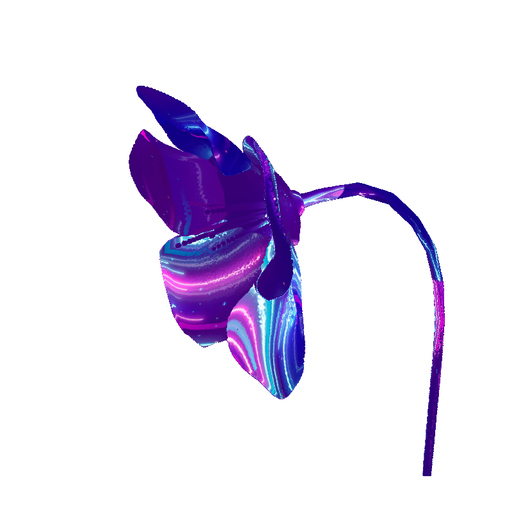}
    & \includegraphics[width=\hsize,valign=m]{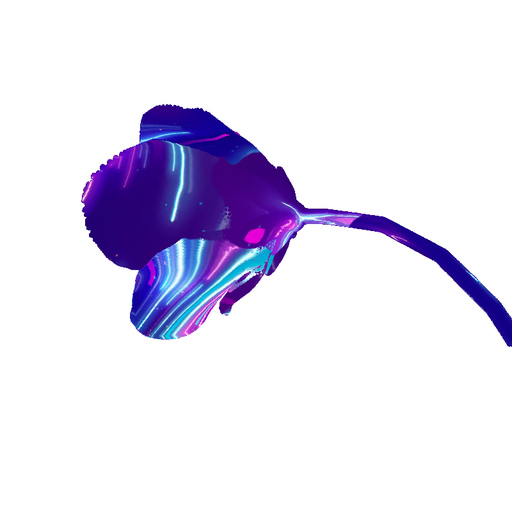}
    & \includegraphics[width=\hsize,valign=m]{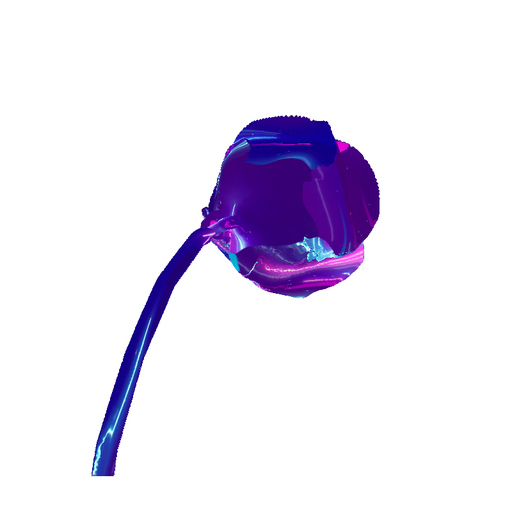}
    & \includegraphics[width=\hsize,valign=m]{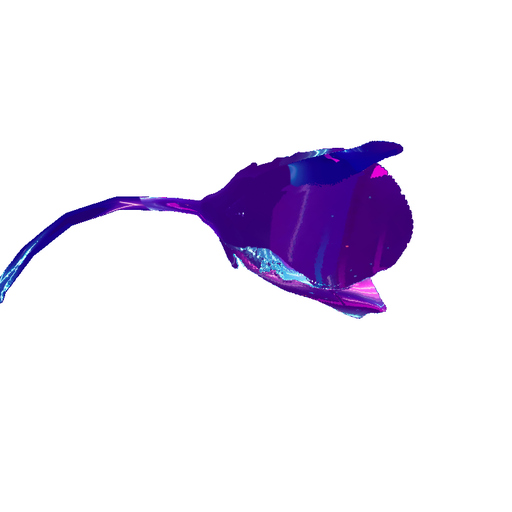} \\
\rotatebox[origin=c]{90}{Dress}
    & \includegraphics[width=\hsize,valign=m]{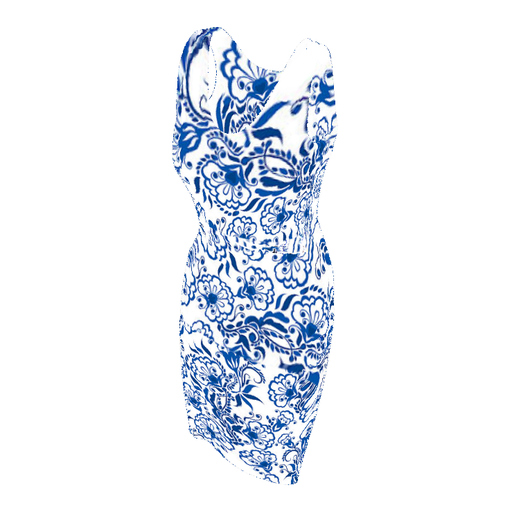}
    & \includegraphics[width=\hsize,valign=m]{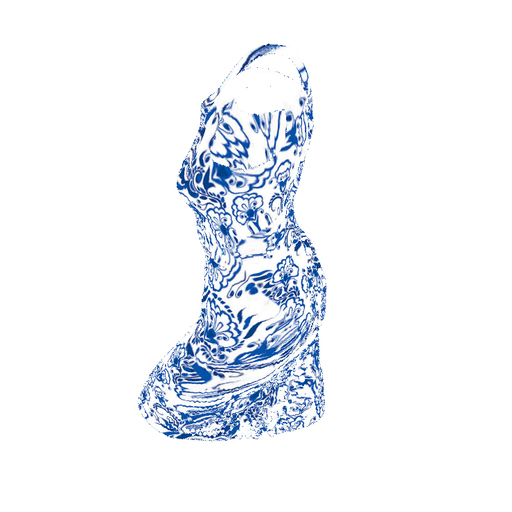}
    & \includegraphics[width=\hsize,valign=m]{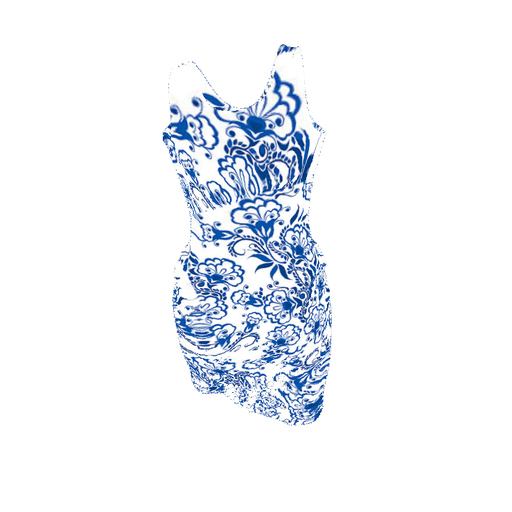}
    & \includegraphics[width=\hsize,valign=m]{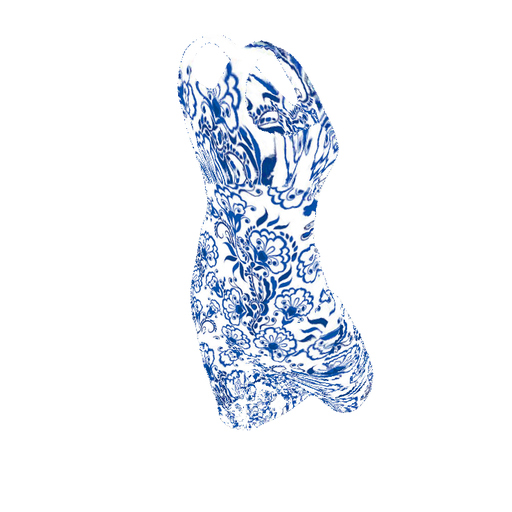}
    & \includegraphics[width=\hsize,valign=m]{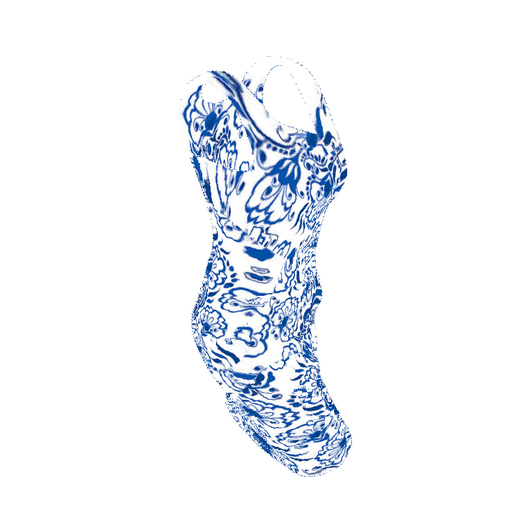} \\
\rotatebox[origin=c]{90}{Flag}
    & \includegraphics[width=\hsize,valign=m]{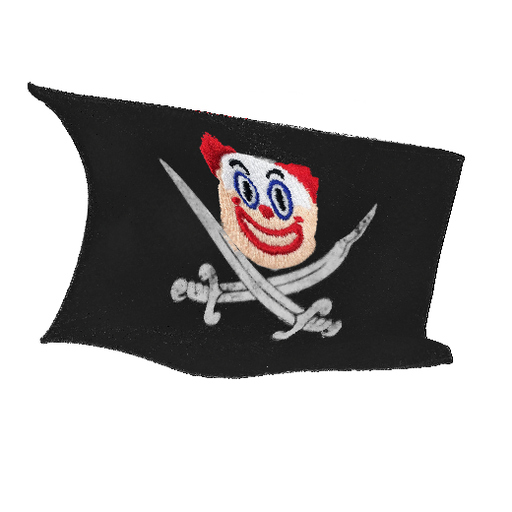}
    & \includegraphics[width=\hsize,valign=m]{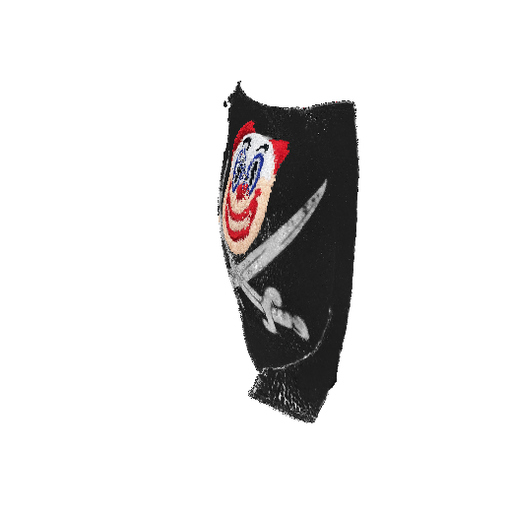}
    & \includegraphics[width=\hsize,valign=m]{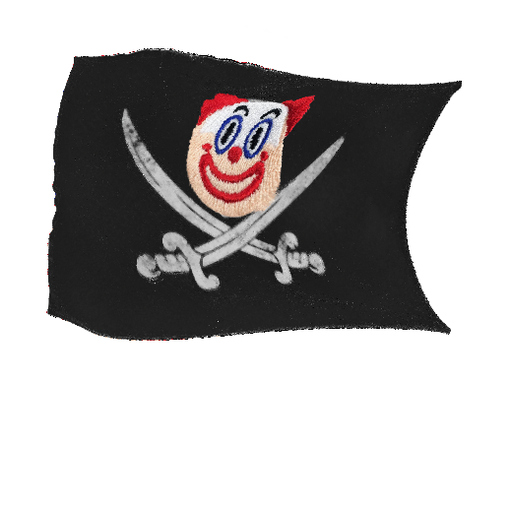}
    & \includegraphics[width=\hsize,valign=m]{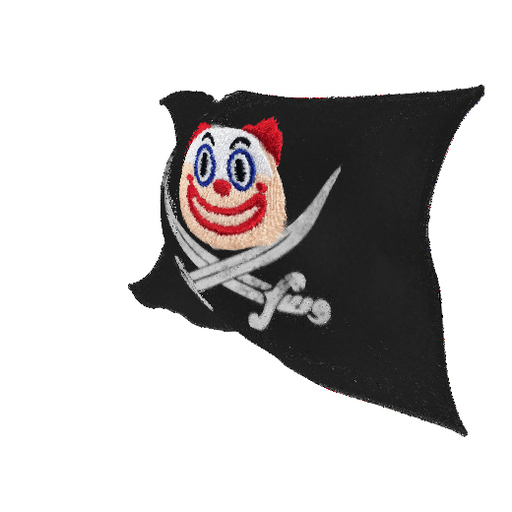}
    & \includegraphics[width=\hsize,valign=m]{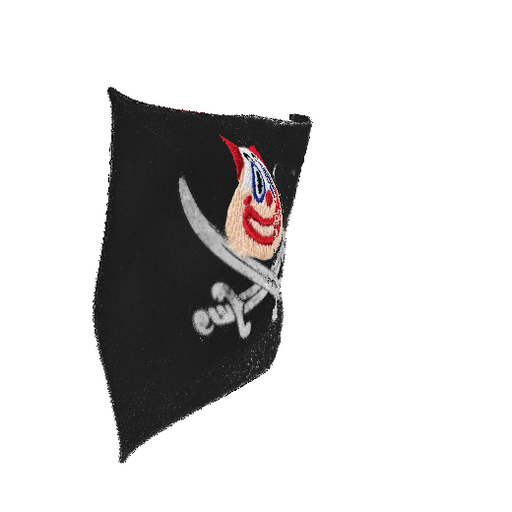} \\
\rotatebox[origin=c]{90}{Rexy}
    & \includegraphics[width=\hsize,valign=m]{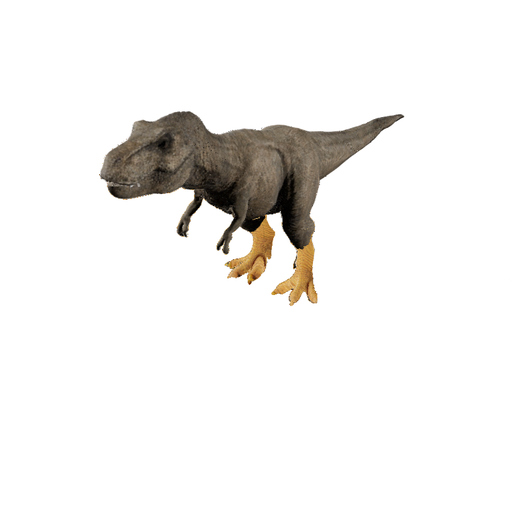}
    & \includegraphics[width=\hsize,valign=m]{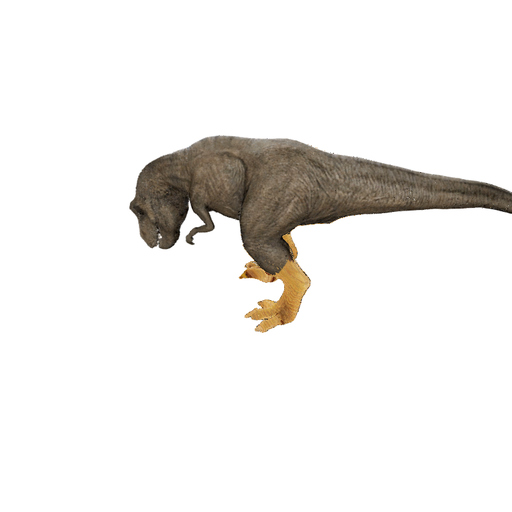}
    & \includegraphics[width=\hsize,valign=m]{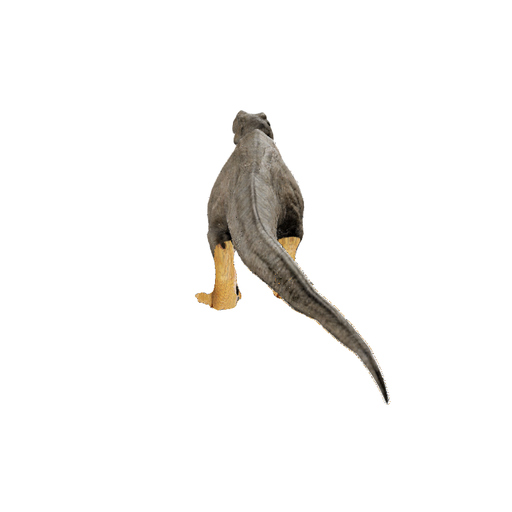}
    & \includegraphics[width=\hsize,valign=m]{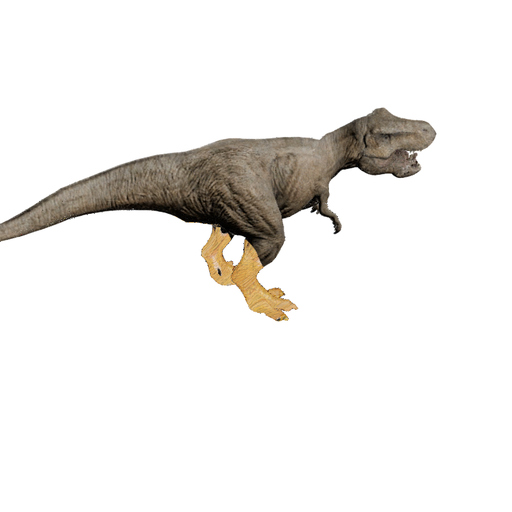}
    & \includegraphics[width=\hsize,valign=m]{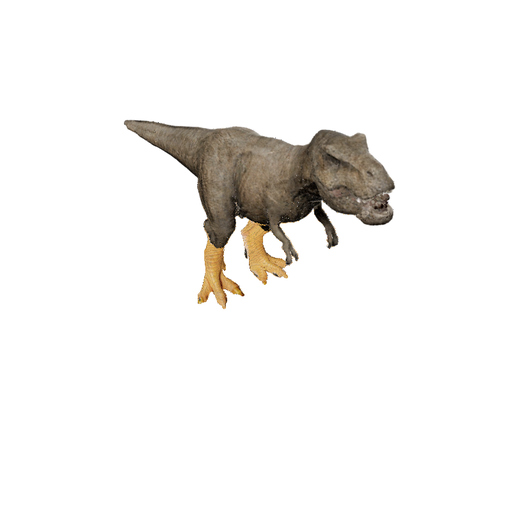} \\
\end{tabularx}
\caption{
    Novel views of our simultaneous geometry and texture editing results. For each scene, we render a $360^\circ$ orbit of the edited and non-rigidly deformed object and show five evenly spaced viewpoints. The edited texture stays sharp and remains consistently attached to the deforming surface across all views.
}
\label{supp-fig:multiview}
\end{figure}

We show our simultaneous geometry and texture editing results rendered from multiple viewpoints. For each of the five Objaverse scenes in Figure~\ref{fig:simul-edit-large-deform}, we render a $360^\circ$ orbit of the edited and non-rigidly deformed object---the same orbit renderings used for the VBench evaluation in Table~\ref{tab:vbench}---and show five evenly spaced views in Figure~\ref{supp-fig:multiview}. Across all viewpoints, the edited texture stays sharp and remains firmly attached to the deformed surface, with no visible seams, drifting, or loss of resolution, confirming that PointGT preserves both view consistency and edit attachment under large non-rigid deformation.

\subsection{Part Removal and Duplication}
\label{supp-sec:remove-duplicate}

\begin{figure}[htb]
\centering
\setlength\tabcolsep{2pt}
\renewcommand{\arraystretch}{1.0}
\scriptsize
\begin{tabular}{@{}c@{\hskip 3pt}cc@{}}
 & UV mapping & Rendering and texture map \\
\rotatebox[origin=c]{90}{Step 1} &
    \includegraphics[height=3cm,valign=m]{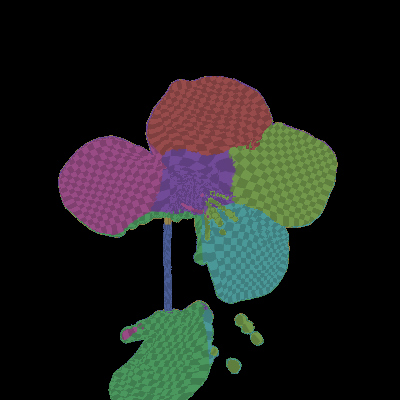} &
    \includegraphics[height=3cm,valign=m]{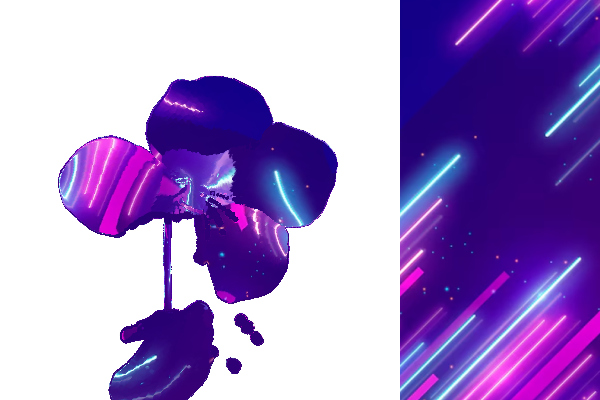} \\
\rotatebox[origin=c]{90}{Step 2} &
    \includegraphics[height=3cm,valign=m]{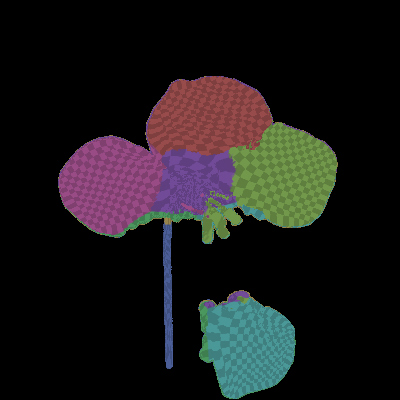} &
    \includegraphics[height=3cm,valign=m]{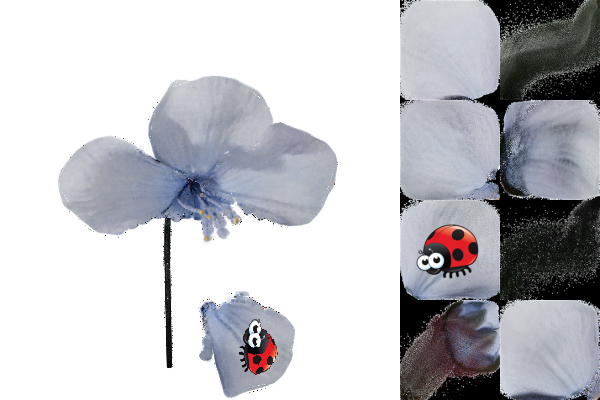} \\
\rotatebox[origin=c]{90}{Step 3} &
    \includegraphics[height=3cm,valign=m]{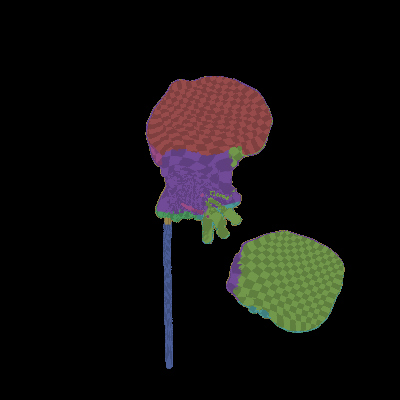} &
    \includegraphics[height=3cm,valign=m]{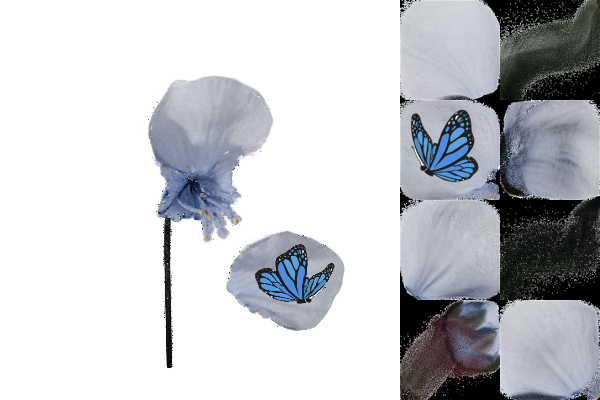} \\
\end{tabular}
\caption{
    \textbf{Part removal} on the Blossom scene. From top to bottom, petals are sequentially removed. \emph{UV mapping}: checkerboard visualization of the learned charts on the deforming geometry. \emph{Rendering and texture map}: the rendered result (left) and its 2D texture map (right); the applied texture is switched across the sequence to show that the remaining geometry stays editable.
}
\label{supp-fig:removal}
\end{figure}

\begin{figure}[htb]
\centering
\setlength\tabcolsep{2pt}
\renewcommand{\arraystretch}{1.0}
\scriptsize
\begin{tabular}{@{}c@{\hskip 3pt}ccc@{}}
 & UV mapping & Edited texture & Texture map \\
\rotatebox[origin=c]{90}{Edit 1} &
    \includegraphics[height=3cm,valign=m]{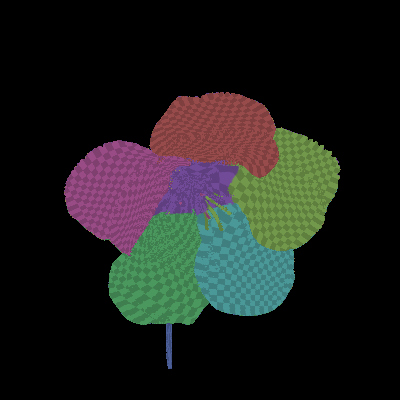} &
    \includegraphics[height=3cm,valign=m]{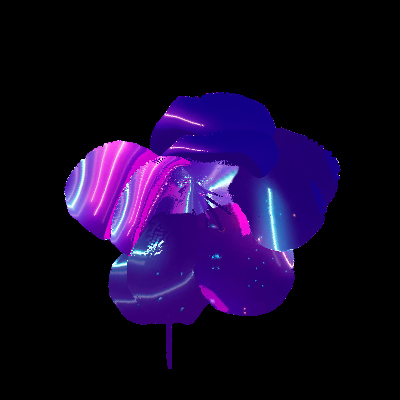} &
    \includegraphics[height=3cm,valign=m]{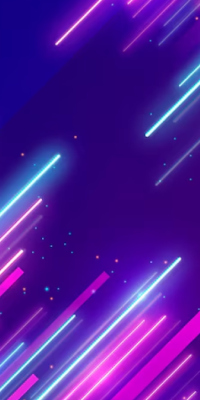} \\
\rotatebox[origin=c]{90}{Edit 2} &
    \includegraphics[height=3cm,valign=m]{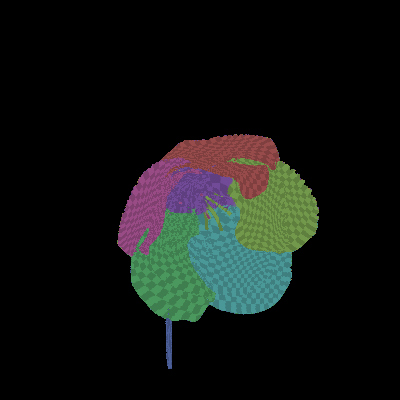} &
    \includegraphics[height=3cm,valign=m]{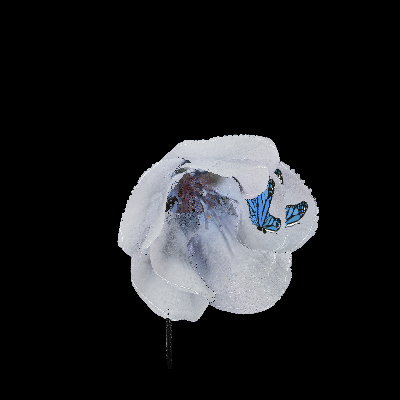} &
    \includegraphics[height=3cm,valign=m]{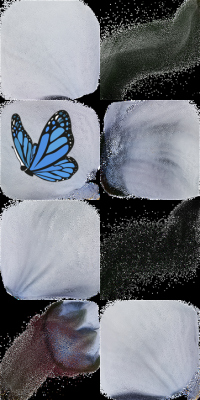} \\
\end{tabular}
\caption{
    \textbf{Part duplication} on the Blossom scene. The duplicated geometry is shown under two different texture edits. \emph{UV mapping}: checkerboard visualization of the learned charts; \emph{Edited texture}: the rendered result; \emph{Texture map}: the corresponding 2D texture map. The texture stays correctly attached to the duplicated parts through the canonical UV mapping.
}
\label{supp-fig:duplication}
\end{figure}

Because PointGT decouples geometry and appearance through a canonical UV mapping, it also supports editing operations that change the object's topology, while keeping the edited texture correctly attached to the modified geometry. We demonstrate this on the Blossom scene with two operations. Figure~\ref{supp-fig:removal} shows \emph{part removal}: petals are sequentially detached from the flower (top to bottom). The learned UV mapping (left, visualized as a checkerboard with one color per chart) and the rendered result together with its 2D texture map (right) stay consistent throughout; since appearance is decoupled, the geometry can be retextured at the same time, and we switch the applied texture across the sequence. Figure~\ref{supp-fig:duplication} shows \emph{part duplication}, where the duplicated geometry is rendered under two different texture edits, with the texture remaining correctly mapped onto the duplicated parts through the canonical UV mapping. 

\section{Additional Ablation Results}

\begin{figure*}[htb]
\centering
\setlength{\tabcolsep}{1pt}
\renewcommand{\arraystretch}{1.05}
\scriptsize
\begin{tabularx}{\linewidth}{Y@{\hspace{0.015\linewidth}}YY@{\hspace{0.015\linewidth}}YY@{\hspace{0.015\linewidth}}YY@{}}
edited scene &
\multicolumn{2}{c}{(1) baseline} &
\multicolumn{2}{c}{(2) + on-ray regularizer} &
\multicolumn{2}{c}{(3) + deform-aware corr.} \\
\includegraphics[width=\linewidth]{figs/main-results/cropped_blossom_edited_v2_fog_ours.jpg} &
\includegraphics[width=\linewidth]{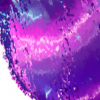} &
\includegraphics[width=\linewidth]{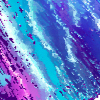} &
\includegraphics[width=\linewidth]{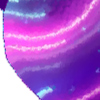} &
\includegraphics[width=\linewidth]{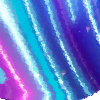} &
\includegraphics[width=\linewidth]{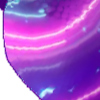} &
\includegraphics[width=\linewidth]{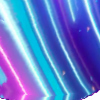} \\
\includegraphics[width=\linewidth]{figs/main-results/cropped_dress_edited_v16_fog_ours.jpg} &
\includegraphics[width=\linewidth]{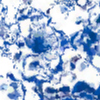} &
\includegraphics[width=\linewidth]{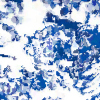} &
\includegraphics[width=\linewidth]{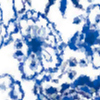} &
\includegraphics[width=\linewidth]{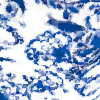} &
\includegraphics[width=\linewidth]{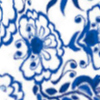} &
\includegraphics[width=\linewidth]{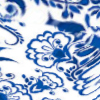} \\
\includegraphics[width=\linewidth]{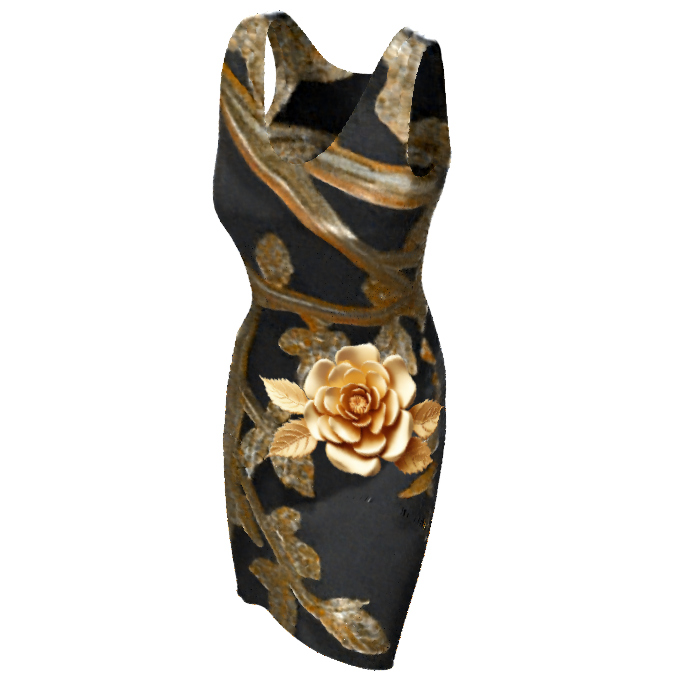} &
\includegraphics[width=\linewidth]{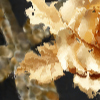} &
\includegraphics[width=\linewidth]{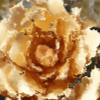} &
\includegraphics[width=\linewidth]{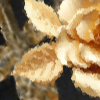} &
\includegraphics[width=\linewidth]{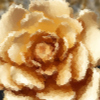} &
\includegraphics[width=\linewidth]{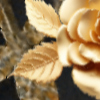} &
\includegraphics[width=\linewidth]{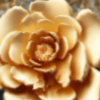}
\end{tabularx}
\caption{Additional texture editing results. We show both edited scene and close-up views of two different regions of the scene. As shown, the proposed on-ray regularizer and deformation-aware correspondence maintain the high-fidelity.}
\label{fig:qual-ablation}
\end{figure*}

We show ablation results for the proposed on-ray regularizer~\ref{eq:on-ray-reg} and the deformation-aware correspondence~\ref{eq:deformation-aware-correspondence} in Figure~\ref{fig:qual-ablation}, in addition to Figure~\ref{fig:abla-correspondence-stability}. As shown, the proposed on-ray regularizer significantly reduces the artifacts of the texture editing, and the deformation-aware correspondence effectively boosts the editing fidelity.

\section{Discussion of Other Gaussian Splatting Baselines}

As noted in the main paper, we choose GSTex~\cite{Rong2024GStexPT} as the primary baseline for simultaneous geometry and texture editing. Here we provide additional discussion and evidence for why the other baselines are limited in their ability to perform simultaneous geometry and texture editing.

\begin{figure}[htb]
\centering
\setlength\tabcolsep{1pt}
\renewcommand{\arraystretch}{2}
\tiny
\newlength{\figheigh}
\setlength{\figheigh}{3cm}
\begin{tabularx}{0.7\linewidth}{l@{\hskip 0.2em}YYY}
\rotatebox[origin=c]{90}{Flag}
    &   \includegraphics[height=\figheigh, width=\hsize, keepaspectratio, valign=m]{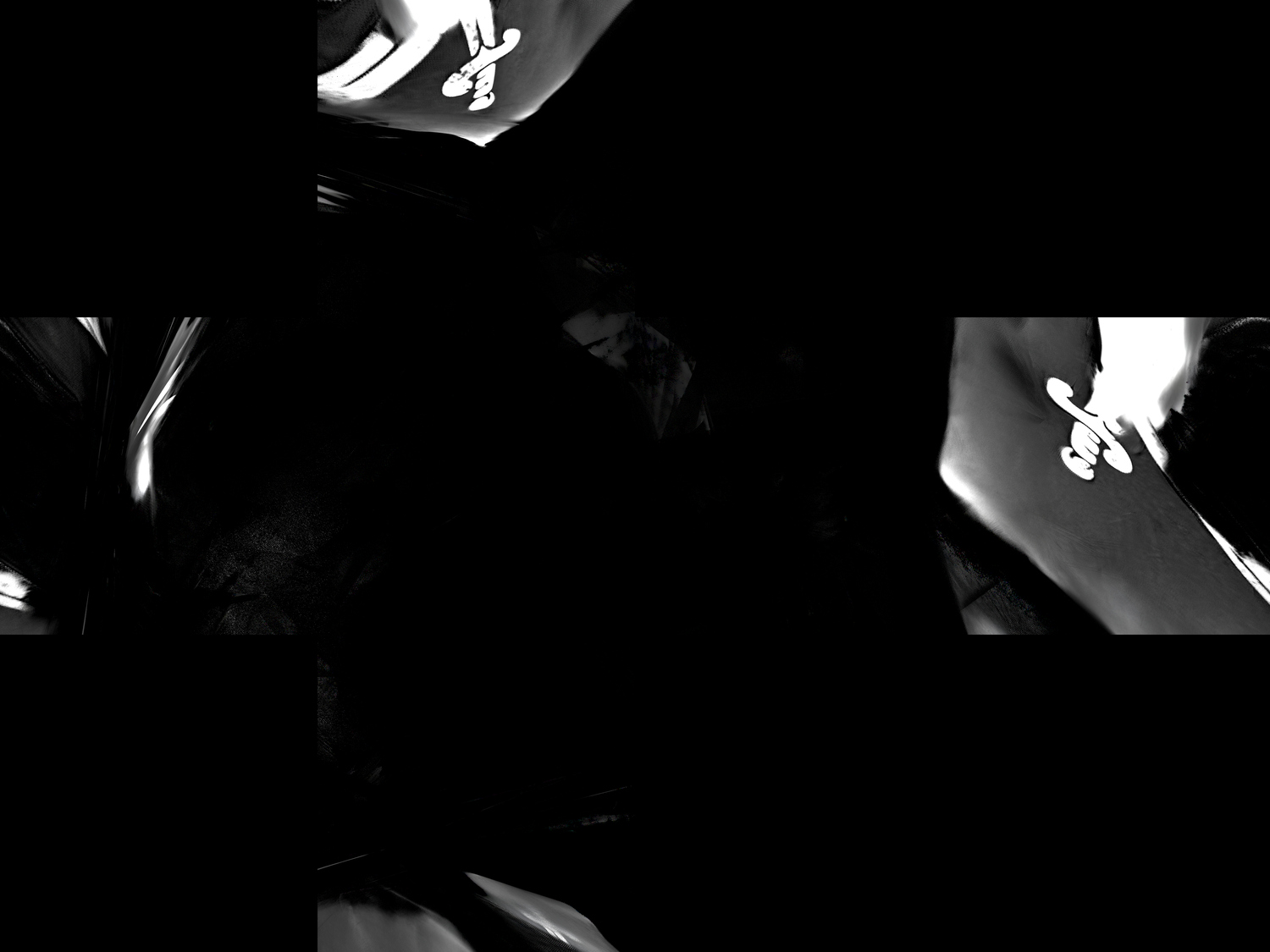}
    &   \includegraphics[height=\figheigh, width=\hsize, keepaspectratio, valign=m]{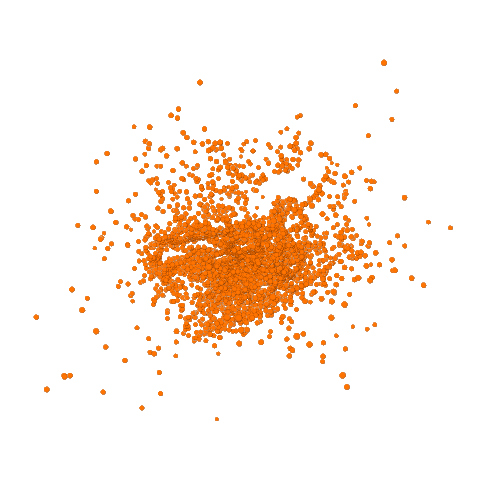}
    &   \includegraphics[height=\figheigh, width=\hsize, keepaspectratio, valign=m]{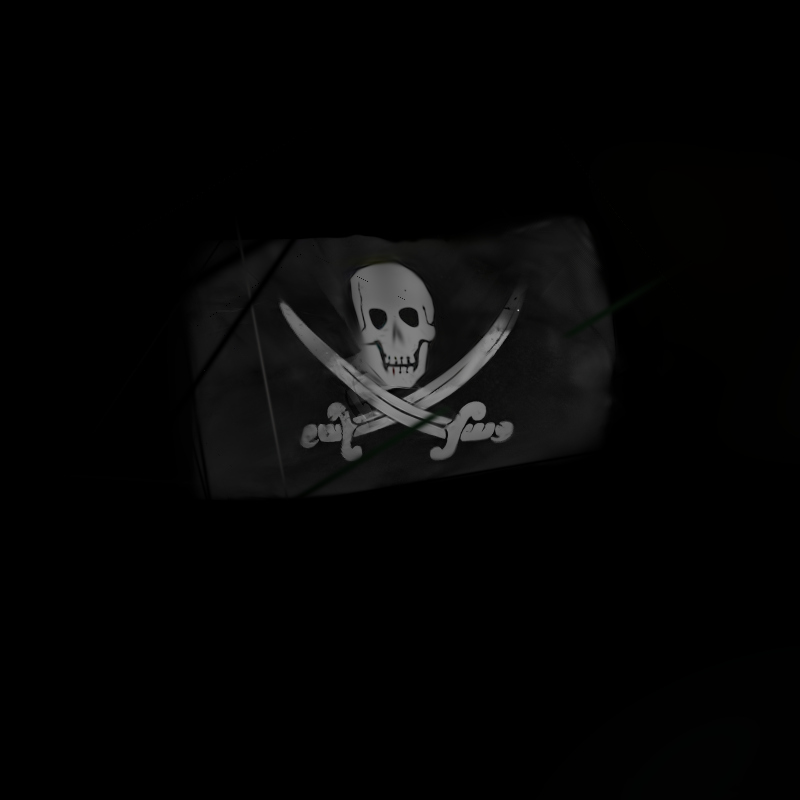}
    \\
\rotatebox[origin=c]{90}{Dress}
    &   \includegraphics[height=\figheigh, width=\hsize, keepaspectratio, valign=m]{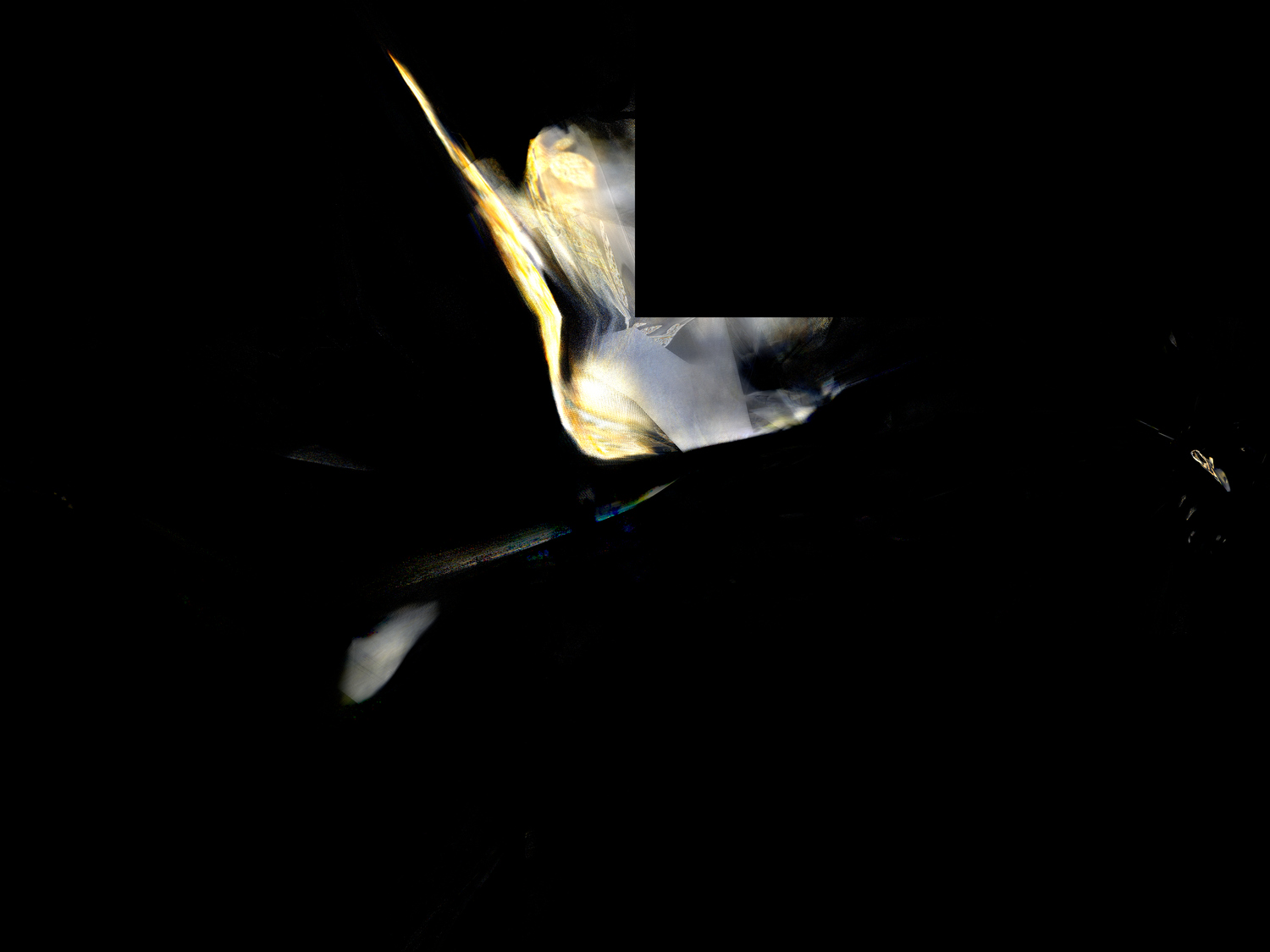}
    &   \includegraphics[height=\figheigh, width=\hsize, keepaspectratio, valign=m]{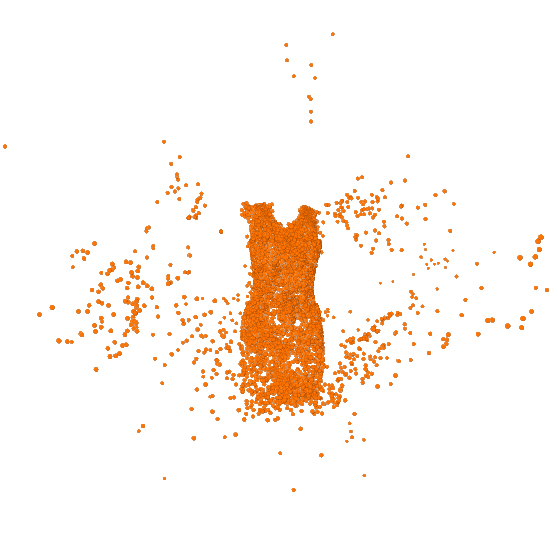}
    &   \includegraphics[height=\figheigh, width=\hsize, keepaspectratio, valign=m]{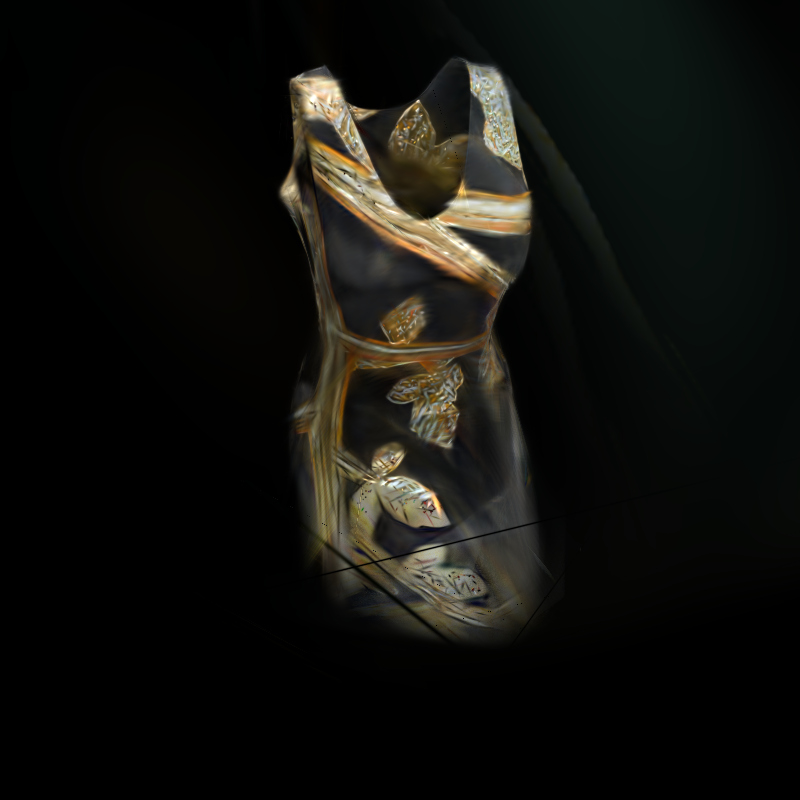}
    \\
\rotatebox[origin=c]{90}{Bird}
    &   \includegraphics[height=\figheigh, width=\hsize, keepaspectratio, valign=m]{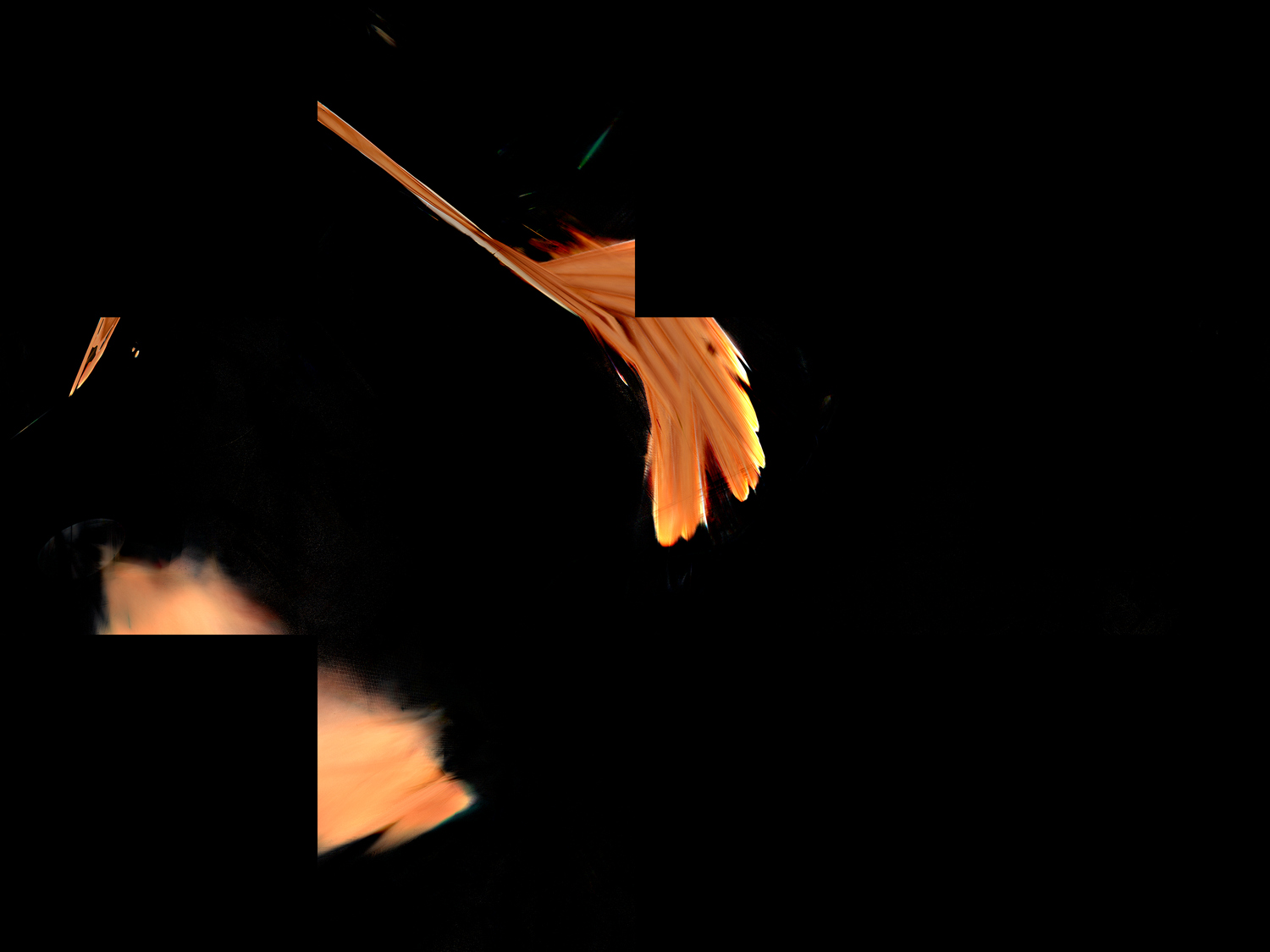}
    &   \includegraphics[height=\figheigh, width=\hsize, keepaspectratio, valign=m]{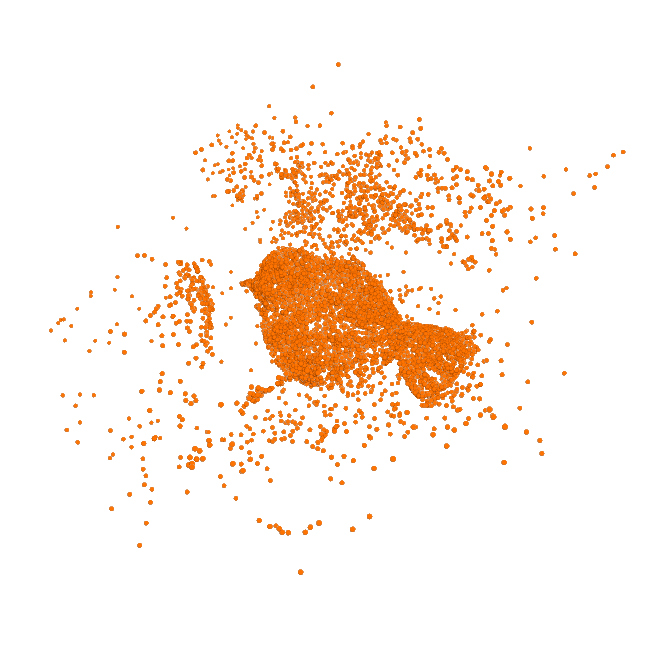}
    &   \includegraphics[height=\figheigh, width=\hsize, keepaspectratio, valign=m]{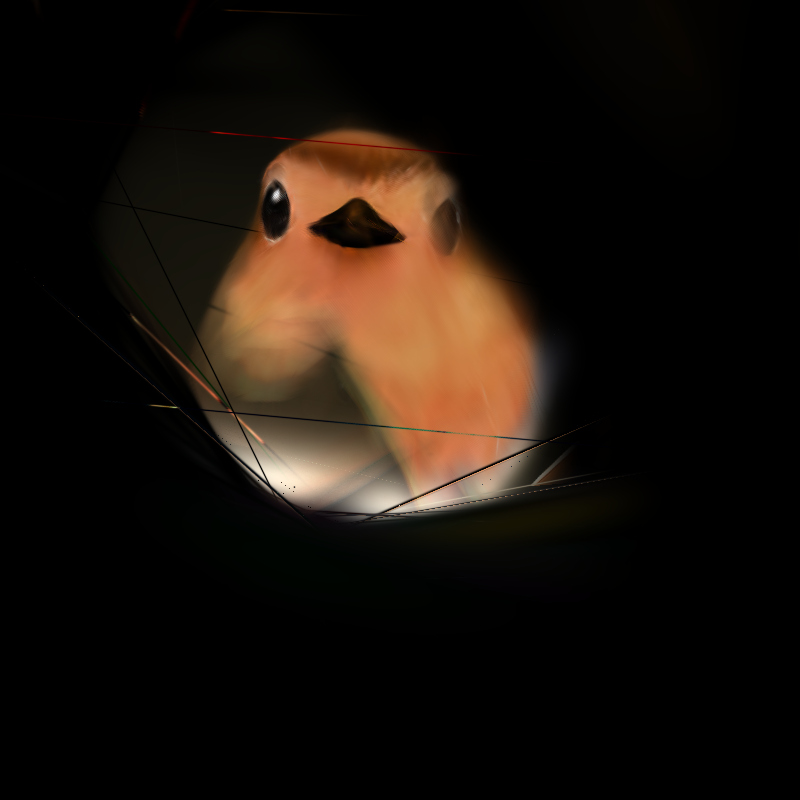}
    \\
    &   Learned Texture Map from Texture-GS
    &   Learned Gaussians
    &   Rendering
\end{tabularx}
\caption{
    Texture-GS~\cite{Xu2024TextureGSDT} may fail to learn a reasonable texture map that is valid for texture editing.
}
\label{supp-fig:texture-gs-failure}
\end{figure}

\begin{figure}[htb]
\centering
\setlength\tabcolsep{1pt}
\renewcommand{\arraystretch}{2}
\tiny
\newlength{\figheight}
\setlength{\figheight}{2.2cm}
\begin{tabularx}{1.0\linewidth}{YYYYYY}
    \includegraphics[height=\figheight, width=\hsize, keepaspectratio, valign=m]{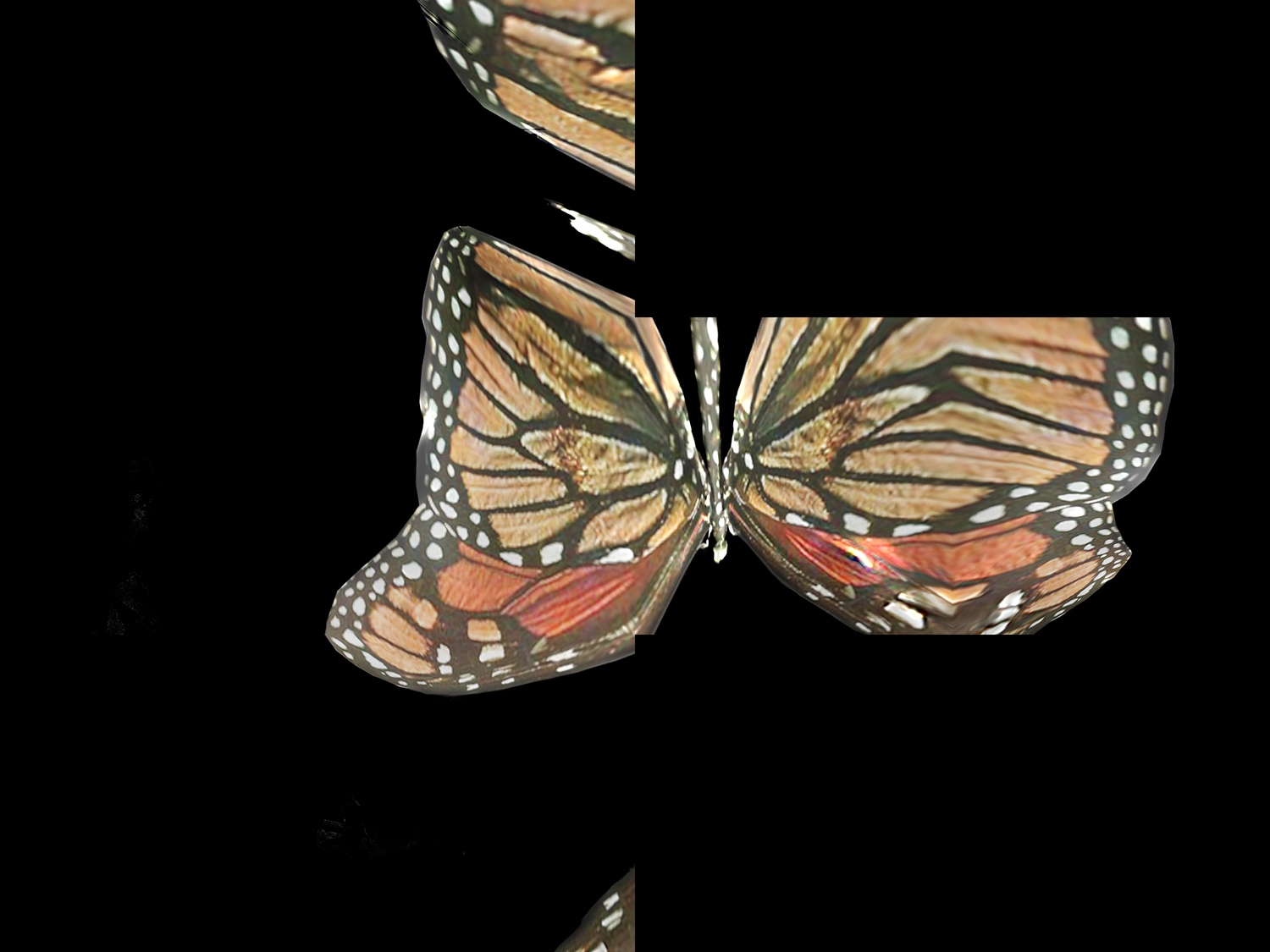}
    &   \includegraphics[height=\figheight, width=\hsize, keepaspectratio, valign=m]{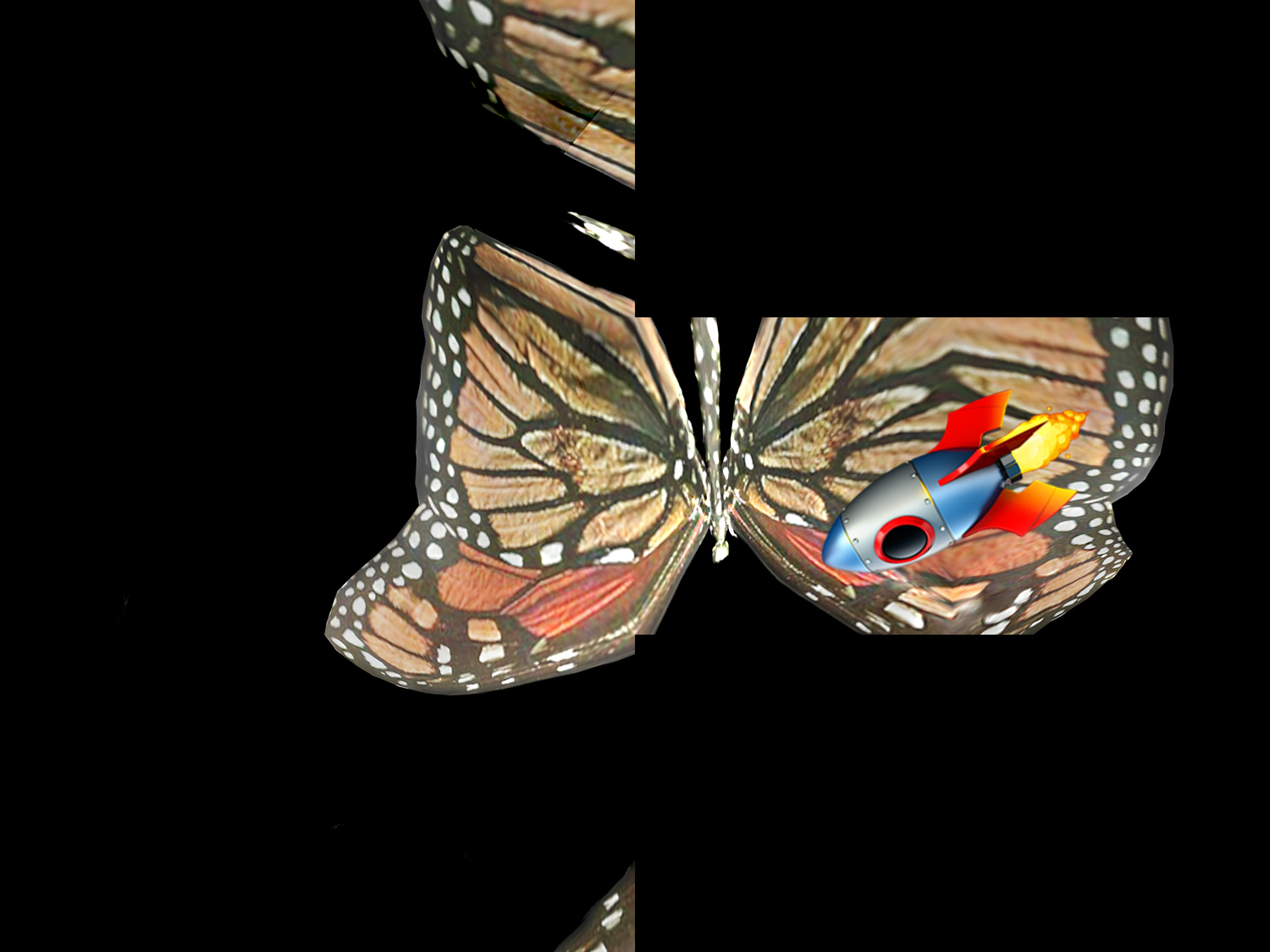}
    &   \includegraphics[height=\figheight, width=\hsize, keepaspectratio, valign=m]{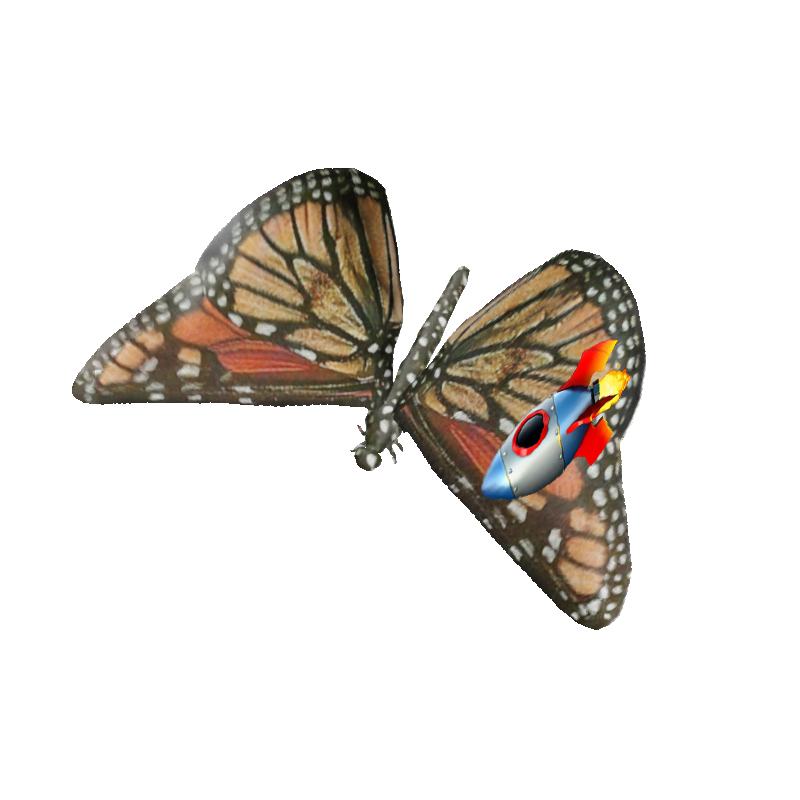}
    &   \includegraphics[height=\figheight, width=\hsize, keepaspectratio, valign=m]{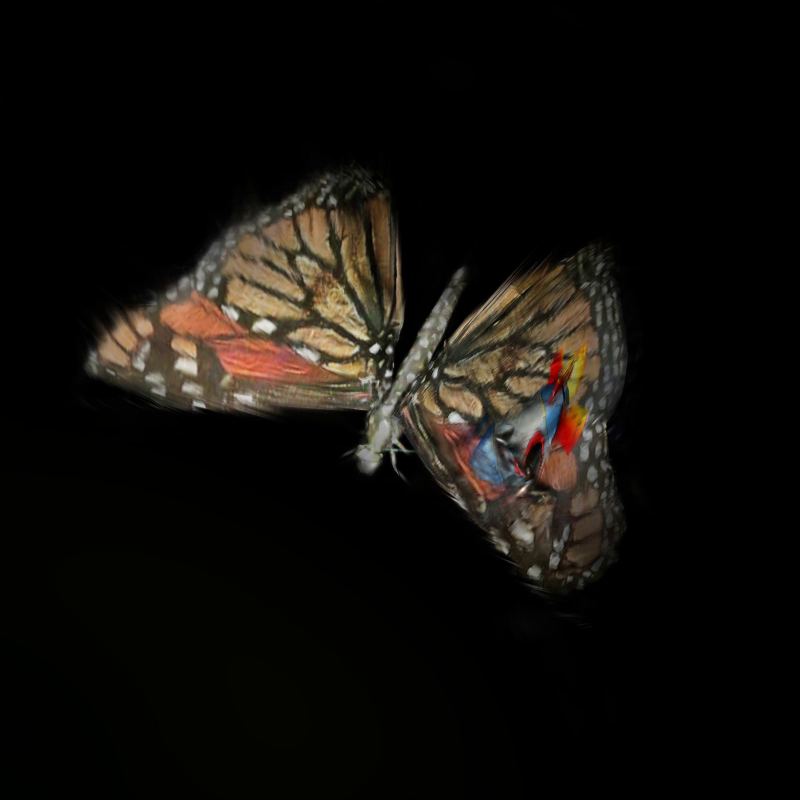}
    &   \includegraphics[height=\figheight, width=\hsize, keepaspectratio, valign=m]{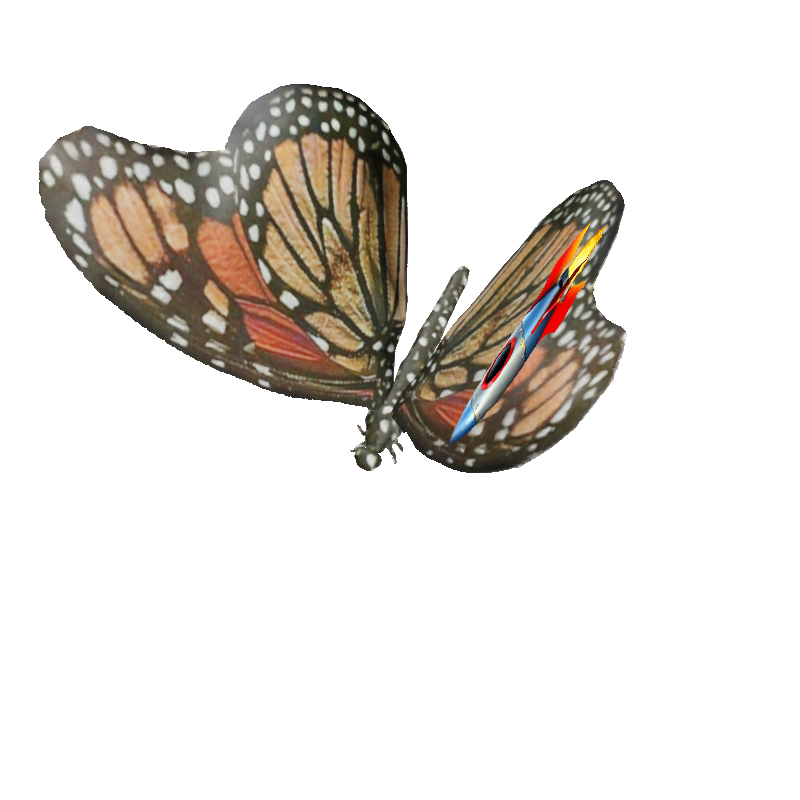}
    &   \includegraphics[height=\figheight, width=\hsize, keepaspectratio, valign=m]{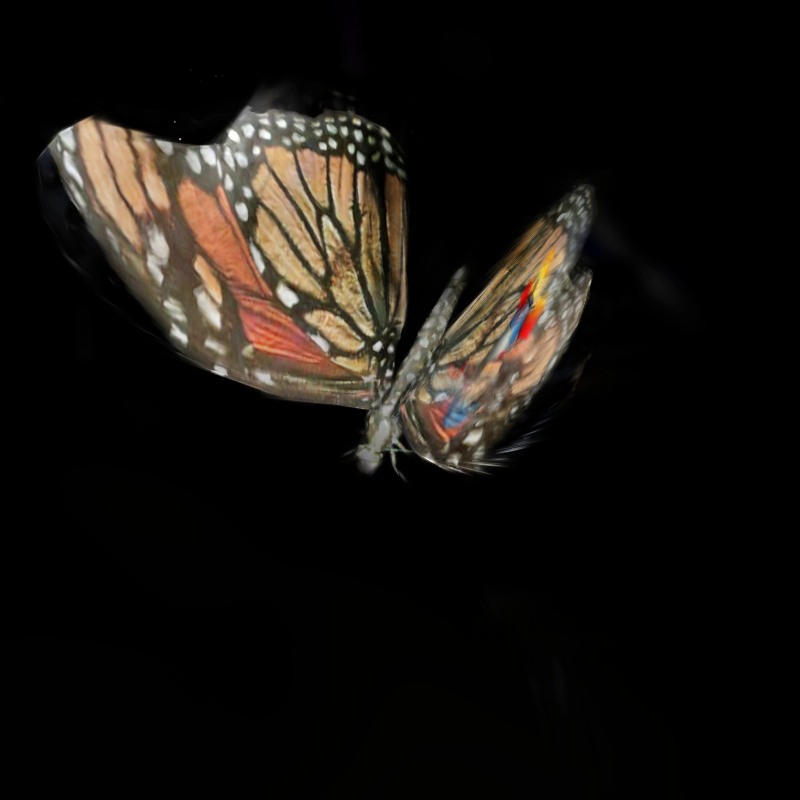}
    \\
    Texture Map from Texture-GS
    &   Edited Texture Map
    &   Ours Edit 1
    &   Texture-GS Edit 1
    &   Ours Edit 2
    &   Texture-GS Edit 2
\end{tabularx}
\caption{
    Simultaneous geometry and texture editing comparisons with Texture-GS~\cite{Xu2024TextureGSDT} on the Butterfly from Objaverse~\cite{Deitke2022ObjaverseAU}. We use the original settings of Texture-GS without capping the number of Gaussians.
}
\label{supp-fig:both-editing-texture-gs}
\end{figure}

Texture-GS~\cite{Xu2024TextureGSDT} supports texture editing through a single-chart global texture map. However, this representation often struggles with complex geometries, resulting in texture maps that are insufficient for reliable texture-space editing in practice (see Figure~\ref{supp-fig:texture-gs-failure}). For a fair comparison, Figure~\ref{supp-fig:both-editing-texture-gs} shows the comparison of simultaneous geometry and texture editing results on the `Butterfly' scene from Objaverse, where Texture-GS successfully generates a valid map. As shown, unlike Texture-GS, which suffers from broken surface continuity and loss of resolution after deformation, our method ensures that texture edits remain sharp and continuous.

Textured Gaussians~\cite{Chao2024TexturedGF} adopts the same core idea as GSTex~\cite{Rong2024GStexPT}: both methods attach a learnable per-Gaussian texture map to each Gaussian primitive and query it via ray--Gaussian intersection with UV mapping defined by the Gaussian's local coordinate axes. Both also decompose appearance into a spatially varying texture and spatially constant spherical harmonics, and both follow a two-stage optimization that first trains a standard Gaussian model before adding textures. The main differences are that Textured Gaussians builds on 3D Gaussians rather than 2D, uses a fixed texture resolution across all primitives (instead of adapting the resolution to each Gaussian's scale), and optionally includes an alpha channel in the texture map. Given the strong architectural similarity and the fact that texture editing is not the main focus of Textured Gaussians (their released codebase does not include an editing pipeline), we compare against GSTex as a representative method from this family.

NeST-Splatting~\cite{Zhang2025NeuralST} decouples geometry and appearance by replacing per-Gaussian spherical harmonics with a global shell texture, parameterized as a multi-resolution hash grid encoding. During rendering, each ray--Gaussian intersection is queried against the hash grid in \emph{world space} to obtain textures. While this design yields a compact representation that achieves high-fidelity rendering with fewer primitives, the hash grid remains anchored in world space rather than attached to individual Gaussians. Consequently, displacing or deforming the Gaussian geometry causes the ray-Gaussian intersections to sample the hash grid at the wrong locations. NeST-Splatting also does not provide an editable texture pipeline in their released codebase. As a result, NeST-Splatting is unable to perform geometry editing without the use of additional proxies, such as cages~\cite{xu2022cagedeforming} or explicit meshes~\cite{yuan2022nerfedit}, to establish a correspondence between the deformed and canonical space.

\section{Additional Novel View Synthesis Results}

\begin{figure*}[htb]
\centering
\setlength\tabcolsep{0pt}
\renewcommand{\arraystretch}{1}
\scriptsize
\begin{tabularx}{1.0\linewidth}{l@{\hskip 0.2em}YYYYYYYY}
\rotatebox[origin=c]{90}{Chair}
    &   \includegraphics[width=\hsize,valign=m]{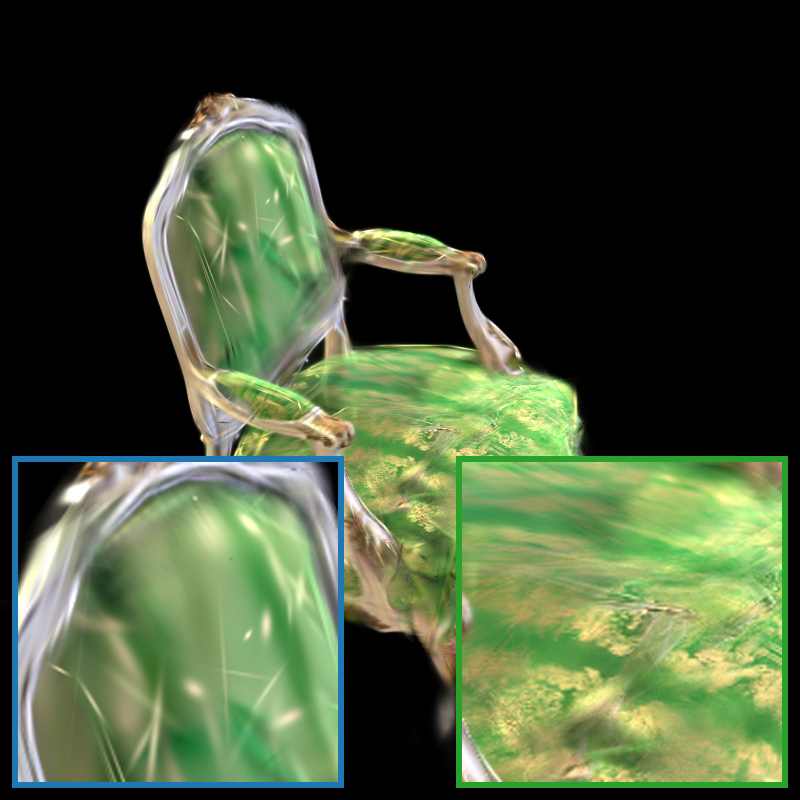}
    &   \includegraphics[width=\hsize,valign=m]{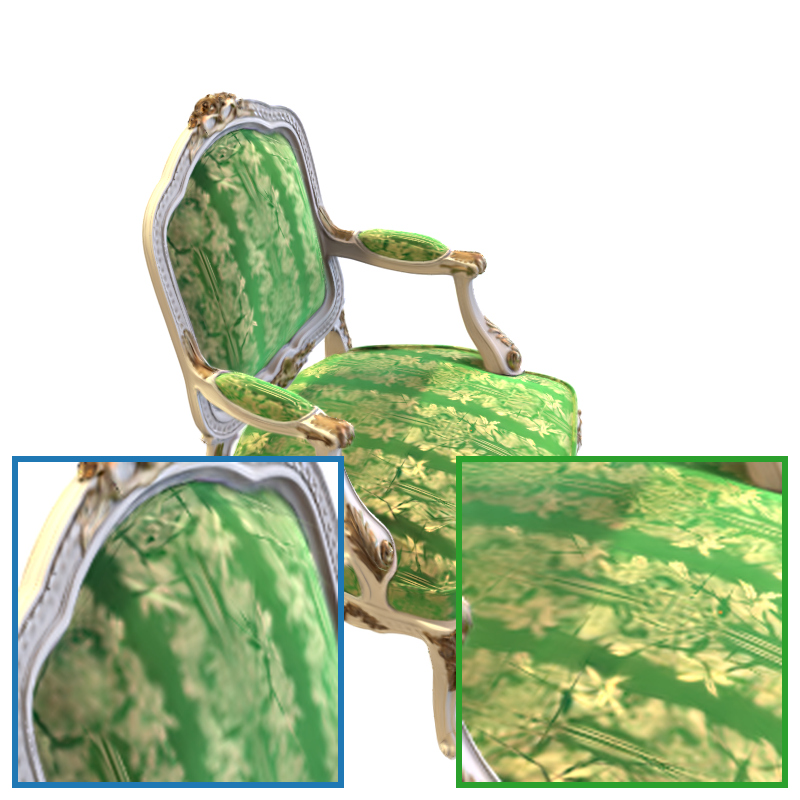}
    &   \includegraphics[width=\hsize,valign=m]{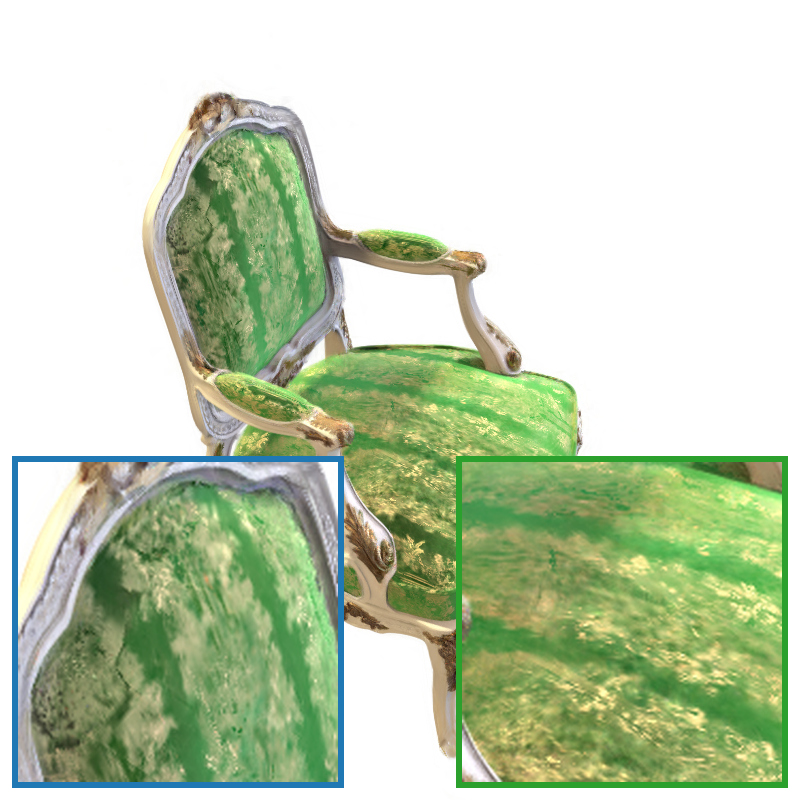}
    &   \includegraphics[width=\hsize,valign=m]{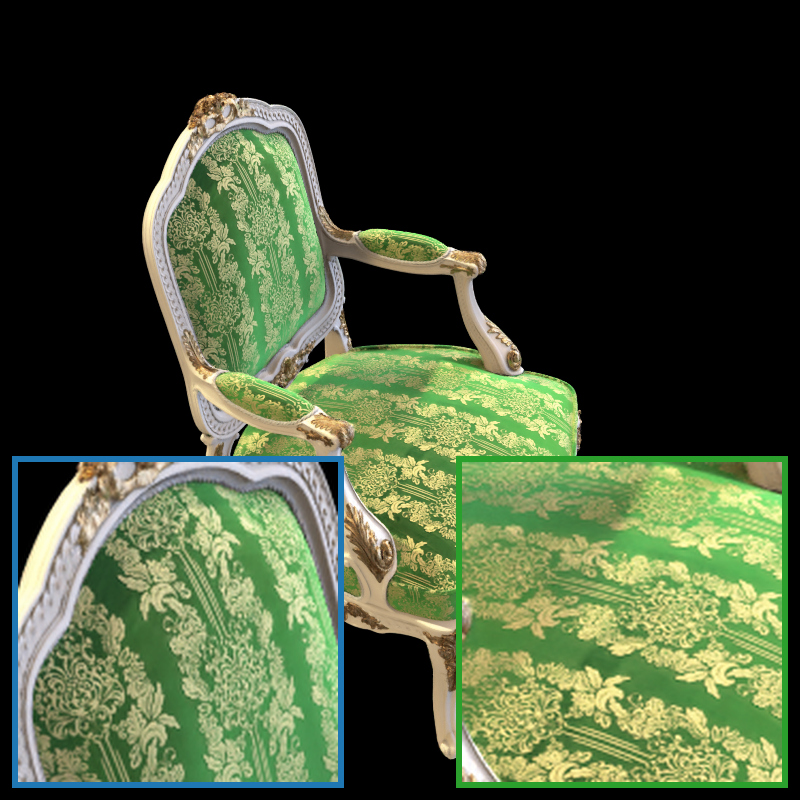}
    &   \includegraphics[width=\hsize,valign=m]{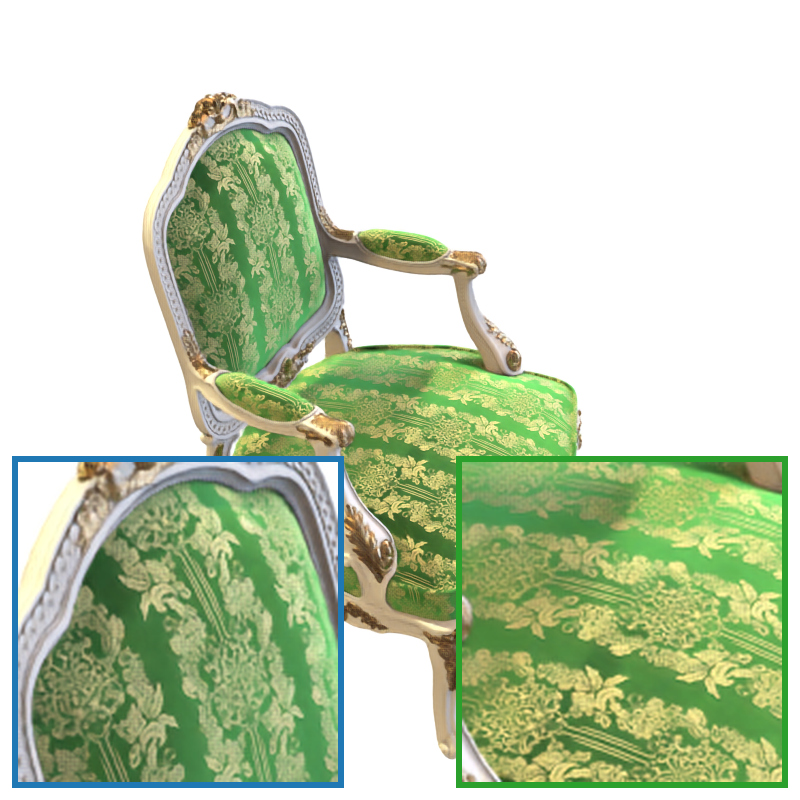}
    &   \includegraphics[width=\hsize,valign=m]{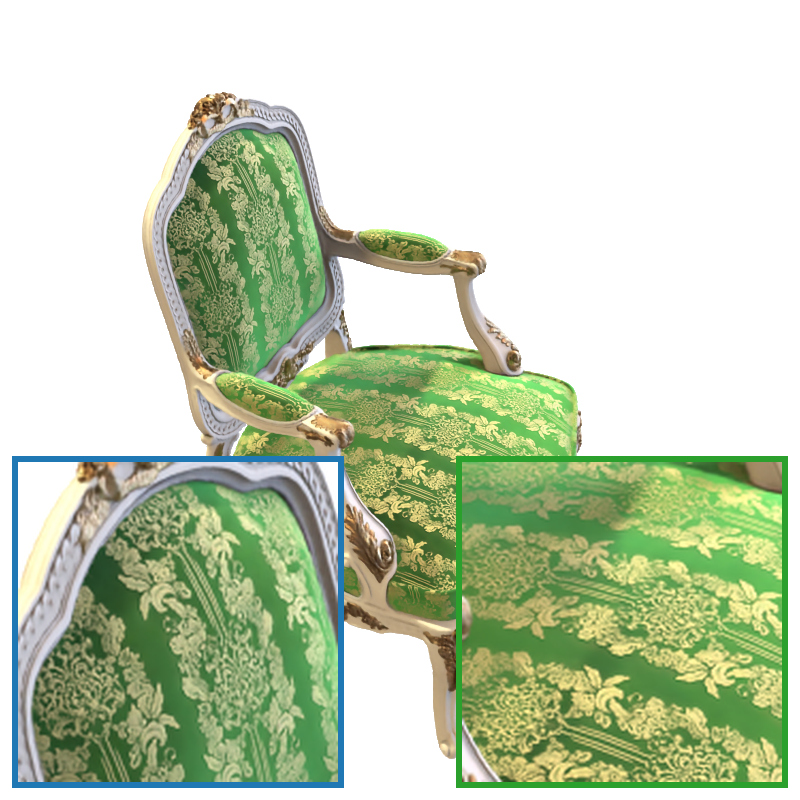}
    &   \includegraphics[width=\hsize,valign=m]{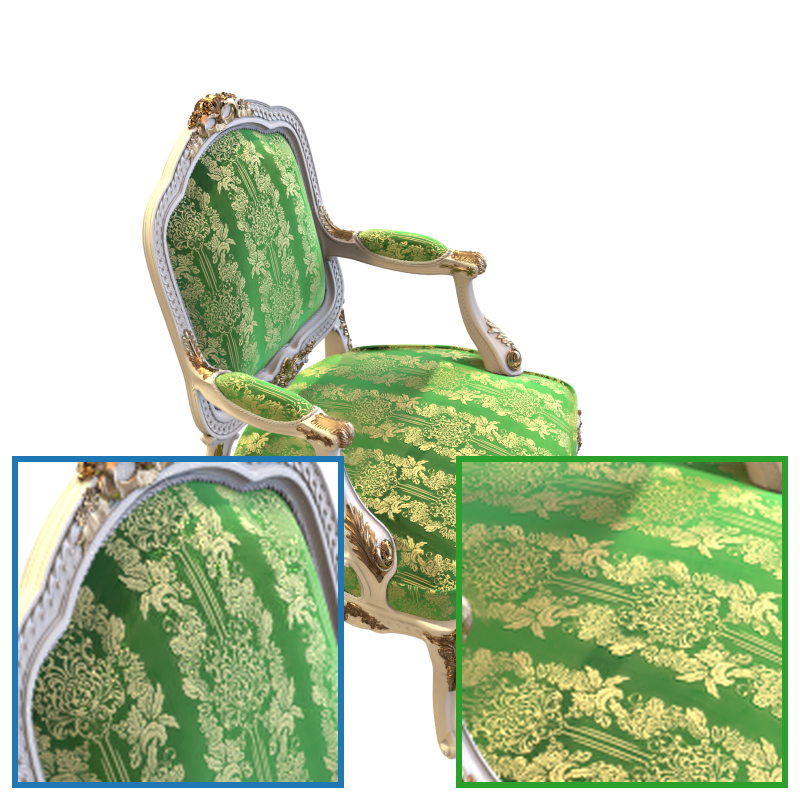}
    \\  
\rotatebox[origin=c]{90}{Hotdog}
    &   \includegraphics[width=\hsize,valign=m]{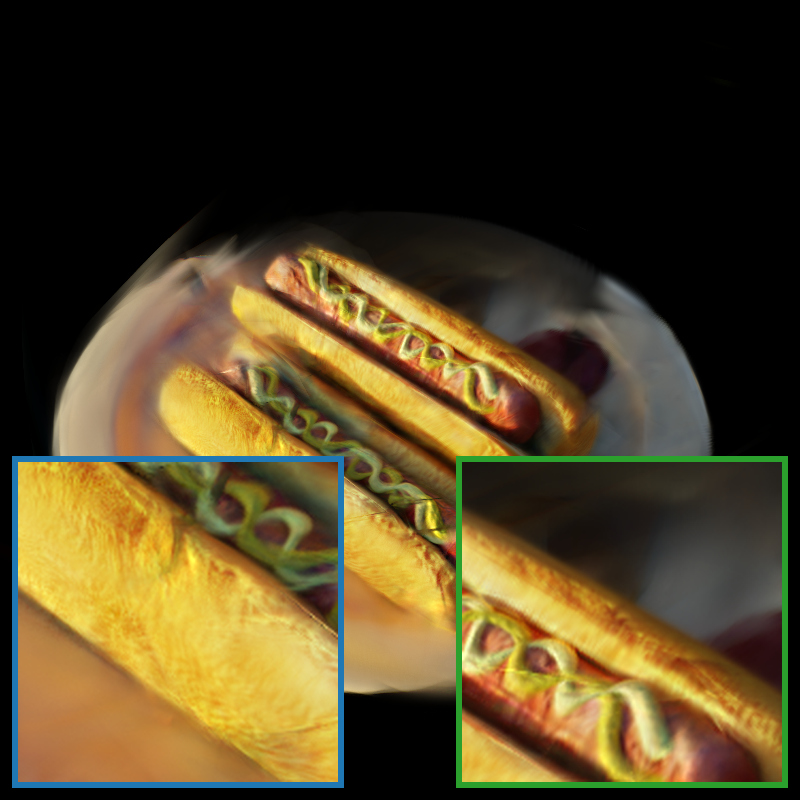}
    &   \includegraphics[width=\hsize,valign=m]{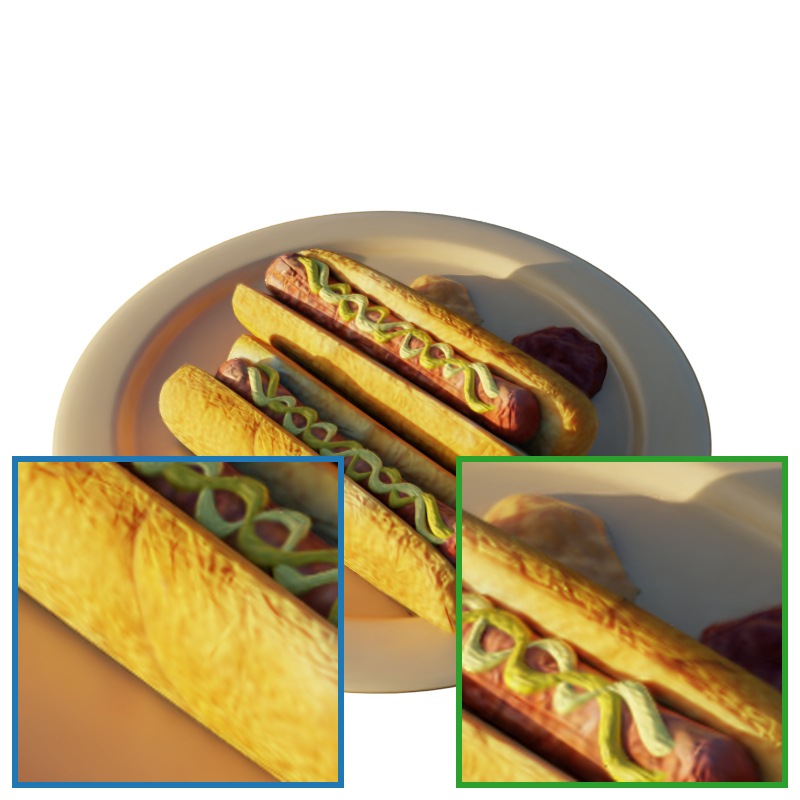}
    &   \includegraphics[width=\hsize,valign=m]{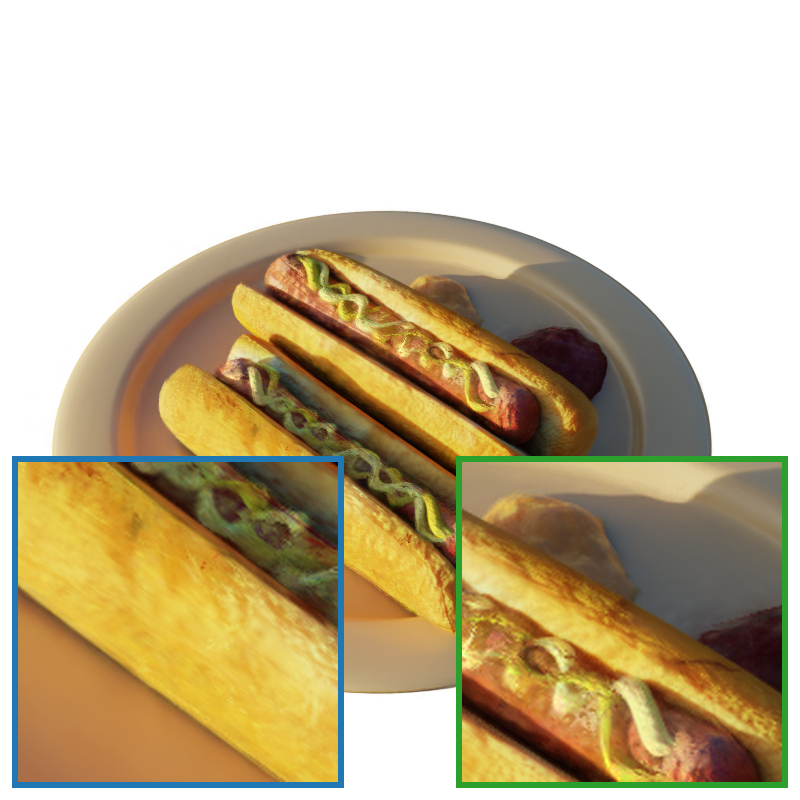}
    &   \includegraphics[width=\hsize,valign=m]{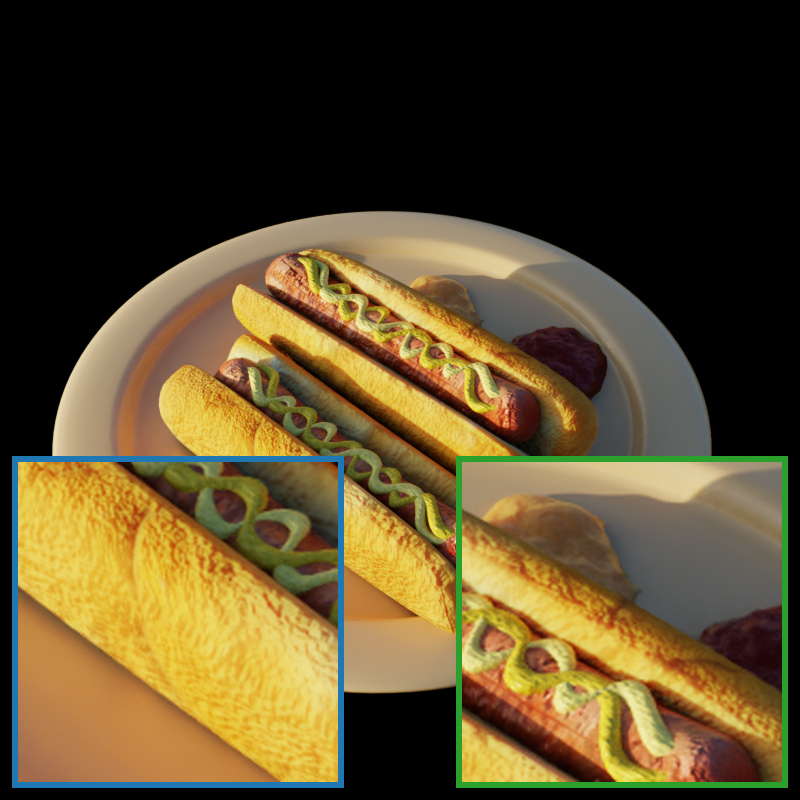}
    &   \includegraphics[width=\hsize,valign=m]{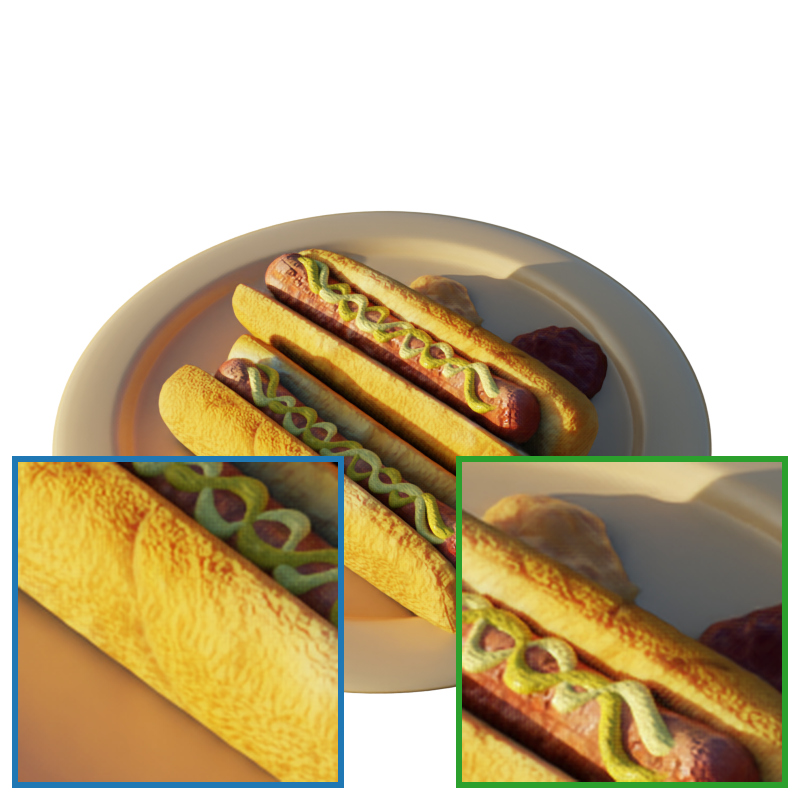}
    &   \includegraphics[width=\hsize,valign=m]{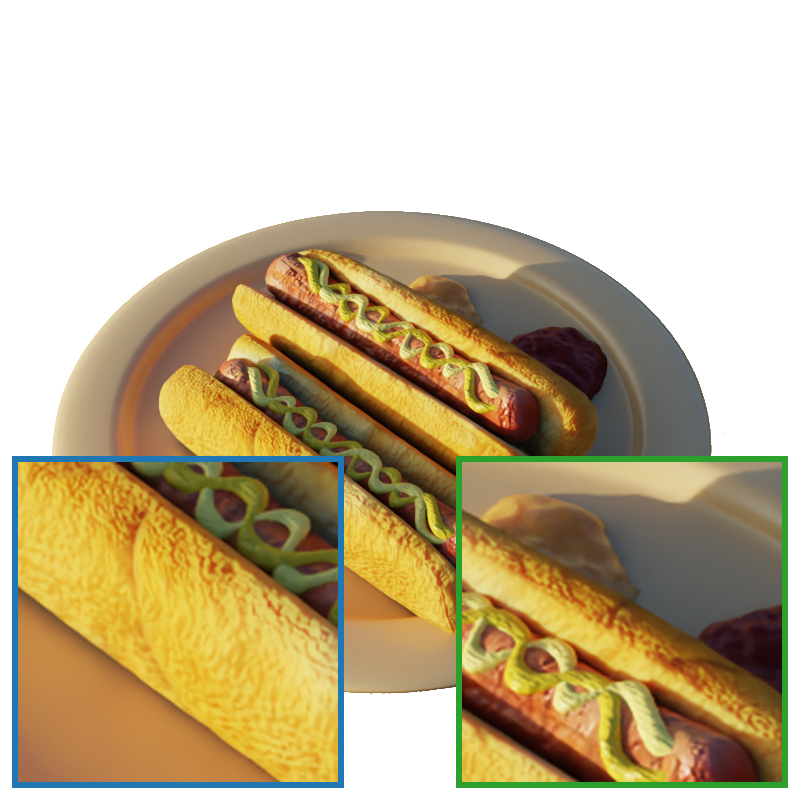}
    &   \includegraphics[width=\hsize,valign=m]{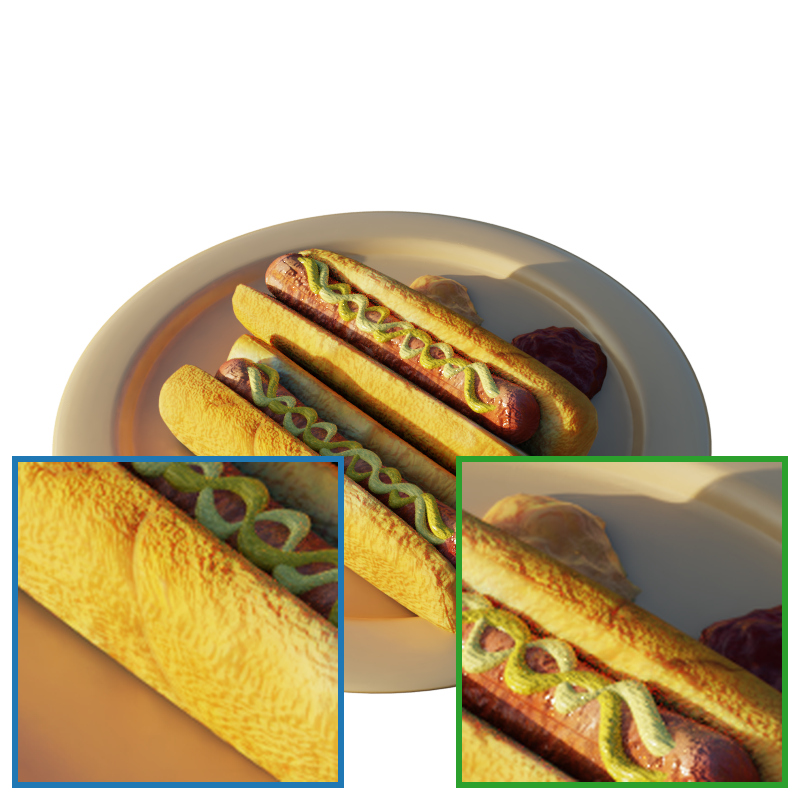}
    \\
\rotatebox[origin=c]{90}{Scan 24}
    &   \includegraphics[width=\hsize,valign=m]{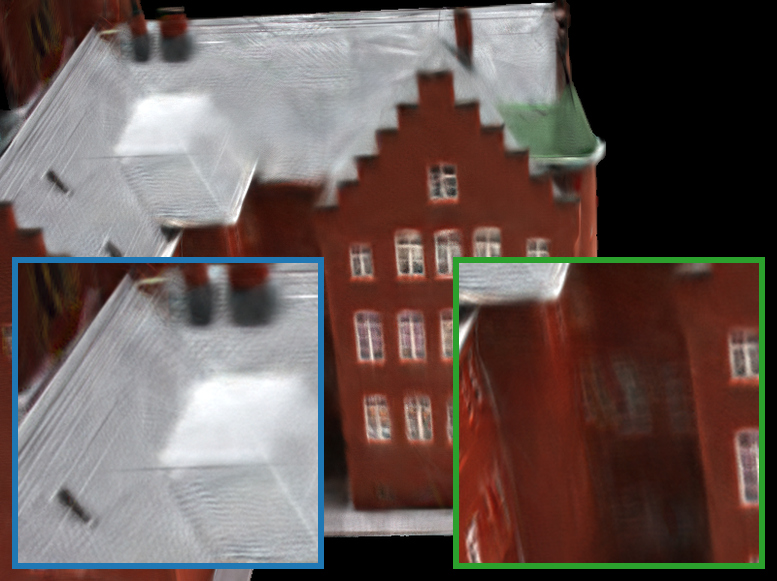}
    &   \includegraphics[width=\hsize,valign=m]{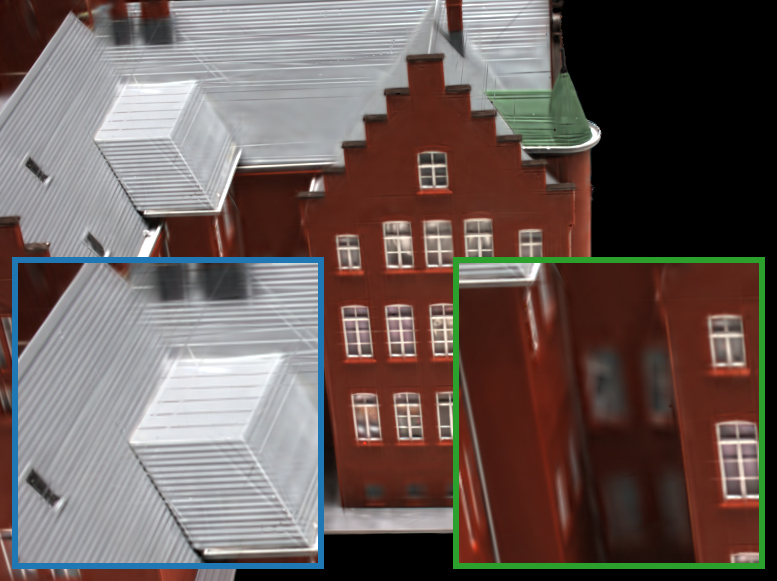}
    &   \includegraphics[width=\hsize,valign=m]{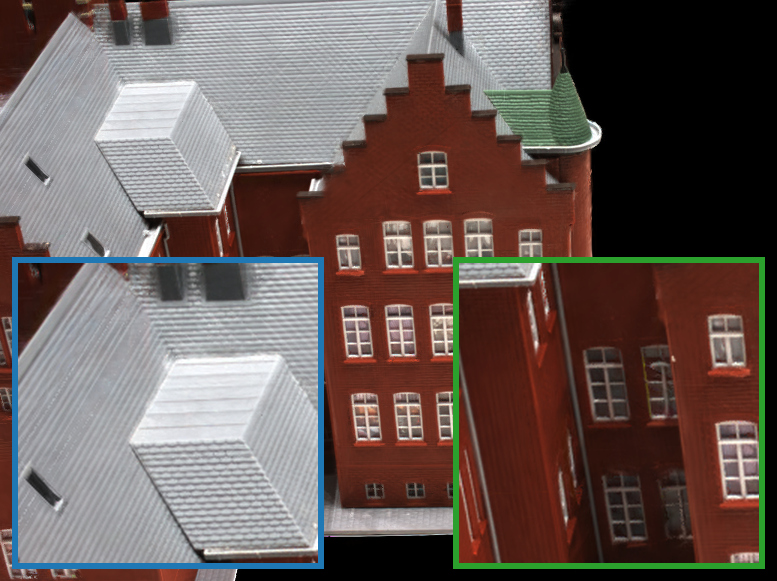}
    &   \includegraphics[width=\hsize,valign=m]{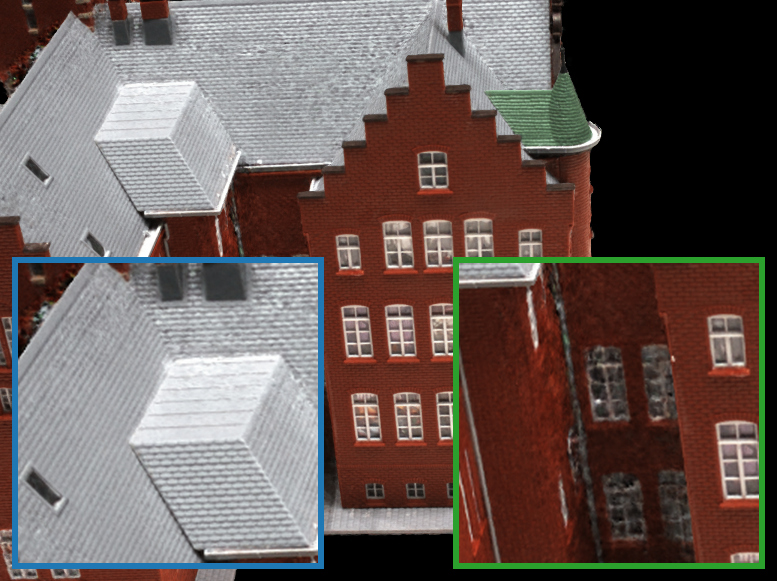}
    &   \includegraphics[width=\hsize,valign=m]{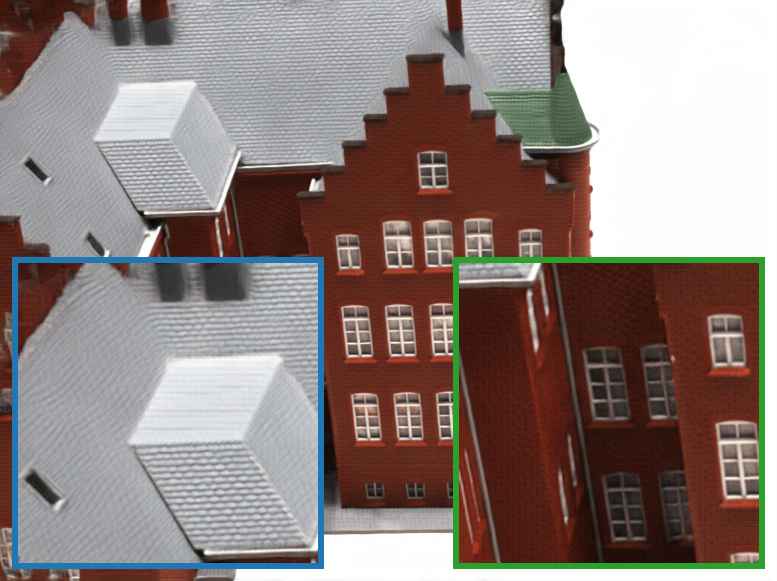}
    &   \includegraphics[width=\hsize,valign=m]{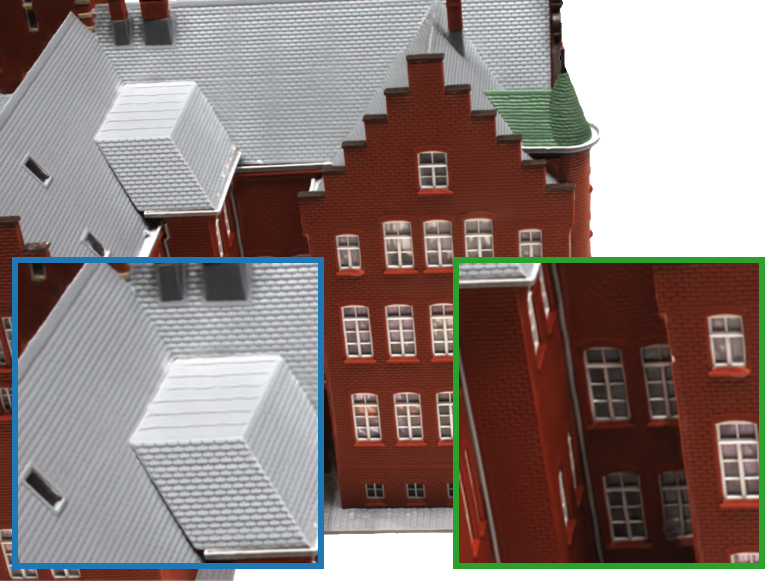}
    &   \includegraphics[width=\hsize,valign=m]{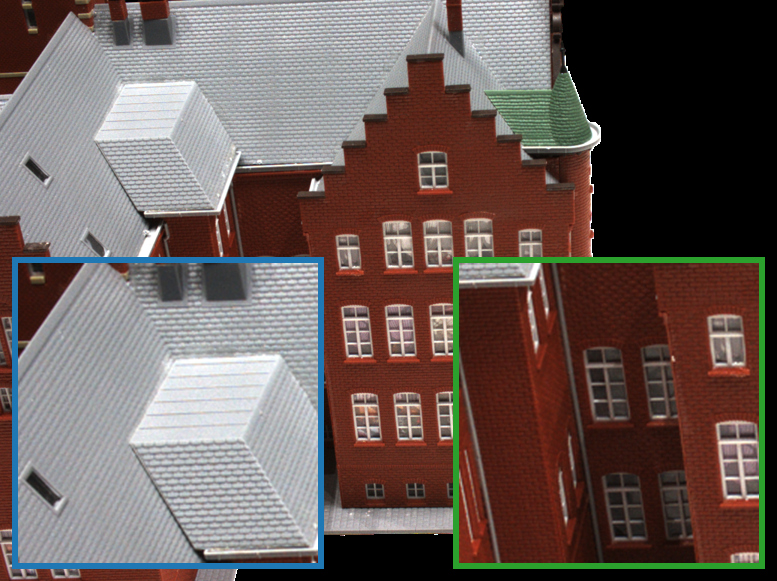}
    \\
\rotatebox[origin=c]{90}{Scan 97}
    &   \includegraphics[width=\hsize,valign=m]{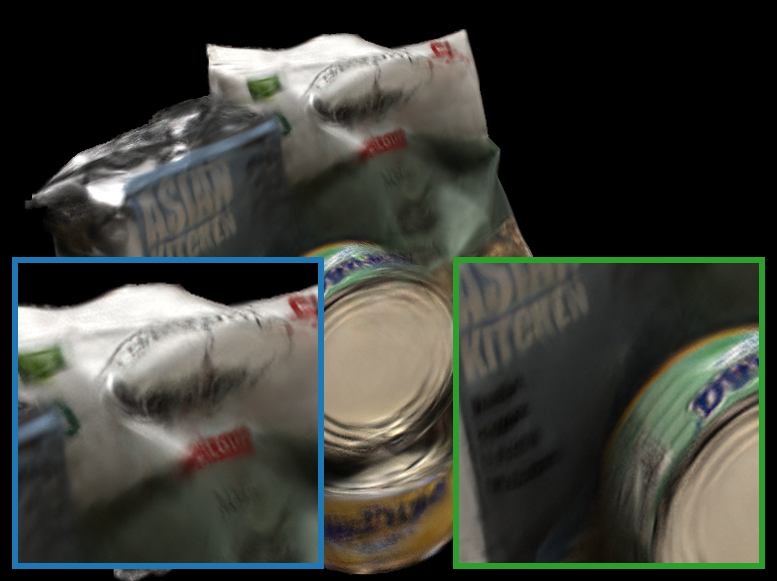}
    &   \includegraphics[width=\hsize,valign=m]{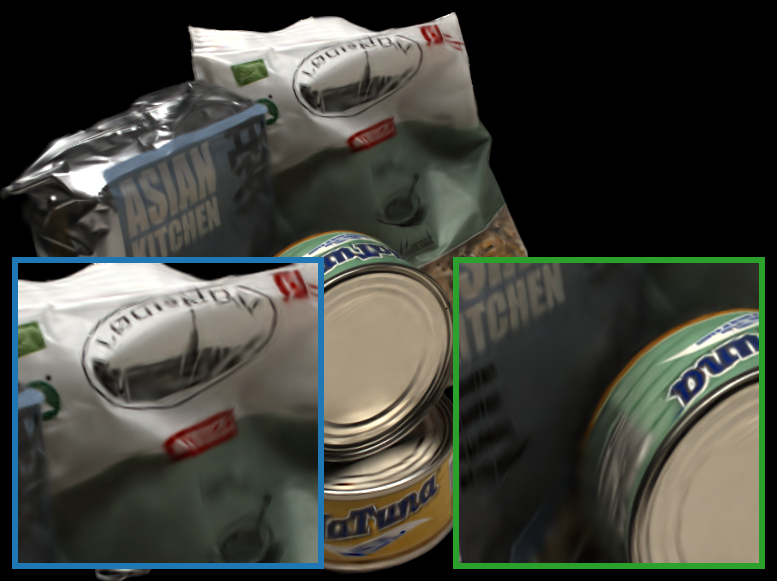}
    &   \includegraphics[width=\hsize,valign=m]{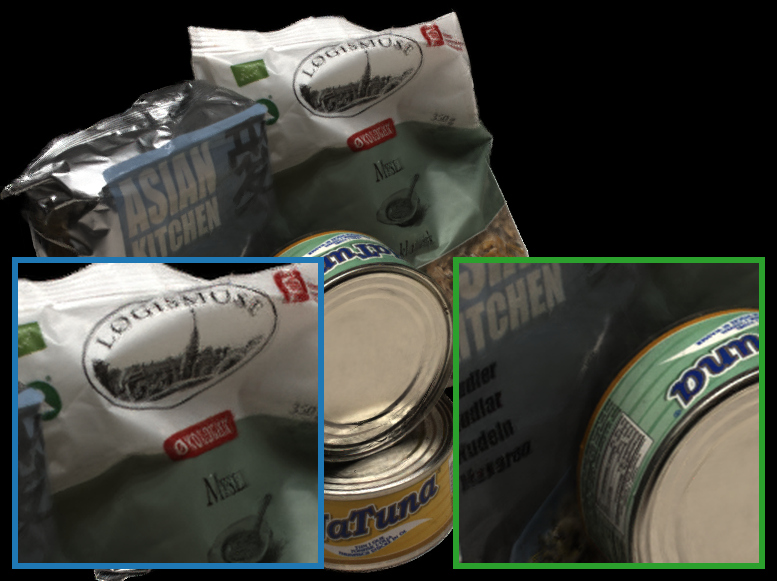}
    &   \includegraphics[width=\hsize,valign=m]{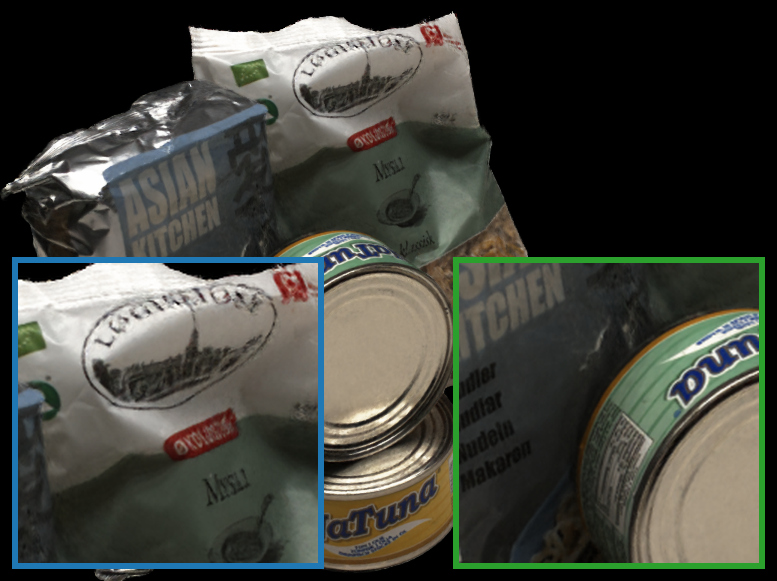}
    &   \includegraphics[width=\hsize,valign=m]{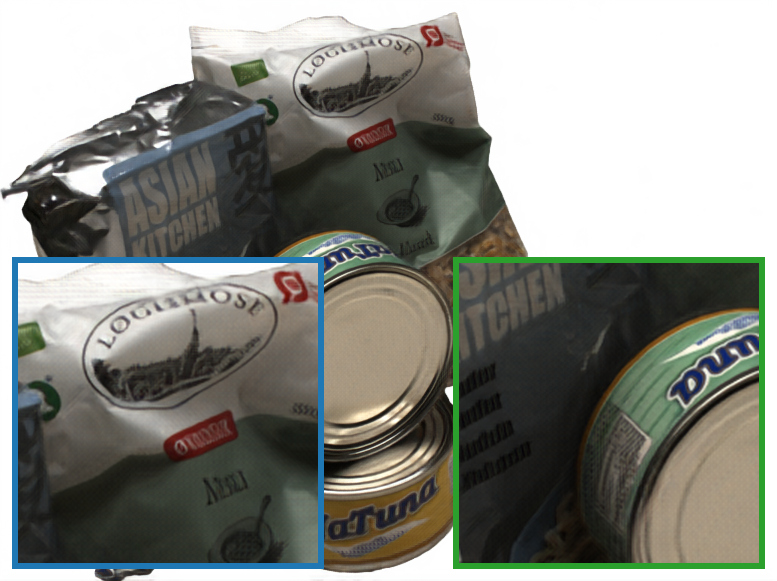}
    &   \includegraphics[width=\hsize,valign=m]{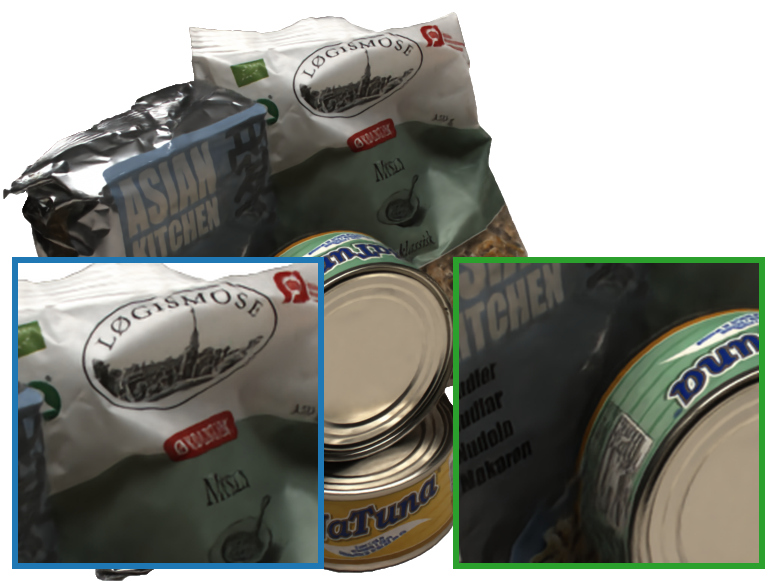}
    &   \includegraphics[width=\hsize,valign=m]{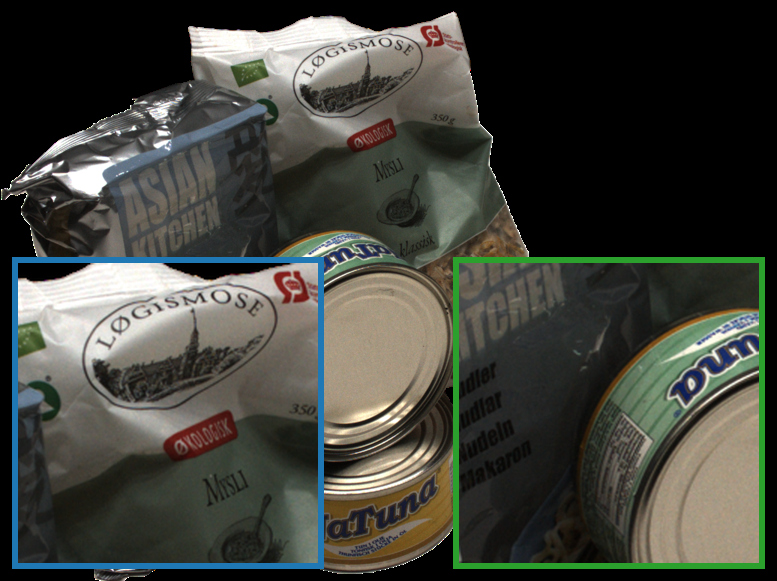}
    \\
    &   Texture-GS
    &   GSTex
    &   Textured-GS
    &   NeST-Splat
    &   PAPR
    &   Ours
    &   GT
\end{tabularx}
\caption{
    Qualitative comparison of novel view synthesis with 5,000 primitives. As shown, our method achieves better rendering fidelity with the same number of points.
}
\label{fig:qualitative-results}
\label{fig:quali-nerf}
\end{figure*}

\begin{figure}[htb]
\centering
\setlength\tabcolsep{1pt}
\renewcommand{\arraystretch}{2}
\scriptsize
\begin{tabularx}{1.0\linewidth}{l@{\hskip 0.2em}YYYYYYYY}
\rotatebox[origin=c]{90}{Chair}
    &   \includegraphics[width=\hsize,valign=m]{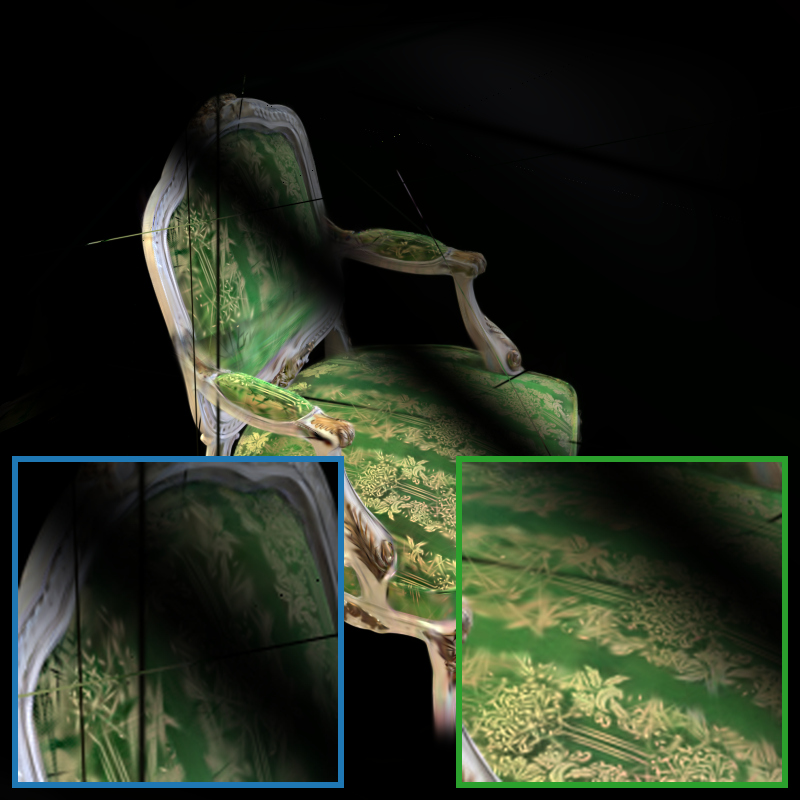}
    &   \includegraphics[width=\hsize,valign=m]{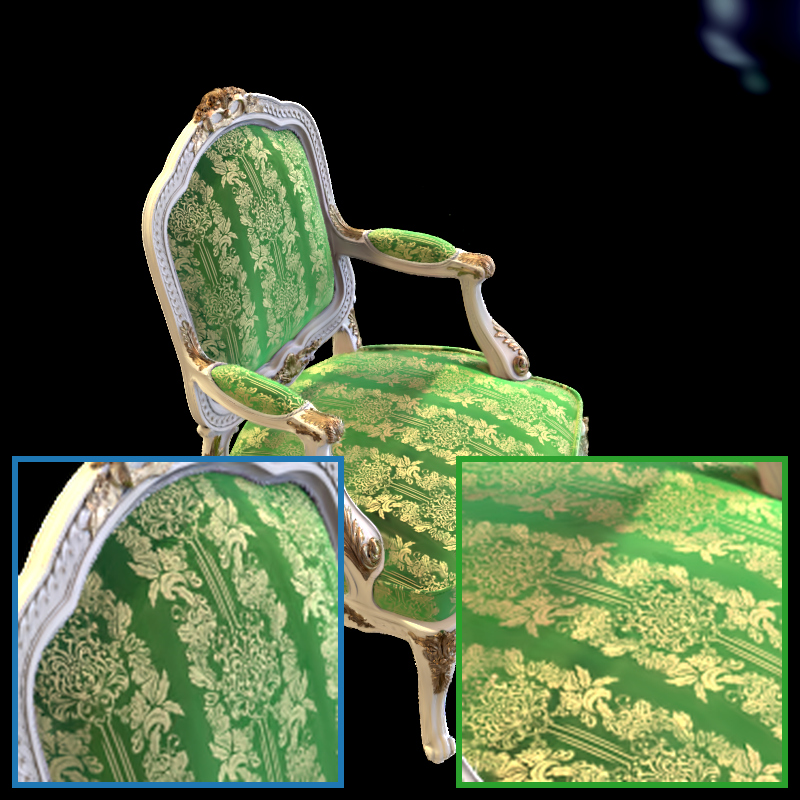}
    &   \includegraphics[width=\hsize,valign=m]{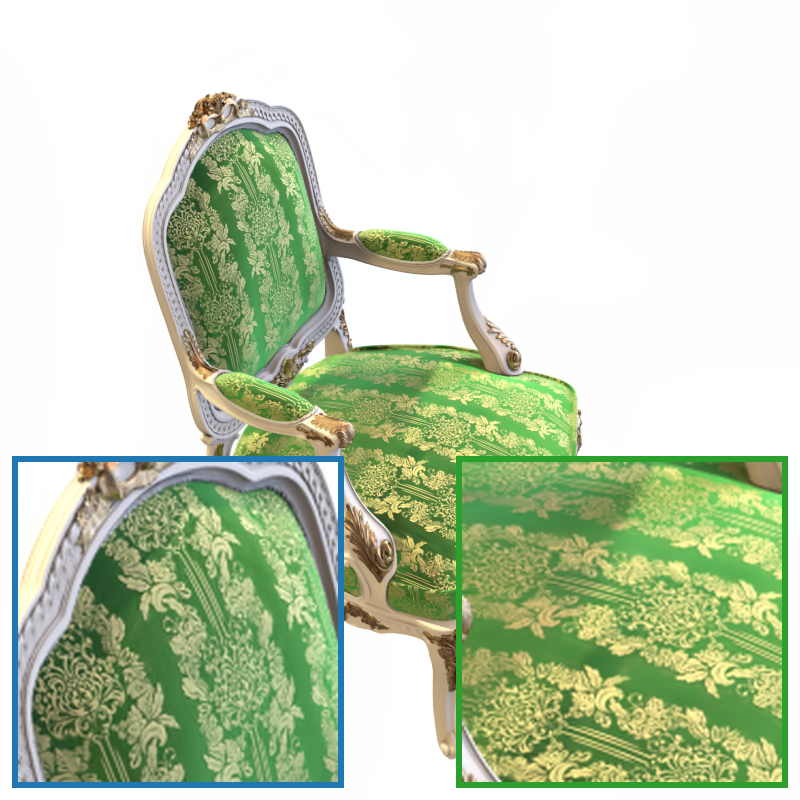}
    &   \includegraphics[width=\hsize,valign=m]{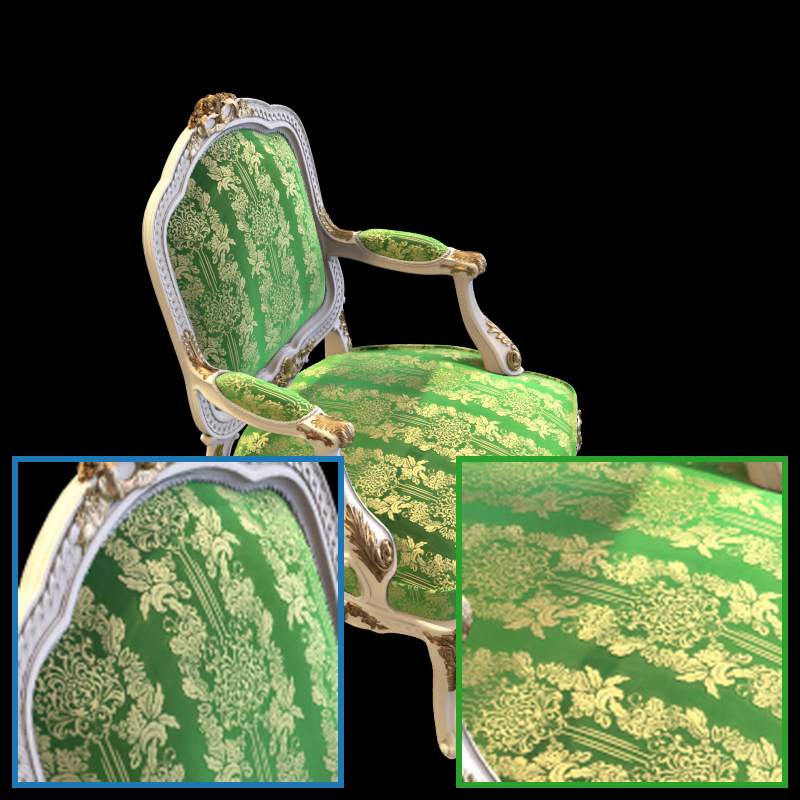}
    &   \includegraphics[width=\hsize,valign=m]{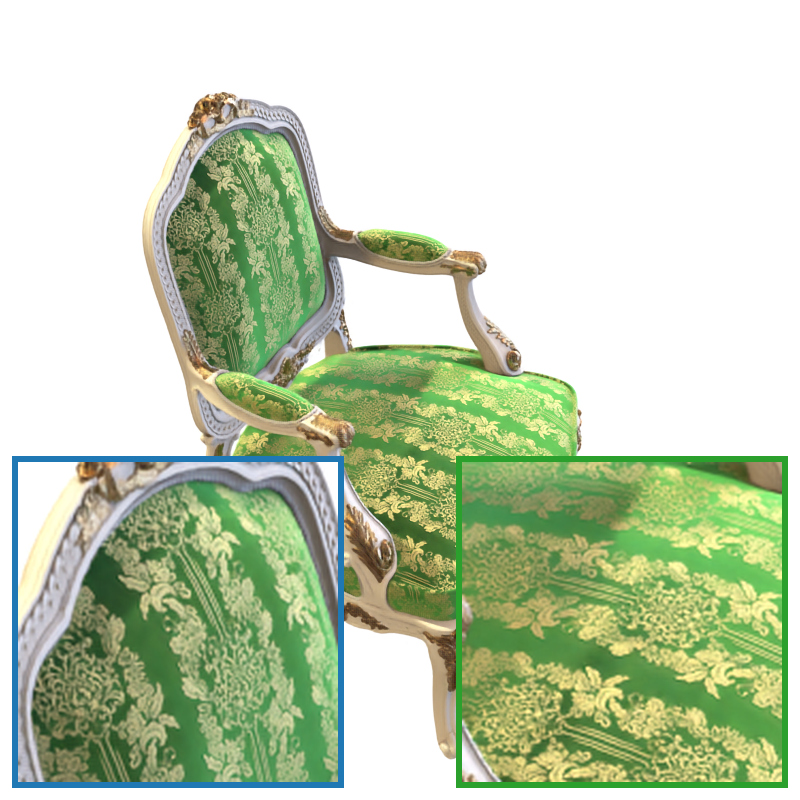}
    &   \includegraphics[width=\hsize,valign=m]{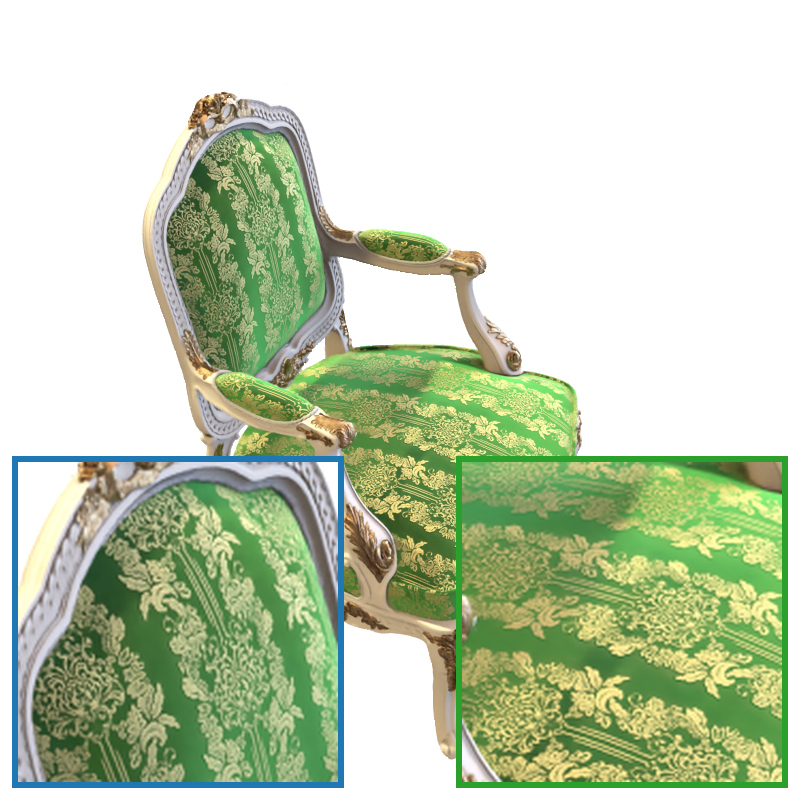}
    &   \includegraphics[width=\hsize,valign=m]{figs/nvs/chair-gt-v32_combined.jpg}
    \\  
\rotatebox[origin=c]{90}{Hotdog}
    &   \includegraphics[width=\hsize,valign=m]{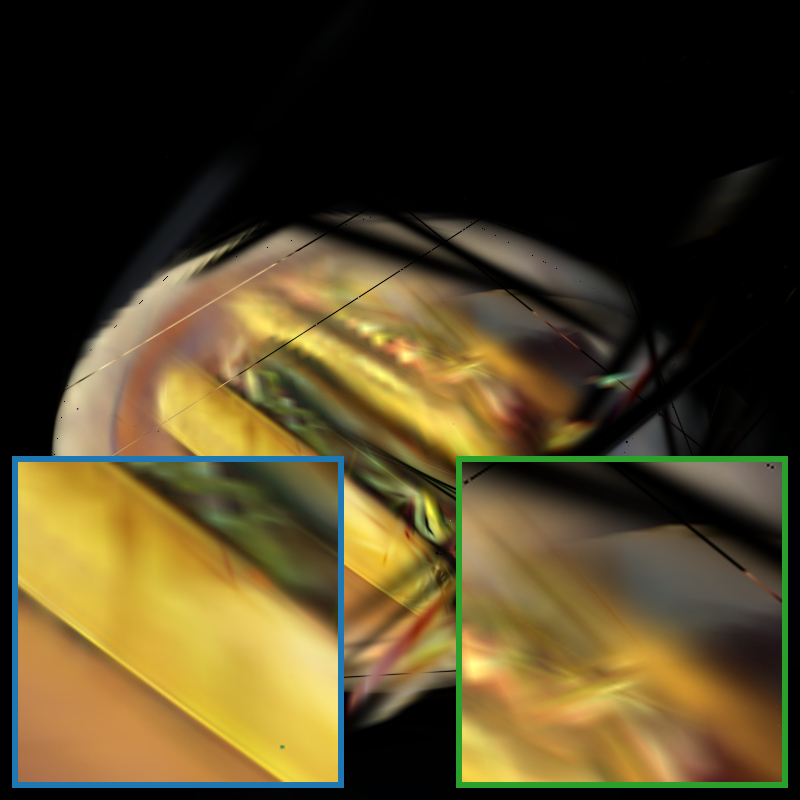}
    &   \includegraphics[width=\hsize,valign=m]{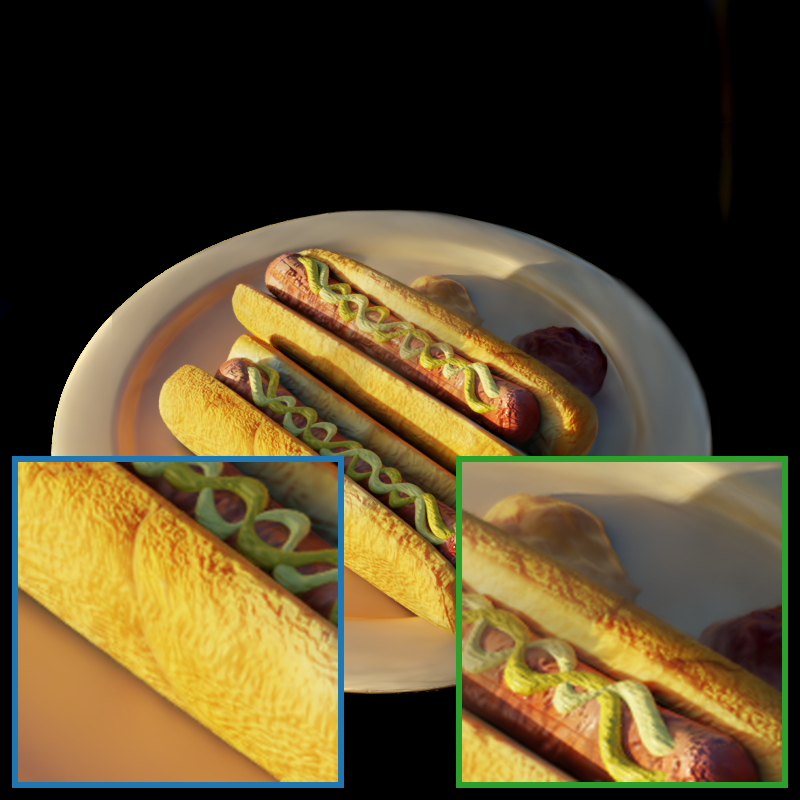}
    &   \includegraphics[width=\hsize,valign=m]{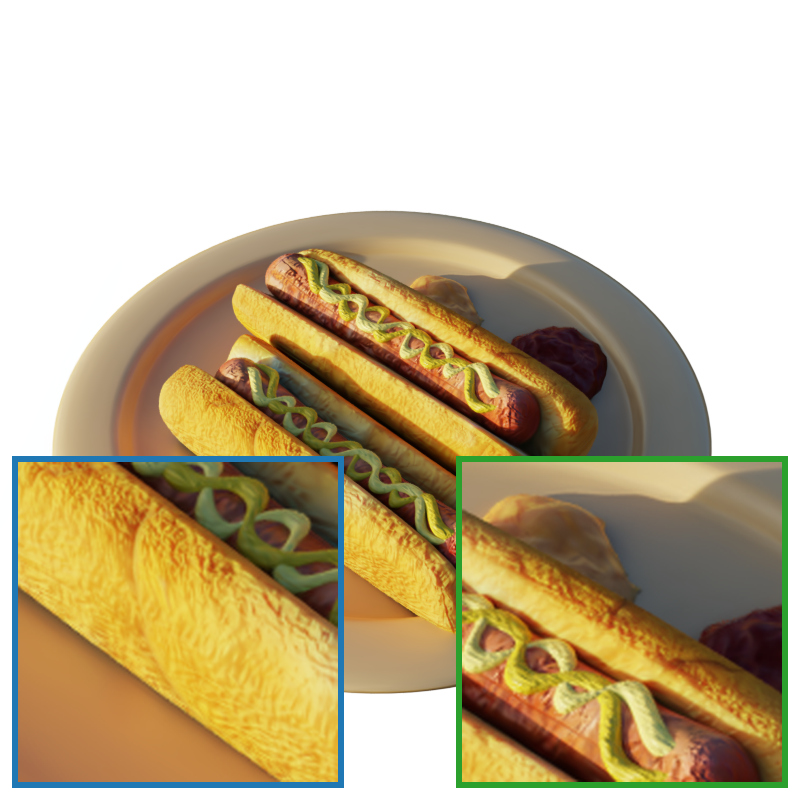}
    &   \includegraphics[width=\hsize,valign=m]{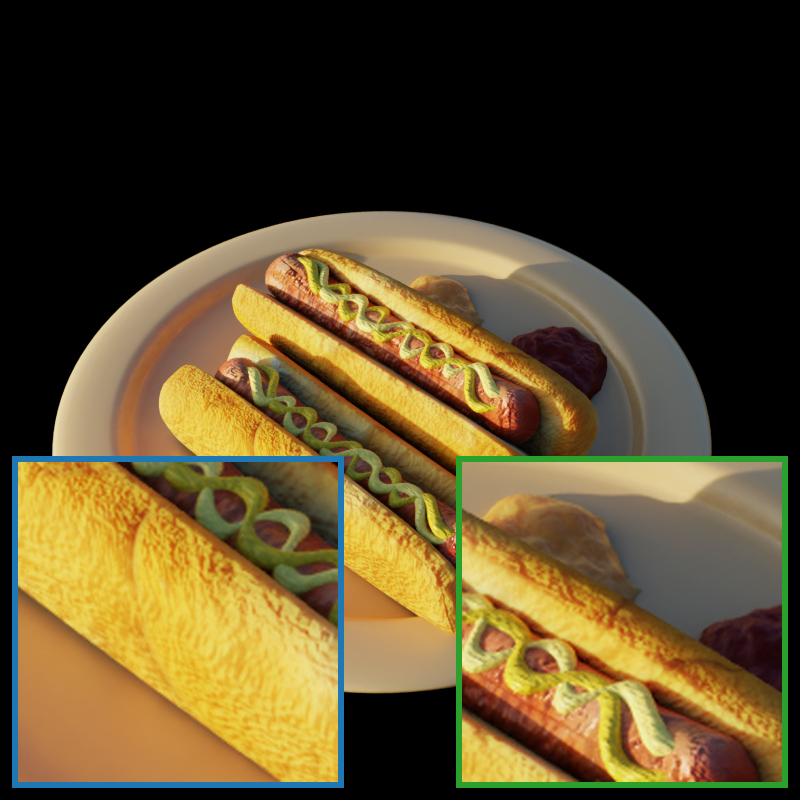}
    &   \includegraphics[width=\hsize,valign=m]{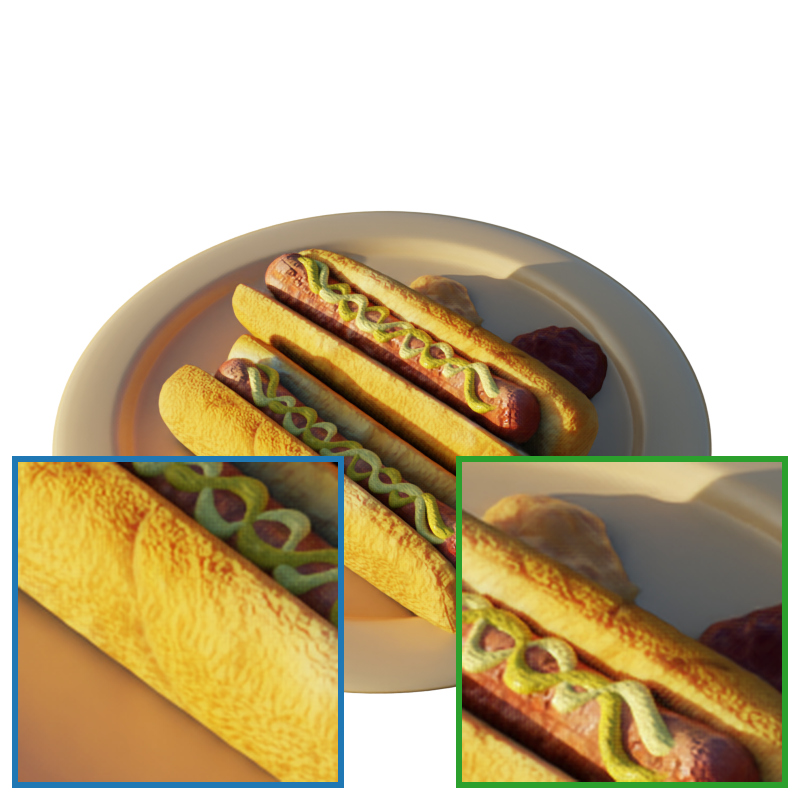}
    &   \includegraphics[width=\hsize,valign=m]{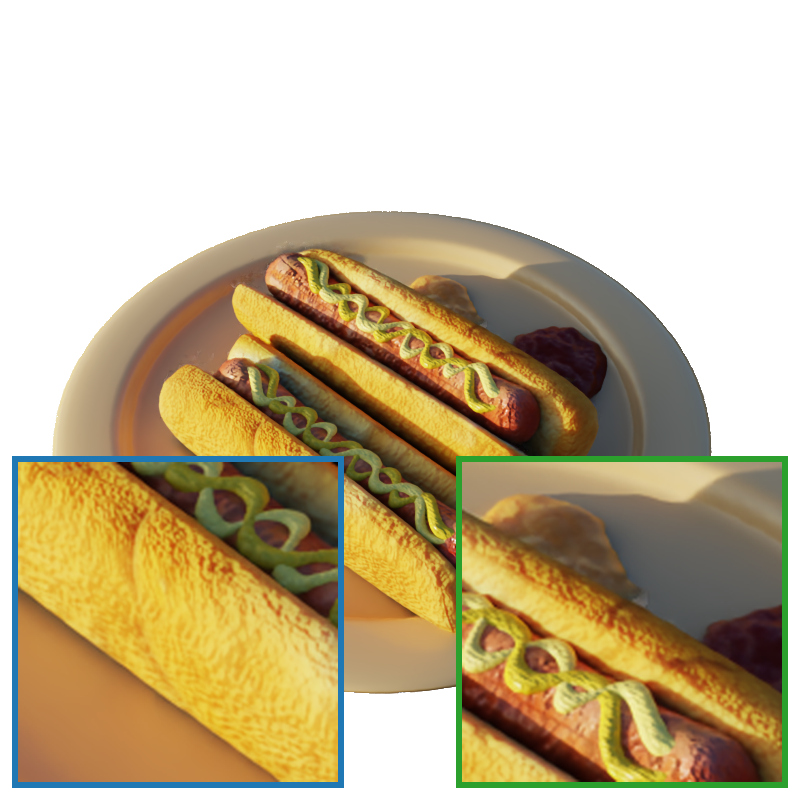}
    &   \includegraphics[width=\hsize,valign=m]{figs/nvs/hotdog-gt-v187_combined.jpg}
    \\
\rotatebox[origin=c]{90}{Scan 24}
    &   \includegraphics[width=\hsize,valign=m]{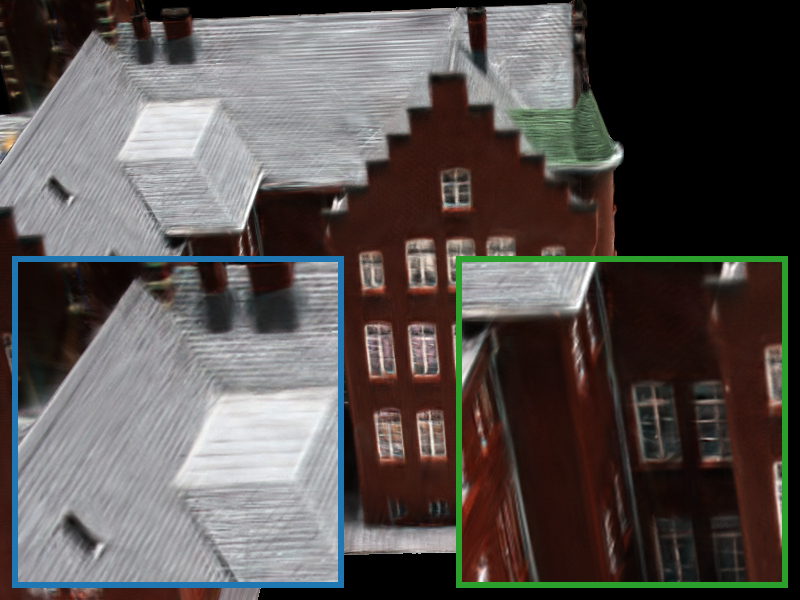}
    &   \includegraphics[width=\hsize,valign=m]{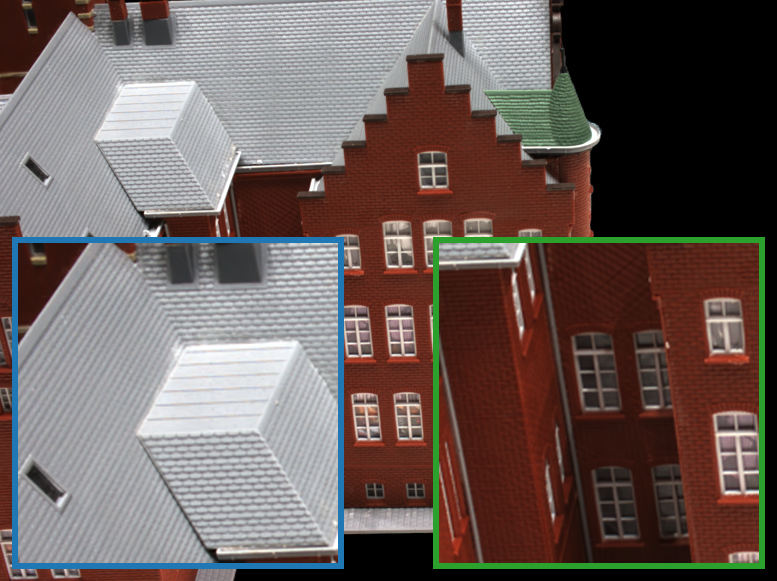}
    &   \includegraphics[width=\hsize,valign=m]{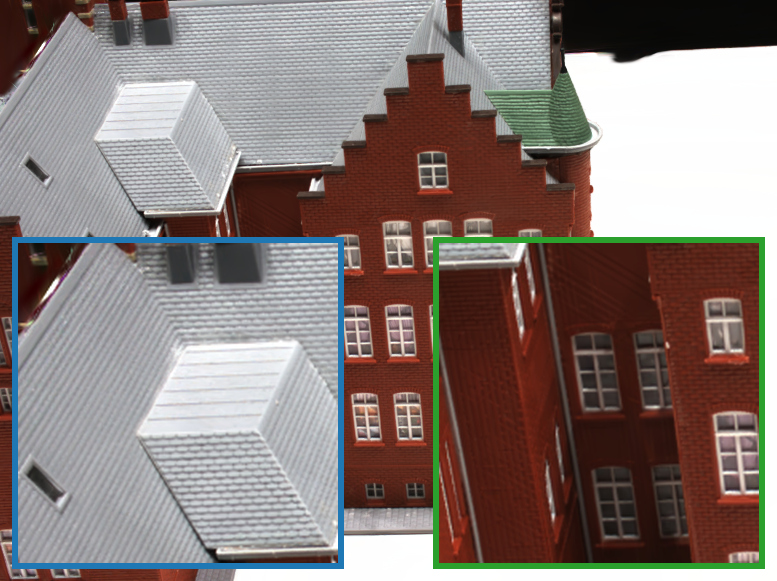}
    &   \includegraphics[width=\hsize,valign=m]{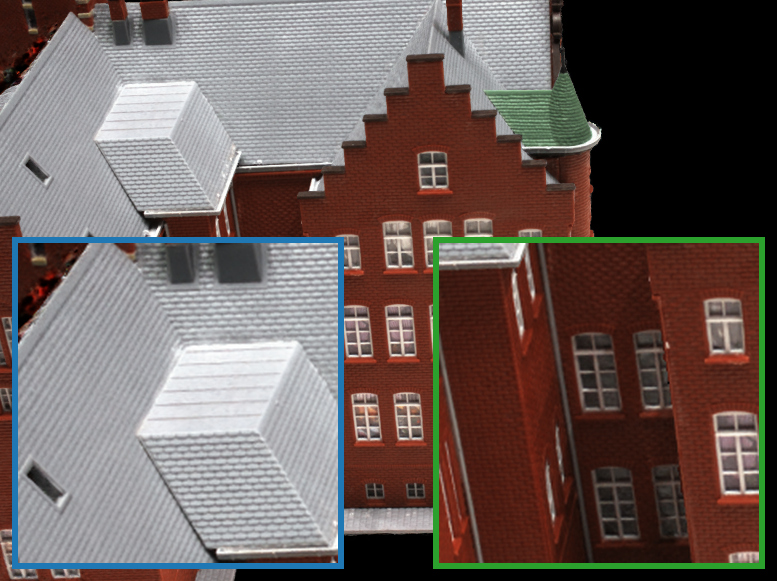}
    &   \includegraphics[width=\hsize,valign=m]{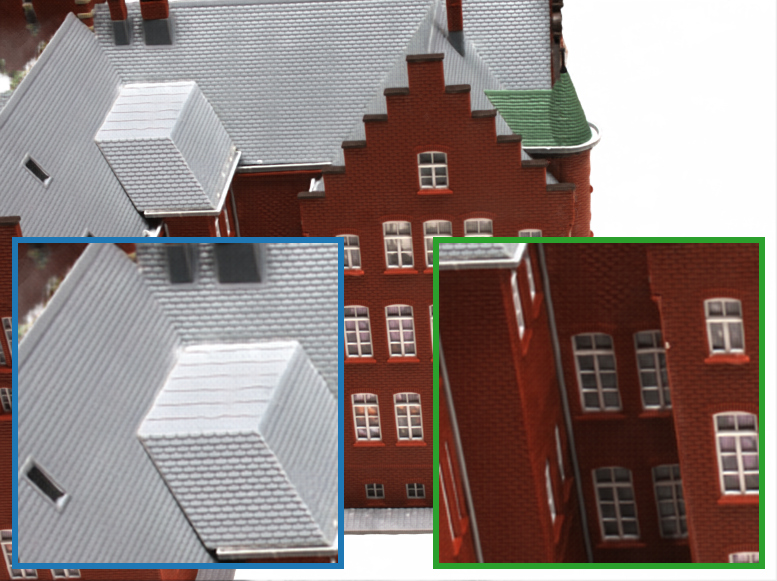}
    &   \includegraphics[width=\hsize,valign=m]{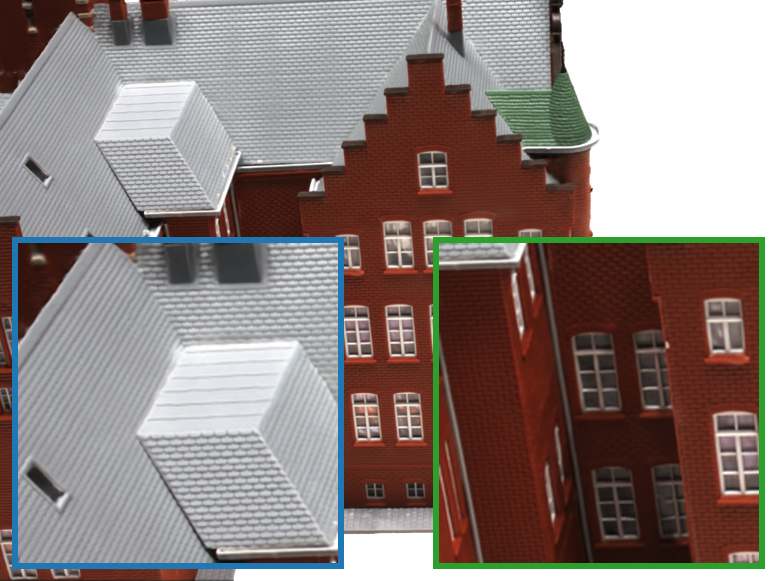}
    &   \includegraphics[width=\hsize,valign=m]{figs/nvs/scan24-gt-v26_combined.jpg}
    \\
\rotatebox[origin=c]{90}{Scan 97}
    &   \includegraphics[width=\hsize,valign=m]{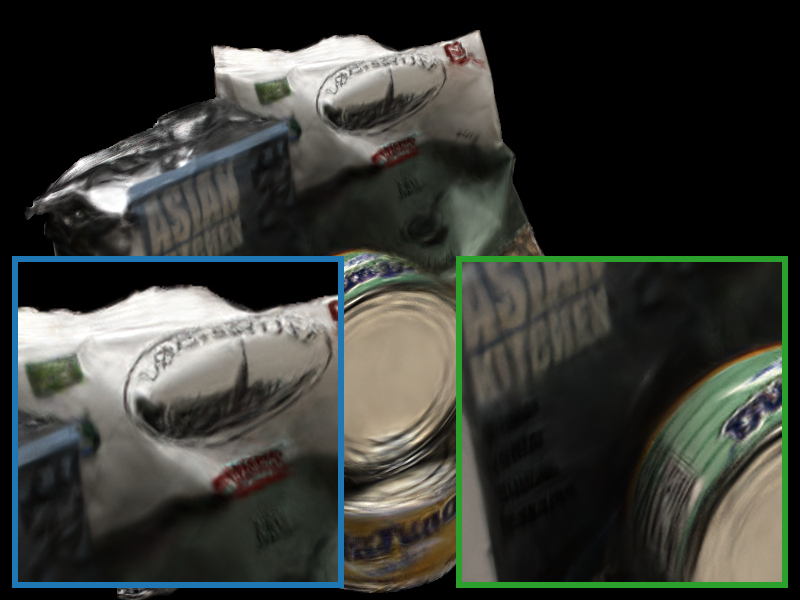}
    &   \includegraphics[width=\hsize,valign=m]{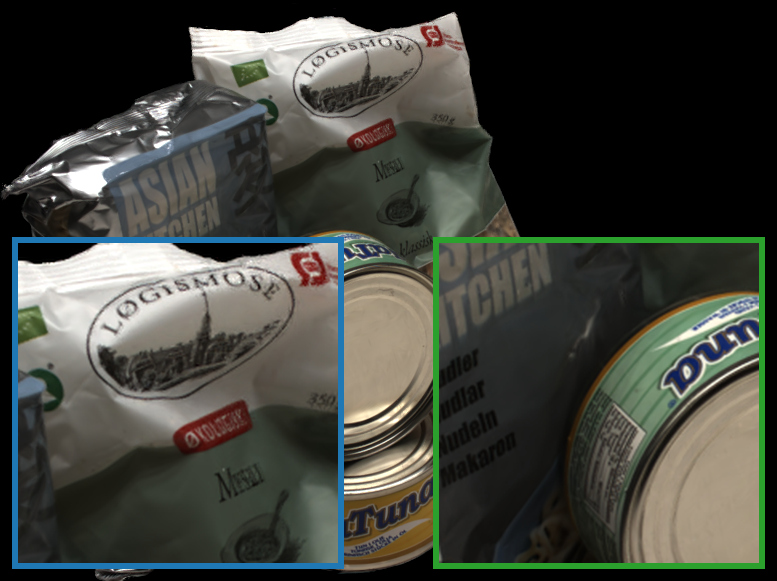}
    &   \includegraphics[width=\hsize,valign=m]{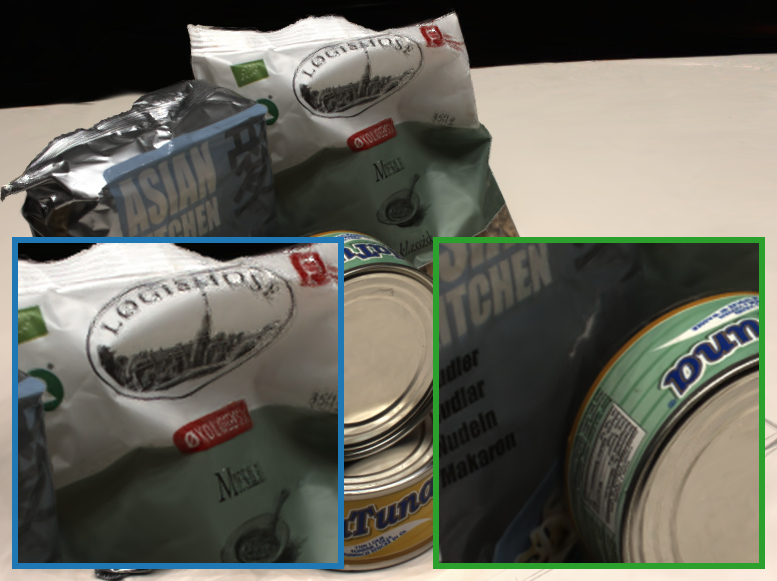}
    &   \includegraphics[width=\hsize,valign=m]{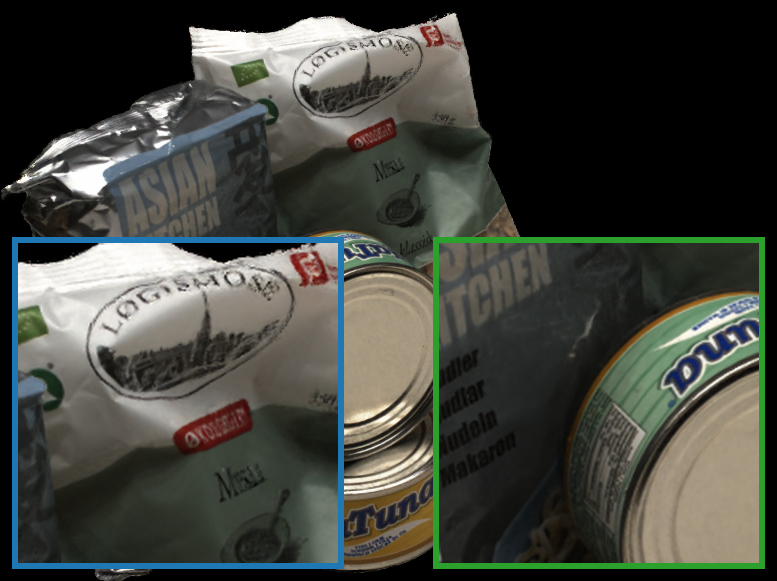}
    &   \includegraphics[width=\hsize,valign=m]{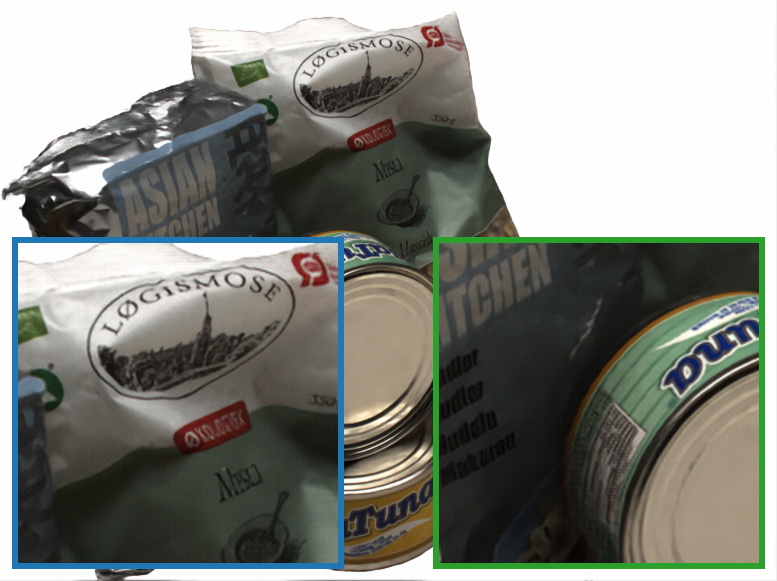}
    &   \includegraphics[width=\hsize,valign=m]{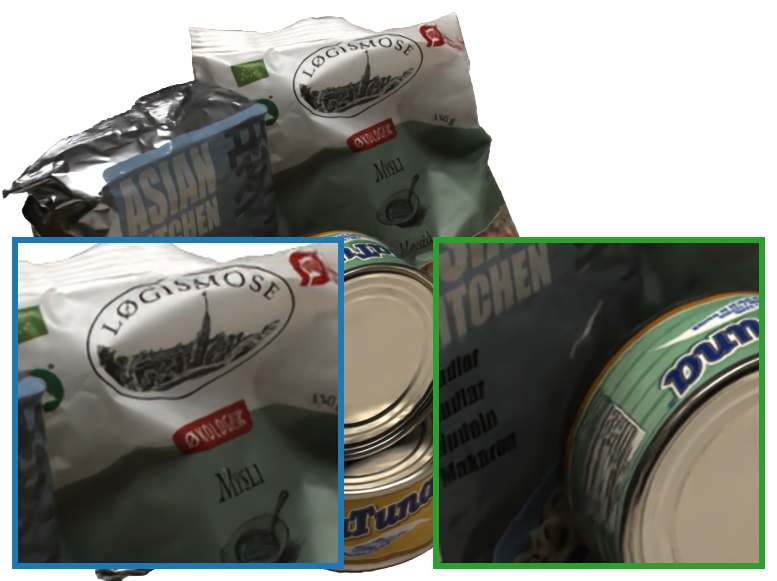}
    &   \includegraphics[width=\hsize,valign=m]{figs/nvs/scan97-gt-v2_combined.jpg}
    \\
    &   Texture-GS
    &   GSTex
    &   Textured-GS
    &   NeST-Splat
    &   PAPR
    &   Ours
    &   GT
\end{tabularx}
\caption{
    Qualitative comparison of novel view synthesis with the original number of primitives proposed by the authors.
}
\label{supp-fig:qualitative-results}
\end{figure}

We show qualitative novel view synthesis comparison using either 5,000 primitives or their originally suggested number of primitives for each method in Figure~\ref{fig:qualitative-results} and Figure~\ref{supp-fig:qualitative-results}. As shown, our method achieves rendering quality comparable to or better than the baselines while using only 30,000 primitives.

We further evaluate PointGT for novel view synthesis on the real-world scenes of the MipNeRF360 dataset~\cite{Barron2021MipNeRF3U}. Table~\ref{tab:m360_nvs} reports per-scene PSNR, SSIM, and LPIPS. We compare against the 3D Gaussian Splatting~\cite{Kerbl20233DGS}, the textured Gaussian baselines, and the PAPR~\cite{zhang2023papr}. As full resolution training exceeds the memory of our GPUs, we evaluate both PAPR and PointGT at \emph{half} the resolution of the MipNeRF360 dataset. The numbers for the Gaussian splatting-based baselines are from their original papers.

\begin{table*}[htb]
  \centering
  \setlength{\tabcolsep}{6pt}
  \scriptsize
  \caption{Quantitative novel view synthesis on the MipNeRF360~\cite{Barron2021MipNeRF3U} scenes. $\ddagger$~Both PAPR and PointGT were trained and rendered at the half-resolution of the dataset, as full resolution evaluation exceeds our GPU memory.}
  \label{tab:m360_nvs}
  \resizebox{\linewidth}{!}{%
  \begin{tabular}{l ccccc c}
  \toprule
  Method & bicycle & counter & flowers & room & stump & Mean \\
  \midrule
  \multicolumn{7}{l}{\emph{PSNR}~$\uparrow$} \\
  3DGS~\cite{Kerbl20233DGS}                    & 24.71 & 28.96 & 21.09 & 31.50 & 26.45 & 26.54 \\
  GSTex~\cite{Rong2024GStexPT}                 & 24.68 & 28.50 & 21.17 & 31.15 & 26.24 & 26.35 \\
  Textured Gaussians~\cite{Chao2024TexturedGF} & --    & --    & --    & --    & --    & 27.35 \\
  NeST-Splatting~\cite{Zhang2025NeuralST}      & 24.49 & 28.45 & 20.05 & 31.30 & 25.87 & 26.03 \\
  \cmidrule(lr){1-7}
  PAPR$^\ddagger$~\cite{zhang2023papr}       & 24.21 & 27.13 & --    & 29.36 & 26.23 & 26.73 \\
  PointGT (Ours)$^\ddagger$                  & \textbf{26.61} & \textbf{28.58} & \textbf{22.40} & \textbf{34.66} & \textbf{27.96} & \textbf{28.04} \\
  \midrule
  \multicolumn{7}{l}{\emph{SSIM}~$\uparrow$} \\
  3DGS~\cite{Kerbl20233DGS}                    & 0.729 & 0.905 & 0.571 & 0.915 & 0.762 & 0.776 \\
  GSTex~\cite{Rong2024GStexPT}                 & 0.730 & 0.896 & 0.582 & 0.910 & 0.758 & 0.775 \\
  Textured Gaussians~\cite{Chao2024TexturedGF} & --    & --    & --    & --    & --    & 0.827 \\
  NeST-Splatting~\cite{Zhang2025NeuralST}      & 0.729 & 0.888 & 0.521 & 0.909 & 0.737 & 0.757 \\
  \cmidrule(lr){1-7}
  PAPR$^\ddagger$~\cite{zhang2023papr}       & 0.736 & 0.838 & --    & 0.822 & 0.818 & 0.804 \\
  PointGT (Ours)$^\ddagger$                  & \textbf{0.757} & \textbf{0.850} & \textbf{0.640} & \textbf{0.962} & \textbf{0.832} & \textbf{0.808} \\
  \midrule
  \multicolumn{7}{l}{\emph{LPIPS}~$\downarrow$} \\
  3DGS~\cite{Kerbl20233DGS}                    & 0.265 & 0.201 & 0.377 & 0.219 & 0.266 & 0.266 \\
  GSTex~\cite{Rong2024GStexPT}                 & 0.265 & 0.221 & 0.365 & 0.237 & 0.248 & 0.267 \\
  Textured Gaussians~\cite{Chao2024TexturedGF} & --    & --    & --    & --    & --    & 0.186 \\
  NeST-Splatting~\cite{Zhang2025NeuralST}      & 0.236 & 0.202 & 0.358 & 0.194 & 0.246 & 0.247 \\
  \cmidrule(lr){1-7}
  PAPR$^\ddagger$~\cite{zhang2023papr}       & 0.201 & 0.226 & --    & 0.209 & 0.202 & 0.210 \\
  PointGT (Ours)$^\ddagger$                  & \textbf{0.199} & \textbf{0.184} & \textbf{0.303} & \textbf{0.049} & \textbf{0.182} & \textbf{0.183} \\
  \bottomrule
  \end{tabular}}
\end{table*}

\end{document}